\documentclass{article}

\usepackage[preprint]{neurips_2026}
\usepackage[utf8]{inputenc}
\usepackage[T1]{fontenc}
\usepackage{booktabs}
\usepackage{hyperref, url, booktabs, amsfonts, amsmath, nicefrac, microtype}
\usepackage{graphicx, xcolor, subcaption}
\usepackage[capitalize,noabbrev]{cleveref}
\crefname{appendix}{App.}{Apps.}
\Crefname{appendix}{App.}{Apps.}
\usepackage{multirow}
\usepackage{xurl}

\providecommand{\doi}[1]{doi:\,\href{https://doi.org/#1}{\nolinkurl{#1}}}

\makeatletter
\newcommand{\myparagraph}{%
  \@startsection{paragraph}{4}{\z@}%
    {1.0ex \@plus 0.2ex \@minus 0.1ex}%
    {-0.5em}%
    {\normalfont\normalsize\bfseries}%
}
\let\paragraph\myparagraph
\makeatother
\definecolor{catambtile}{HTML}{44AA99}
\definecolor{catambink}{HTML}{2E7A6E}
\definecolor{catambfill}{HTML}{238D7C}

\definecolor{catinctile}{HTML}{B8923F}
\definecolor{catincink}{HTML}{8A6A25}
\definecolor{catincfill}{HTML}{9E7927}

\definecolor{catcontile}{HTML}{B05A75}
\definecolor{catconink}{HTML}{8E4259}
\definecolor{catconfill}{HTML}{D35079}

\definecolor{catgentile}{HTML}{7B6D9E}
\definecolor{catgenink}{HTML}{5B4F7E}
\definecolor{catgenfill}{HTML}{8B6BDA}

\newcommand{\catrender}[2]{\textcolor{#1ink}{\textbf{#2}}}             %

\newcommand{\amblabel}{ambiguity}
\newcommand{\Amblabel}{Ambiguity}
\newcommand{\amb}{\amblabel}
\newcommand{\Amb}{\Amblabel}
\newcommand{\catamb}{\catrender{catamb}{\amblabel}}
\newcommand{\Catamb}{\catrender{catamb}{\Amblabel}}

\newcommand{\inclabel}{incompleteness}
\newcommand{\Inclabel}{Incompleteness}
\newcommand{\inc}{\inclabel}

\newcommand{\catinc}{\catrender{catinc}{\inclabel}}
\newcommand{\Catinc}{\catrender{catinc}{\Inclabel}}

\newcommand{\conlabel}{conflicting evidence}
\newcommand{\Conlabel}{Conflicting evidence}
\newcommand{\con}{\conlabel}

\newcommand{\catcon}{\catrender{catcon}{\conlabel}}
\newcommand{\Catcon}{\catrender{catcon}{\Conlabel}}

\newcommand{\genlabel}{general uncertainty}
\newcommand{\Genlabel}{General uncertainty}
\newcommand{\gen}{\genlabel}

\newcommand{\catgen}{\catrender{catgen}{\genlabel}}
\newcommand{\Catgen}{\catrender{catgen}{\Genlabel}}

\newcommand{\catshadecol}[1]{#1tile!30}
\makeatletter
\DeclareRobustCommand{\catshade}[2]{\begingroup\setlength{\fboxsep}{1.2pt}%
  \edef\cs@col{\catshadecol{#1}}\cs@loop#2 \@nil\endgroup}
\def\cs@loop#1 #2\@nil{%
  \colorbox{\cs@col}{\strut #1}%
  \ifx\relax#2\relax\else\ \cs@loop#2\@nil\fi}
\makeatother

\newcommand{\uspace}{\textsc{U-Space}}

\newcommand{\ulens}{\textsc{U-Lens}}

\newcommand{\blrender}[1]{#1}                   %

\newcommand{\msp}{\blrender{MSP}}
\newcommand{\predent}{\blrender{Predictive Entropy}}
\newcommand{\maxent}{\blrender{Maximum Entropy}}
\newcommand{\meannll}{\blrender{Mean NLL}}
\newcommand{\sumnll}{\blrender{Sum NLL}}

\newcommand{\selfcert}{\blrender{Self-Certainty}}
\newcommand{\deepconf}{\blrender{DeepConf}}
\newcommand{\genlen}{\blrender{Generation length}}

\newcommand{\ptrue}{\blrender{P(True)}}
\newcommand{\tokur}{\blrender{TokUR}}
\newcommand{\sement}{\blrender{Semantic Entropy}}
\newcommand{\sentsar}{\blrender{SentenceSAR}}

\newcommand{\featgaps}{\blrender{Feature-Gaps}}

\newcommand{\methodLong}{U-Space: Uncovering When and Why Uncertainty Arises in Language Models}
\newcommand{\jlens}{\textsc{J-Lens}}

\newcommand{\rlens}{\blrender{R-Lens}}

\newcommand{\qwenfull}{Qwen3.5-27B}
\newcommand{\qwen}{Qwen3.5}
\newcommand{\gemmafull}{Gemma-4-31B-it}
\newcommand{\gemma}{Gemma 4}
\newcommand{\magistralfull}{Magistral-Small-2507}
\newcommand{\magistral}{Magistral 1.1}

\newcommand{\gain}{\gamma}                      %

\usepackage{tikz}
\usetikzlibrary{arrows.meta,calc}
\definecolor{unadj}{HTML}{C05B40}
\definecolor{matched}{HTML}{44AA99}
\newcommand{\gl}[1]{\textit{#1}}
\definecolor{genlen}{HTML}{555555}
\newcommand{\dropbar}[3][]{%
  \makebox[28.5mm][l]{\begin{tikzpicture}[baseline=-0.55ex, x=1.1mm, y=1mm, remember picture]
    \path (0,-1.8) (25,1.8);   %
    #1%
    \pgfmathsetmacro{\xr}{#2-48}\pgfmathsetmacro{\xm}{#3-48}
    \pgfmathparse{#2-#3 > 0.9 ? 1 : 0}\ifnum\pgfmathresult=1
      \draw[line width=0.6pt, shorten <=2.35pt, shorten >=2.35pt,
            -{Stealth[length=2.8pt, width=2.5pt]}] (\xr,0) -- (\xm,0);\fi
    \fill[unadj] (\xr,0) circle (2.0pt);
    \fill[matched] (\xm,0) circle (2.0pt);
  \end{tikzpicture}}}
\makeatletter
\newcommand{\dropmark}[2]{%
  \path[overlay] #2 coordinate (dm@pt);   %
  \path[overlay] (current page.south west) coordinate (dm@sw);
  \path[overlay] let \p1=(dm@pt), \p2=(dm@sw) in
    \pgfextra{\protected@write\@auxout{}{%
      \string\expandafter\string\gdef\string\csname\space dropmark@#1\string\endcsname
      {(\the\dimexpr\x1-\x2\relax,\the\dimexpr\y1-\y2\relax)}}};}
\newcommand{\dropmarkpos}[1]{\csname dropmark@#1\endcsname}
\newcommand{\drawGLline}{%
  \ifcsname dropmark@glTop\endcsname\ifcsname dropmark@glBot\endcsname
  \begin{tikzpicture}[remember picture, overlay]
    \draw[genlen, line width=0.5pt, dash pattern=on 0.45pt off 0.55pt]
      ([shift={\dropmarkpos{glTop}}]current page.south west) coordinate (a)
      ([shift={\dropmarkpos{glBot}}]current page.south west) coordinate (b)
      (a) -- (a |- b);
  \end{tikzpicture}\fi\fi}
\makeatother
\newcommand{\dropbarGL}[2]{\dropbar[\dropmark{glTop}{(#1-48,0)}]{#1}{#2}}
\newcommand{\dropbarEnd}[2]{\dropbar[\dropmark{glBot}{(0,-1)}]{#1}{#2}}
\makeatletter
\newcommand{\dropruler}{%
  \makebox[28.5mm][l]{\begin{tikzpicture}[baseline=0pt, x=1.1mm, y=1mm]
    \path (0,0) (25,1);
    \pgfmathsetlengthmacro{\dropruleY}{-(\dp\@arstrutbox+\aboverulesep+\lightrulewidth)}
    \foreach \t in {50,55,60,65,70} {
      \node[anchor=base, inner sep=0pt] at (\t-48,0) {\t};
      \draw[line width=0.4pt, overlay] ([yshift=\dropruleY]\t-48,0) -- ++(0,-2.5pt);}
  \end{tikzpicture}}}
\makeatother

\title{\methodLong}

\author{%
  Tobias Braun$^{1,2,*}$ \quad Nils Loose$^{2,3,*}$ \quad Alexander Herzog$^{2}$ \quad
  Virginia Ceccatelli$^{2,4}$\\[3pt]
  \bfseries Marcus Rohrbach$^{1}$ \quad Thomas Eisenbarth$^{3}$ \quad Lorenzo Cavallaro$^{5,2}$\\[6pt]
  {\small\mdseries $^{1}$Technische Universit\"at Darmstadt \quad
   $^{2}$University College London \quad $^{3}$Universit\"at zu L\"ubeck}\\
  {\small\mdseries $^{4}$Mila -- Quebec Artificial Intelligence Institute \enspace
   $^{5}$Mohamed bin Zayed University of Artificial Intelligence}\\[4pt]
  {\small\mdseries $^{*}$Equal contribution; listed alphabetically.}
}

\begin{document}

\maketitle
\providecommand{\ULensDef}{}
\newcommand{\BlockGemma}{40}
\newcommand{\NTotalGemma}{3{,}964}
\newcommand{\ULensGemma}{0.687}
\newcommand{\AconeGemma}{0.657}
\newcommand{\ULensRawGemma}{0.738}
\newcommand{\AconeRawGemma}{0.725}
\newcommand{\PEGemma}{0.667}
\newcommand{\PEThinkGemma}{0.662}
\newcommand{\MarginPEGemma}{+0.020}
\newcommand{\MarginBlGemma}{+0.013}
\newcommand{\MarginFreeBlGemma}{+0.019}
\newcommand{\BestBlGemma}{0.674}
\newcommand{\BestBlNameGemma}{Feature-Gaps}
\newcommand{\BestFreeBlGemma}{0.668}
\newcommand{\BestFreeBlNameGemma}{TokUR (eu_sum)}
\newcommand{\SULensGemma}{0.687}
\newcommand{\SULensGemmaMmlupro}{0.739}
\newcommand{\SULensGemmaOmnimath}{0.707}
\newcommand{\SULensGemmaSupergpqa}{0.662}
\newcommand{\SULensGemmaTriviaqa}{0.639}
\newcommand{\AuprcULensGemma}{0.436}
\newcommand{\AuprcLmULensGemma}{0.397}
\newcommand{\AurcULensGemma}{0.116}
\newcommand{\AurcLmULensGemma}{0.133}
\newcommand{\AurocULensGemma}{0.738}
\newcommand{\AurocLmULensGemma}{0.687}
\newcommand{\RLoglenULensGemma}{0.30}
\newcommand{\TopFiftyULensGemma}{0.884}
\newcommand{\TopFiftyLmULensGemma}{0.870}
\newcommand{\SAConeGemma}{0.657}
\newcommand{\SAConeGemmaMmlupro}{0.704}
\newcommand{\SAConeGemmaOmnimath}{0.711}
\newcommand{\SAConeGemmaSupergpqa}{0.620}
\newcommand{\SAConeGemmaTriviaqa}{0.594}
\newcommand{\AuprcAConeGemma}{0.438}
\newcommand{\AuprcLmAConeGemma}{0.381}
\newcommand{\AurcAConeGemma}{0.133}
\newcommand{\AurcLmAConeGemma}{0.154}
\newcommand{\AurocAConeGemma}{0.725}
\newcommand{\AurocLmAConeGemma}{0.657}
\newcommand{\RLoglenAConeGemma}{0.45}
\newcommand{\TopFiftyAConeGemma}{0.883}
\newcommand{\TopFiftyLmAConeGemma}{0.853}
\newcommand{\SPEGemma}{0.667}
\newcommand{\SPEGemmaMmlupro}{0.735}
\newcommand{\SPEGemmaOmnimath}{0.663}
\newcommand{\SPEGemmaSupergpqa}{0.647}
\newcommand{\SPEGemmaTriviaqa}{0.624}
\newcommand{\AuprcPEGemma}{0.382}
\newcommand{\AuprcLmPEGemma}{0.373}
\newcommand{\AurcPEGemma}{0.131}
\newcommand{\AurcLmPEGemma}{0.142}
\newcommand{\AurocPEGemma}{0.696}
\newcommand{\AurocLmPEGemma}{0.667}
\newcommand{\RLoglenPEGemma}{0.20}
\newcommand{\TopFiftyPEGemma}{0.869}
\newcommand{\TopFiftyLmPEGemma}{0.859}
\newcommand{\SMeanNLLGemma}{0.657}
\newcommand{\SMeanNLLGemmaMmlupro}{0.737}
\newcommand{\SMeanNLLGemmaOmnimath}{0.655}
\newcommand{\SMeanNLLGemmaSupergpqa}{0.649}
\newcommand{\SMeanNLLGemmaTriviaqa}{0.588}
\newcommand{\AuprcMeanNLLGemma}{0.379}
\newcommand{\AuprcLmMeanNLLGemma}{0.370}
\newcommand{\AurcMeanNLLGemma}{0.132}
\newcommand{\AurcLmMeanNLLGemma}{0.143}
\newcommand{\AurocMeanNLLGemma}{0.692}
\newcommand{\AurocLmMeanNLLGemma}{0.657}
\newcommand{\RLoglenMeanNLLGemma}{0.19}
\newcommand{\TopFiftyMeanNLLGemma}{0.868}
\newcommand{\TopFiftyLmMeanNLLGemma}{0.855}
\newcommand{\SDeepConfGemma}{0.665}
\newcommand{\SDeepConfGemmaMmlupro}{0.727}
\newcommand{\SDeepConfGemmaOmnimath}{0.649}
\newcommand{\SDeepConfGemmaSupergpqa}{0.635}
\newcommand{\SDeepConfGemmaTriviaqa}{0.650}
\newcommand{\AuprcDeepConfGemma}{0.396}
\newcommand{\AuprcLmDeepConfGemma}{0.361}
\newcommand{\AurcDeepConfGemma}{0.126}
\newcommand{\AurcLmDeepConfGemma}{0.142}
\newcommand{\AurocDeepConfGemma}{0.711}
\newcommand{\AurocLmDeepConfGemma}{0.665}
\newcommand{\RLoglenDeepConfGemma}{0.32}
\newcommand{\TopFiftyDeepConfGemma}{0.875}
\newcommand{\TopFiftyLmDeepConfGemma}{0.858}
\newcommand{\SSumNLLGemma}{0.664}
\newcommand{\SSumNLLGemmaMmlupro}{0.757}
\newcommand{\SSumNLLGemmaOmnimath}{0.661}
\newcommand{\SSumNLLGemmaSupergpqa}{0.643}
\newcommand{\SSumNLLGemmaTriviaqa}{0.595}
\newcommand{\AuprcSumNLLGemma}{0.470}
\newcommand{\AuprcLmSumNLLGemma}{0.375}
\newcommand{\AurcSumNLLGemma}{0.128}
\newcommand{\AurcLmSumNLLGemma}{0.142}
\newcommand{\AurocSumNLLGemma}{0.750}
\newcommand{\AurocLmSumNLLGemma}{0.664}
\newcommand{\RLoglenSumNLLGemma}{0.79}
\newcommand{\TopFiftySumNLLGemma}{0.893}
\newcommand{\TopFiftyLmSumNLLGemma}{0.861}
\newcommand{\SEDISGemma}{0.653}
\newcommand{\SEDISGemmaMmlupro}{0.753}
\newcommand{\SEDISGemmaOmnimath}{0.647}
\newcommand{\SEDISGemmaSupergpqa}{0.625}
\newcommand{\SEDISGemmaTriviaqa}{0.588}
\newcommand{\AuprcEDISGemma}{0.463}
\newcommand{\AuprcLmEDISGemma}{0.368}
\newcommand{\AurcEDISGemma}{0.133}
\newcommand{\AurcLmEDISGemma}{0.148}
\newcommand{\AurocEDISGemma}{0.742}
\newcommand{\AurocLmEDISGemma}{0.653}
\newcommand{\RLoglenEDISGemma}{0.83}
\newcommand{\TopFiftyEDISGemma}{0.891}
\newcommand{\TopFiftyLmEDISGemma}{0.859}
\newcommand{\SSelfCertaintyGemma}{0.566}
\newcommand{\SSelfCertaintyGemmaMmlupro}{0.439}
\newcommand{\SSelfCertaintyGemmaOmnimath}{0.646}
\newcommand{\SSelfCertaintyGemmaSupergpqa}{0.582}
\newcommand{\SSelfCertaintyGemmaTriviaqa}{0.598}
\newcommand{\AuprcSelfCertaintyGemma}{0.395}
\newcommand{\AuprcLmSelfCertaintyGemma}{0.313}
\newcommand{\AurcSelfCertaintyGemma}{0.141}
\newcommand{\AurcLmSelfCertaintyGemma}{0.175}
\newcommand{\AurocSelfCertaintyGemma}{0.681}
\newcommand{\AurocLmSelfCertaintyGemma}{0.566}
\newcommand{\RLoglenSelfCertaintyGemma}{0.67}
\newcommand{\TopFiftySelfCertaintyGemma}{0.863}
\newcommand{\TopFiftyLmSelfCertaintyGemma}{0.823}
\newcommand{\SMSPGemma}{0.490}
\newcommand{\SMSPGemmaMmlupro}{0.493}
\newcommand{\SMSPGemmaOmnimath}{0.492}
\newcommand{\SMSPGemmaSupergpqa}{0.476}
\newcommand{\SMSPGemmaTriviaqa}{0.500}
\newcommand{\AuprcMSPGemma}{0.215}
\newcommand{\AuprcLmMSPGemma}{0.241}
\newcommand{\AurcMSPGemma}{0.221}
\newcommand{\AurcLmMSPGemma}{0.220}
\newcommand{\AurocMSPGemma}{0.497}
\newcommand{\AurocLmMSPGemma}{0.490}
\newcommand{\RLoglenMSPGemma}{0.05}
\newcommand{\TopFiftyMSPGemma}{0.791}
\newcommand{\TopFiftyLmMSPGemma}{0.784}
\newcommand{\SMaxEntropyGemma}{0.598}
\newcommand{\SMaxEntropyGemmaMmlupro}{0.617}
\newcommand{\SMaxEntropyGemmaOmnimath}{0.583}
\newcommand{\SMaxEntropyGemmaSupergpqa}{0.562}
\newcommand{\SMaxEntropyGemmaTriviaqa}{0.629}
\newcommand{\AuprcMaxEntropyGemma}{0.358}
\newcommand{\AuprcLmMaxEntropyGemma}{0.318}
\newcommand{\AurcMaxEntropyGemma}{0.153}
\newcommand{\AurcLmMaxEntropyGemma}{0.170}
\newcommand{\AurocMaxEntropyGemma}{0.654}
\newcommand{\AurocLmMaxEntropyGemma}{0.598}
\newcommand{\RLoglenMaxEntropyGemma}{0.36}
\newcommand{\TopFiftyMaxEntropyGemma}{0.848}
\newcommand{\TopFiftyLmMaxEntropyGemma}{0.826}
\newcommand{\STokURGemma}{0.668}
\newcommand{\ExtTokURGemma}{1}
\newcommand{\STokURGemmaMmlupro}{0.733}
\newcommand{\STokURGemmaOmnimath}{0.683}
\newcommand{\STokURGemmaSupergpqa}{0.546}
\newcommand{\STokURGemmaTriviaqa}{0.712}
\newcommand{\AuprcTokURGemma}{0.488}
\newcommand{\AuprcLmTokURGemma}{0.369}
\newcommand{\AurcTokURGemma}{0.128}
\newcommand{\AurcLmTokURGemma}{0.152}
\newcommand{\AurocTokURGemma}{0.756}
\newcommand{\AurocLmTokURGemma}{0.668}
\newcommand{\RLoglenTokURGemma}{0.79}
\newcommand{\TopFiftyTokURGemma}{0.886}
\newcommand{\TopFiftyLmTokURGemma}{0.845}
\newcommand{\SFeatureGapsGemma}{0.674}
\newcommand{\ExtFeatureGapsGemma}{1}
\newcommand{\SFeatureGapsGemmaMmlupro}{0.714}
\newcommand{\SFeatureGapsGemmaOmnimath}{0.591}
\newcommand{\SFeatureGapsGemmaSupergpqa}{0.682}
\newcommand{\SFeatureGapsGemmaTriviaqa}{0.708}
\newcommand{\AuprcFeatureGapsGemma}{0.421}
\newcommand{\AuprcLmFeatureGapsGemma}{0.372}
\newcommand{\AurcFeatureGapsGemma}{0.114}
\newcommand{\AurcLmFeatureGapsGemma}{0.138}
\newcommand{\AurocFeatureGapsGemma}{0.733}
\newcommand{\AurocLmFeatureGapsGemma}{0.674}
\newcommand{\RLoglenFeatureGapsGemma}{0.45}
\newcommand{\TopFiftyFeatureGapsGemma}{0.887}
\newcommand{\TopFiftyLmFeatureGapsGemma}{0.864}
\newcommand{\SGenLengthGemma}{0.540}
\newcommand{\SGenLengthGemmaMmlupro}{0.610}
\newcommand{\SGenLengthGemmaOmnimath}{0.540}
\newcommand{\SGenLengthGemmaSupergpqa}{0.485}
\newcommand{\SGenLengthGemmaTriviaqa}{0.524}
\newcommand{\AuprcGenLengthGemma}{0.399}
\newcommand{\AuprcLmGenLengthGemma}{0.267}
\newcommand{\AurcGenLengthGemma}{0.148}
\newcommand{\AurcLmGenLengthGemma}{0.200}
\newcommand{\AurocGenLengthGemma}{0.696}
\newcommand{\AurocLmGenLengthGemma}{0.540}
\newcommand{\RLoglenGenLengthGemma}{0.91}
\newcommand{\TopFiftyGenLengthGemma}{0.868}
\newcommand{\TopFiftyLmGenLengthGemma}{0.800}
\newcommand{\SPEThinkGemma}{0.662}
\newcommand{\SPEThinkGemmaMmlupro}{0.741}
\newcommand{\SPEThinkGemmaOmnimath}{0.643}
\newcommand{\SPEThinkGemmaSupergpqa}{0.642}
\newcommand{\SPEThinkGemmaTriviaqa}{0.623}
\newcommand{\AuprcPEThinkGemma}{0.367}
\newcommand{\AuprcLmPEThinkGemma}{0.367}
\newcommand{\AurcPEThinkGemma}{0.137}
\newcommand{\AurcLmPEThinkGemma}{0.146}
\newcommand{\AurocPEThinkGemma}{0.679}
\newcommand{\AurocLmPEThinkGemma}{0.662}
\newcommand{\RLoglenPEThinkGemma}{0.15}
\newcommand{\TopFiftyPEThinkGemma}{0.859}
\newcommand{\TopFiftyLmPEThinkGemma}{0.853}
\newcommand{\SPTrueGemma}{0.531}
\newcommand{\ExtPTrueGemma}{1}
\newcommand{\SPTrueGemmaMmlupro}{0.582}
\newcommand{\SPTrueGemmaOmnimath}{0.510}
\newcommand{\SPTrueGemmaSupergpqa}{0.487}
\newcommand{\SPTrueGemmaTriviaqa}{0.547}
\newcommand{\AuprcPTrueGemma}{0.224}
\newcommand{\AuprcLmPTrueGemma}{0.250}
\newcommand{\AurcPTrueGemma}{0.199}
\newcommand{\AurcLmPTrueGemma}{0.203}
\newcommand{\AurocPTrueGemma}{0.533}
\newcommand{\AurocLmPTrueGemma}{0.531}
\newcommand{\RLoglenPTrueGemma}{0.04}
\newcommand{\TopFiftyPTrueGemma}{0.801}
\newcommand{\TopFiftyLmPTrueGemma}{0.799}
\newcommand{\GainAuprcGemma}{+0.054}
\newcommand{\GainAuprcLmGemma}{+0.024}
\newcommand{\GainAurcGemma}{+0.014}
\newcommand{\GainAurcLmGemma}{+0.009}
\newcommand{\GainAurocGemma}{+0.042}
\newcommand{\GainAurocLmGemma}{+0.020}
\newcommand{\GainTopFiftyGemma}{+0.015}
\newcommand{\GainTopFiftyLmGemma}{+0.011}
\newcommand{\NGemmaMmlupro}{1000}
\newcommand{\ErrGemmaMmlupro}{10.9}
\newcommand{\NGemmaOmnimath}{1000}
\newcommand{\ErrGemmaOmnimath}{19.4}
\newcommand{\NGemmaSupergpqa}{999}
\newcommand{\ErrGemmaSupergpqa}{36.6}
\newcommand{\NGemmaTriviaqa}{965}
\newcommand{\ErrGemmaTriviaqa}{17.9}
\newcommand{\BlockQwen}{42}
\newcommand{\NTotalQwen}{3{,}880}
\newcommand{\ULensQwen}{0.717}
\newcommand{\AconeQwen}{0.676}
\newcommand{\ULensRawQwen}{0.737}
\newcommand{\AconeRawQwen}{0.710}
\newcommand{\PEQwen}{0.699}
\newcommand{\PEThinkQwen}{0.699}
\newcommand{\MarginPEQwen}{+0.017}
\newcommand{\MarginBlQwen}{+0.012}
\newcommand{\MarginFreeBlQwen}{+0.012}
\newcommand{\BestBlQwen}{0.704}
\newcommand{\BestBlNameQwen}{Max entropy}
\newcommand{\BestFreeBlQwen}{0.704}
\newcommand{\BestFreeBlNameQwen}{Max entropy}
\newcommand{\SULensQwen}{0.717}
\newcommand{\SULensQwenMmlupro}{0.781}
\newcommand{\SULensQwenOmnimath}{0.679}
\newcommand{\SULensQwenSupergpqa}{0.683}
\newcommand{\SULensQwenTriviaqa}{0.724}
\newcommand{\AuprcULensQwen}{0.427}
\newcommand{\AuprcLmULensQwen}{0.416}
\newcommand{\AurcULensQwen}{0.112}
\newcommand{\AurcLmULensQwen}{0.121}
\newcommand{\AurocULensQwen}{0.737}
\newcommand{\AurocLmULensQwen}{0.717}
\newcommand{\RLoglenULensQwen}{0.37}
\newcommand{\TopFiftyULensQwen}{0.891}
\newcommand{\TopFiftyLmULensQwen}{0.884}
\newcommand{\SAConeQwen}{0.676}
\newcommand{\SAConeQwenMmlupro}{0.725}
\newcommand{\SAConeQwenOmnimath}{0.707}
\newcommand{\SAConeQwenSupergpqa}{0.663}
\newcommand{\SAConeQwenTriviaqa}{0.607}
\newcommand{\AuprcAConeQwen}{0.400}
\newcommand{\AuprcLmAConeQwen}{0.381}
\newcommand{\AurcAConeQwen}{0.117}
\newcommand{\AurcLmAConeQwen}{0.135}
\newcommand{\AurocAConeQwen}{0.710}
\newcommand{\AurocLmAConeQwen}{0.676}
\newcommand{\RLoglenAConeQwen}{0.29}
\newcommand{\TopFiftyAConeQwen}{0.881}
\newcommand{\TopFiftyLmAConeQwen}{0.864}
\newcommand{\SPEQwen}{0.699}
\newcommand{\SPEQwenMmlupro}{0.790}
\newcommand{\SPEQwenOmnimath}{0.623}
\newcommand{\SPEQwenSupergpqa}{0.657}
\newcommand{\SPEQwenTriviaqa}{0.728}
\newcommand{\AuprcPEQwen}{0.405}
\newcommand{\AuprcLmPEQwen}{0.392}
\newcommand{\AurcPEQwen}{0.120}
\newcommand{\AurcLmPEQwen}{0.128}
\newcommand{\AurocPEQwen}{0.723}
\newcommand{\AurocLmPEQwen}{0.699}
\newcommand{\RLoglenPEQwen}{0.45}
\newcommand{\TopFiftyPEQwen}{0.885}
\newcommand{\TopFiftyLmPEQwen}{0.880}
\newcommand{\SMeanNLLQwen}{0.691}
\newcommand{\SMeanNLLQwenMmlupro}{0.784}
\newcommand{\SMeanNLLQwenOmnimath}{0.623}
\newcommand{\SMeanNLLQwenSupergpqa}{0.656}
\newcommand{\SMeanNLLQwenTriviaqa}{0.700}
\newcommand{\AuprcMeanNLLQwen}{0.403}
\newcommand{\AuprcLmMeanNLLQwen}{0.384}
\newcommand{\AurcMeanNLLQwen}{0.122}
\newcommand{\AurcLmMeanNLLQwen}{0.130}
\newcommand{\AurocMeanNLLQwen}{0.719}
\newcommand{\AurocLmMeanNLLQwen}{0.691}
\newcommand{\RLoglenMeanNLLQwen}{0.46}
\newcommand{\TopFiftyMeanNLLQwen}{0.885}
\newcommand{\TopFiftyLmMeanNLLQwen}{0.879}
\newcommand{\SDeepConfQwen}{0.696}
\newcommand{\SDeepConfQwenMmlupro}{0.777}
\newcommand{\SDeepConfQwenOmnimath}{0.623}
\newcommand{\SDeepConfQwenSupergpqa}{0.661}
\newcommand{\SDeepConfQwenTriviaqa}{0.725}
\newcommand{\AuprcDeepConfQwen}{0.382}
\newcommand{\AuprcLmDeepConfQwen}{0.379}
\newcommand{\AurcDeepConfQwen}{0.117}
\newcommand{\AurcLmDeepConfQwen}{0.125}
\newcommand{\AurocDeepConfQwen}{0.719}
\newcommand{\AurocLmDeepConfQwen}{0.696}
\newcommand{\RLoglenDeepConfQwen}{0.37}
\newcommand{\TopFiftyDeepConfQwen}{0.888}
\newcommand{\TopFiftyLmDeepConfQwen}{0.879}
\newcommand{\SSumNLLQwen}{0.677}
\newcommand{\SSumNLLQwenMmlupro}{0.771}
\newcommand{\SSumNLLQwenOmnimath}{0.615}
\newcommand{\SSumNLLQwenSupergpqa}{0.636}
\newcommand{\SSumNLLQwenTriviaqa}{0.686}
\newcommand{\AuprcSumNLLQwen}{0.395}
\newcommand{\AuprcLmSumNLLQwen}{0.378}
\newcommand{\AurcSumNLLQwen}{0.133}
\newcommand{\AurcLmSumNLLQwen}{0.134}
\newcommand{\AurocSumNLLQwen}{0.714}
\newcommand{\AurocLmSumNLLQwen}{0.677}
\newcommand{\RLoglenSumNLLQwen}{0.86}
\newcommand{\TopFiftySumNLLQwen}{0.880}
\newcommand{\TopFiftyLmSumNLLQwen}{0.869}
\newcommand{\SEDISQwen}{0.659}
\newcommand{\SEDISQwenMmlupro}{0.759}
\newcommand{\SEDISQwenOmnimath}{0.597}
\newcommand{\SEDISQwenSupergpqa}{0.632}
\newcommand{\SEDISQwenTriviaqa}{0.648}
\newcommand{\AuprcEDISQwen}{0.384}
\newcommand{\AuprcLmEDISQwen}{0.358}
\newcommand{\AurcEDISQwen}{0.138}
\newcommand{\AurcLmEDISQwen}{0.139}
\newcommand{\AurocEDISQwen}{0.705}
\newcommand{\AurocLmEDISQwen}{0.659}
\newcommand{\RLoglenEDISQwen}{0.88}
\newcommand{\TopFiftyEDISQwen}{0.875}
\newcommand{\TopFiftyLmEDISQwen}{0.867}
\newcommand{\SSelfCertaintyQwen}{0.694}
\newcommand{\SSelfCertaintyQwenMmlupro}{0.782}
\newcommand{\SSelfCertaintyQwenOmnimath}{0.645}
\newcommand{\SSelfCertaintyQwenSupergpqa}{0.624}
\newcommand{\SSelfCertaintyQwenTriviaqa}{0.725}
\newcommand{\AuprcSelfCertaintyQwen}{0.375}
\newcommand{\AuprcLmSelfCertaintyQwen}{0.369}
\newcommand{\AurcSelfCertaintyQwen}{0.120}
\newcommand{\AurcLmSelfCertaintyQwen}{0.129}
\newcommand{\AurocSelfCertaintyQwen}{0.718}
\newcommand{\AurocLmSelfCertaintyQwen}{0.694}
\newcommand{\RLoglenSelfCertaintyQwen}{0.34}
\newcommand{\TopFiftySelfCertaintyQwen}{0.887}
\newcommand{\TopFiftyLmSelfCertaintyQwen}{0.871}
\newcommand{\SMSPQwen}{0.487}
\newcommand{\SMSPQwenMmlupro}{0.533}
\newcommand{\SMSPQwenOmnimath}{0.347}
\newcommand{\SMSPQwenSupergpqa}{0.531}
\newcommand{\SMSPQwenTriviaqa}{0.540}
\newcommand{\AuprcMSPQwen}{0.218}
\newcommand{\AuprcLmMSPQwen}{0.239}
\newcommand{\AurcMSPQwen}{0.197}
\newcommand{\AurcLmMSPQwen}{0.205}
\newcommand{\AurocMSPQwen}{0.515}
\newcommand{\AurocLmMSPQwen}{0.487}
\newcommand{\RLoglenMSPQwen}{0.10}
\newcommand{\TopFiftyMSPQwen}{0.807}
\newcommand{\TopFiftyLmMSPQwen}{0.794}
\newcommand{\SMaxEntropyQwen}{0.704}
\newcommand{\SMaxEntropyQwenMmlupro}{0.819}
\newcommand{\SMaxEntropyQwenOmnimath}{0.704}
\newcommand{\SMaxEntropyQwenSupergpqa}{0.566}
\newcommand{\SMaxEntropyQwenTriviaqa}{0.729}
\newcommand{\AuprcMaxEntropyQwen}{0.376}
\newcommand{\AuprcLmMaxEntropyQwen}{0.383}
\newcommand{\AurcMaxEntropyQwen}{0.126}
\newcommand{\AurcLmMaxEntropyQwen}{0.138}
\newcommand{\AurocMaxEntropyQwen}{0.724}
\newcommand{\AurocLmMaxEntropyQwen}{0.704}
\newcommand{\RLoglenMaxEntropyQwen}{0.42}
\newcommand{\TopFiftyMaxEntropyQwen}{0.892}
\newcommand{\TopFiftyLmMaxEntropyQwen}{0.862}
\newcommand{\STokURQwen}{0.616}
\newcommand{\ExtTokURQwen}{1}
\newcommand{\STokURQwenMmlupro}{0.725}
\newcommand{\STokURQwenOmnimath}{0.540}
\newcommand{\STokURQwenSupergpqa}{0.584}
\newcommand{\STokURQwenTriviaqa}{0.613}
\newcommand{\AuprcTokURQwen}{0.344}
\newcommand{\AuprcLmTokURQwen}{0.345}
\newcommand{\AurcTokURQwen}{0.147}
\newcommand{\AurcLmTokURQwen}{0.152}
\newcommand{\AurocTokURQwen}{0.652}
\newcommand{\AurocLmTokURQwen}{0.616}
\newcommand{\RLoglenTokURQwen}{0.80}
\newcommand{\TopFiftyTokURQwen}{0.864}
\newcommand{\TopFiftyLmTokURQwen}{0.845}
\newcommand{\SFeatureGapsQwen}{0.618}
\newcommand{\ExtFeatureGapsQwen}{1}
\newcommand{\SFeatureGapsQwenMmlupro}{0.726}
\newcommand{\SFeatureGapsQwenOmnimath}{0.530}
\newcommand{\SFeatureGapsQwenSupergpqa}{0.491}
\newcommand{\SFeatureGapsQwenTriviaqa}{0.728}
\newcommand{\AuprcFeatureGapsQwen}{0.376}
\newcommand{\AuprcLmFeatureGapsQwen}{0.359}
\newcommand{\AurcFeatureGapsQwen}{0.146}
\newcommand{\AurcLmFeatureGapsQwen}{0.158}
\newcommand{\AurocFeatureGapsQwen}{0.667}
\newcommand{\AurocLmFeatureGapsQwen}{0.618}
\newcommand{\RLoglenFeatureGapsQwen}{0.46}
\newcommand{\TopFiftyFeatureGapsQwen}{0.858}
\newcommand{\TopFiftyLmFeatureGapsQwen}{0.850}
\newcommand{\SGenLengthQwen}{0.518}
\newcommand{\SGenLengthQwenMmlupro}{0.519}
\newcommand{\SGenLengthQwenOmnimath}{0.539}
\newcommand{\SGenLengthQwenSupergpqa}{0.480}
\newcommand{\SGenLengthQwenTriviaqa}{0.533}
\newcommand{\AuprcGenLengthQwen}{0.346}
\newcommand{\AuprcLmGenLengthQwen}{0.245}
\newcommand{\AurcGenLengthQwen}{0.153}
\newcommand{\AurcLmGenLengthQwen}{0.195}
\newcommand{\AurocGenLengthQwen}{0.663}
\newcommand{\AurocLmGenLengthQwen}{0.518}
\newcommand{\RLoglenGenLengthQwen}{0.91}
\newcommand{\TopFiftyGenLengthQwen}{0.859}
\newcommand{\TopFiftyLmGenLengthQwen}{0.804}
\newcommand{\SPEThinkQwen}{0.699}
\newcommand{\SPEThinkQwenMmlupro}{0.792}
\newcommand{\SPEThinkQwenOmnimath}{0.619}
\newcommand{\SPEThinkQwenSupergpqa}{0.654}
\newcommand{\SPEThinkQwenTriviaqa}{0.730}
\newcommand{\AuprcPEThinkQwen}{0.400}
\newcommand{\AuprcLmPEThinkQwen}{0.392}
\newcommand{\AurcPEThinkQwen}{0.119}
\newcommand{\AurcLmPEThinkQwen}{0.127}
\newcommand{\AurocPEThinkQwen}{0.721}
\newcommand{\AurocLmPEThinkQwen}{0.699}
\newcommand{\RLoglenPEThinkQwen}{0.42}
\newcommand{\TopFiftyPEThinkQwen}{0.885}
\newcommand{\TopFiftyLmPEThinkQwen}{0.878}
\newcommand{\SPTrueQwen}{0.618}
\newcommand{\ExtPTrueQwen}{1}
\newcommand{\SPTrueQwenMmlupro}{0.606}
\newcommand{\SPTrueQwenOmnimath}{0.585}
\newcommand{\SPTrueQwenSupergpqa}{0.517}
\newcommand{\SPTrueQwenTriviaqa}{0.763}
\newcommand{\AuprcPTrueQwen}{0.343}
\newcommand{\AuprcLmPTrueQwen}{0.346}
\newcommand{\AurcPTrueQwen}{0.158}
\newcommand{\AurcLmPTrueQwen}{0.169}
\newcommand{\AurocPTrueQwen}{0.635}
\newcommand{\AurocLmPTrueQwen}{0.618}
\newcommand{\RLoglenPTrueQwen}{0.18}
\newcommand{\TopFiftyPTrueQwen}{0.846}
\newcommand{\TopFiftyLmPTrueQwen}{0.838}
\newcommand{\GainAuprcQwen}{+0.022}
\newcommand{\GainAuprcLmQwen}{+0.024}
\newcommand{\GainAurcQwen}{+0.008}
\newcommand{\GainAurcLmQwen}{+0.006}
\newcommand{\GainAurocQwen}{+0.014}
\newcommand{\GainAurocLmQwen}{+0.017}
\newcommand{\GainTopFiftyQwen}{+0.005}
\newcommand{\GainTopFiftyLmQwen}{+0.004}
\newcommand{\NQwenMmlupro}{989}
\newcommand{\ErrQwenMmlupro}{12.6}
\newcommand{\NQwenOmnimath}{934}
\newcommand{\ErrQwenOmnimath}{12.3}
\newcommand{\NQwenSupergpqa}{984}
\newcommand{\ErrQwenSupergpqa}{36.1}
\newcommand{\NQwenTriviaqa}{973}
\newcommand{\ErrQwenTriviaqa}{19.9}

\newcommand{\BlockMagistral}{27}
\newcommand{\NTotalMagistral}{3{,}808}
\newcommand{\ULensMagistral}{0.677}
\newcommand{\AconeMagistral}{0.670}
\newcommand{\ULensRawMagistral}{0.649}
\newcommand{\AconeRawMagistral}{0.678}
\newcommand{\PEMagistral}{0.628}
\newcommand{\PEThinkMagistral}{0.630}
\newcommand{\MarginPEMagistral}{+0.049}
\newcommand{\MarginBlMagistral}{+0.045}
\newcommand{\MarginFreeBlMagistral}{+0.045}
\newcommand{\BestBlMagistral}{0.631}
\newcommand{\BestBlNameMagistral}{DeepConf}
\newcommand{\BestFreeBlMagistral}{0.631}
\newcommand{\BestFreeBlNameMagistral}{DeepConf}
\newcommand{\SULensMagistral}{0.677}
\newcommand{\SULensMagistralMmlupro}{0.674}
\newcommand{\SULensMagistralOmnimath}{0.732}
\newcommand{\SULensMagistralSupergpqa}{0.688}
\newcommand{\SULensMagistralTriviaqa}{0.612}
\newcommand{\AuprcULensMagistral}{0.508}
\newcommand{\AuprcLmULensMagistral}{0.539}
\newcommand{\AurcULensMagistral}{0.267}
\newcommand{\AurcLmULensMagistral}{0.250}
\newcommand{\AurocULensMagistral}{0.649}
\newcommand{\AurocLmULensMagistral}{0.677}
\newcommand{\RLoglenULensMagistral}{-0.04}
\newcommand{\TopFiftyULensMagistral}{0.757}
\newcommand{\TopFiftyLmULensMagistral}{0.746}
\newcommand{\SAConeMagistral}{0.670}
\newcommand{\SAConeMagistralMmlupro}{0.716}
\newcommand{\SAConeMagistralOmnimath}{0.691}
\newcommand{\SAConeMagistralSupergpqa}{0.683}
\newcommand{\SAConeMagistralTriviaqa}{0.591}
\newcommand{\AuprcAConeMagistral}{0.520}
\newcommand{\AuprcLmAConeMagistral}{0.531}
\newcommand{\AurcAConeMagistral}{0.240}
\newcommand{\AurcLmAConeMagistral}{0.252}
\newcommand{\AurocAConeMagistral}{0.678}
\newcommand{\AurocLmAConeMagistral}{0.670}
\newcommand{\RLoglenAConeMagistral}{0.09}
\newcommand{\TopFiftyAConeMagistral}{0.748}
\newcommand{\TopFiftyLmAConeMagistral}{0.743}
\newcommand{\SPEMagistral}{0.628}
\newcommand{\SPEMagistralMmlupro}{0.584}
\newcommand{\SPEMagistralOmnimath}{0.714}
\newcommand{\SPEMagistralSupergpqa}{0.652}
\newcommand{\SPEMagistralTriviaqa}{0.560}
\newcommand{\AuprcPEMagistral}{0.475}
\newcommand{\AuprcLmPEMagistral}{0.497}
\newcommand{\AurcPEMagistral}{0.298}
\newcommand{\AurcLmPEMagistral}{0.283}
\newcommand{\AurocPEMagistral}{0.609}
\newcommand{\AurocLmPEMagistral}{0.628}
\newcommand{\RLoglenPEMagistral}{-0.06}
\newcommand{\TopFiftyPEMagistral}{0.733}
\newcommand{\TopFiftyLmPEMagistral}{0.715}
\newcommand{\SMeanNLLMagistral}{0.623}
\newcommand{\SMeanNLLMagistralMmlupro}{0.583}
\newcommand{\SMeanNLLMagistralOmnimath}{0.712}
\newcommand{\SMeanNLLMagistralSupergpqa}{0.650}
\newcommand{\SMeanNLLMagistralTriviaqa}{0.545}
\newcommand{\AuprcMeanNLLMagistral}{0.468}
\newcommand{\AuprcLmMeanNLLMagistral}{0.495}
\newcommand{\AurcMeanNLLMagistral}{0.300}
\newcommand{\AurcLmMeanNLLMagistral}{0.285}
\newcommand{\AurocMeanNLLMagistral}{0.605}
\newcommand{\AurocLmMeanNLLMagistral}{0.623}
\newcommand{\RLoglenMeanNLLMagistral}{-0.05}
\newcommand{\TopFiftyMeanNLLMagistral}{0.729}
\newcommand{\TopFiftyLmMeanNLLMagistral}{0.712}
\newcommand{\SDeepConfMagistral}{0.631}
\newcommand{\SDeepConfMagistralMmlupro}{0.607}
\newcommand{\SDeepConfMagistralOmnimath}{0.730}
\newcommand{\SDeepConfMagistralSupergpqa}{0.662}
\newcommand{\SDeepConfMagistralTriviaqa}{0.527}
\newcommand{\AuprcDeepConfMagistral}{0.473}
\newcommand{\AuprcLmDeepConfMagistral}{0.501}
\newcommand{\AurcDeepConfMagistral}{0.291}
\newcommand{\AurcLmDeepConfMagistral}{0.274}
\newcommand{\AurocDeepConfMagistral}{0.605}
\newcommand{\AurocLmDeepConfMagistral}{0.631}
\newcommand{\RLoglenDeepConfMagistral}{-0.13}
\newcommand{\TopFiftyDeepConfMagistral}{0.732}
\newcommand{\TopFiftyLmDeepConfMagistral}{0.718}
\newcommand{\SSumNLLMagistral}{0.620}
\newcommand{\SSumNLLMagistralMmlupro}{0.583}
\newcommand{\SSumNLLMagistralOmnimath}{0.709}
\newcommand{\SSumNLLMagistralSupergpqa}{0.635}
\newcommand{\SSumNLLMagistralTriviaqa}{0.551}
\newcommand{\AuprcSumNLLMagistral}{0.534}
\newcommand{\AuprcLmSumNLLMagistral}{0.488}
\newcommand{\AurcSumNLLMagistral}{0.233}
\newcommand{\AurcLmSumNLLMagistral}{0.287}
\newcommand{\AurocSumNLLMagistral}{0.702}
\newcommand{\AurocLmSumNLLMagistral}{0.620}
\newcommand{\RLoglenSumNLLMagistral}{0.73}
\newcommand{\TopFiftySumNLLMagistral}{0.783}
\newcommand{\TopFiftyLmSumNLLMagistral}{0.716}
\newcommand{\SEDISMagistral}{0.618}
\newcommand{\SEDISMagistralMmlupro}{0.598}
\newcommand{\SEDISMagistralOmnimath}{0.698}
\newcommand{\SEDISMagistralSupergpqa}{0.628}
\newcommand{\SEDISMagistralTriviaqa}{0.547}
\newcommand{\AuprcEDISMagistral}{0.533}
\newcommand{\AuprcLmEDISMagistral}{0.486}
\newcommand{\AurcEDISMagistral}{0.234}
\newcommand{\AurcLmEDISMagistral}{0.288}
\newcommand{\AurocEDISMagistral}{0.703}
\newcommand{\AurocLmEDISMagistral}{0.618}
\newcommand{\RLoglenEDISMagistral}{0.79}
\newcommand{\TopFiftyEDISMagistral}{0.781}
\newcommand{\TopFiftyLmEDISMagistral}{0.714}
\newcommand{\SSelfCertaintyMagistral}{0.617}
\newcommand{\SSelfCertaintyMagistralMmlupro}{0.566}
\newcommand{\SSelfCertaintyMagistralOmnimath}{0.733}
\newcommand{\SSelfCertaintyMagistralSupergpqa}{0.648}
\newcommand{\SSelfCertaintyMagistralTriviaqa}{0.523}
\newcommand{\AuprcSelfCertaintyMagistral}{0.476}
\newcommand{\AuprcLmSelfCertaintyMagistral}{0.487}
\newcommand{\AurcSelfCertaintyMagistral}{0.296}
\newcommand{\AurcLmSelfCertaintyMagistral}{0.286}
\newcommand{\AurocSelfCertaintyMagistral}{0.604}
\newcommand{\AurocLmSelfCertaintyMagistral}{0.617}
\newcommand{\RLoglenSelfCertaintyMagistral}{-0.05}
\newcommand{\TopFiftySelfCertaintyMagistral}{0.727}
\newcommand{\TopFiftyLmSelfCertaintyMagistral}{0.712}
\newcommand{\SMSPMagistral}{0.582}
\newcommand{\SMSPMagistralMmlupro}{0.613}
\newcommand{\SMSPMagistralOmnimath}{0.622}
\newcommand{\SMSPMagistralSupergpqa}{0.606}
\newcommand{\SMSPMagistralTriviaqa}{0.486}
\newcommand{\AuprcMSPMagistral}{0.406}
\newcommand{\AuprcLmMSPMagistral}{0.429}
\newcommand{\AurcMSPMagistral}{0.292}
\newcommand{\AurcLmMSPMagistral}{0.299}
\newcommand{\AurocMSPMagistral}{0.580}
\newcommand{\AurocLmMSPMagistral}{0.582}
\newcommand{\RLoglenMSPMagistral}{-0.02}
\newcommand{\TopFiftyMSPMagistral}{0.705}
\newcommand{\TopFiftyLmMSPMagistral}{0.694}
\newcommand{\SMaxEntropyMagistral}{0.613}
\newcommand{\SMaxEntropyMagistralMmlupro}{0.582}
\newcommand{\SMaxEntropyMagistralOmnimath}{0.645}
\newcommand{\SMaxEntropyMagistralSupergpqa}{0.628}
\newcommand{\SMaxEntropyMagistralTriviaqa}{0.596}
\newcommand{\AuprcMaxEntropyMagistral}{0.509}
\newcommand{\AuprcLmMaxEntropyMagistral}{0.476}
\newcommand{\AurcMaxEntropyMagistral}{0.244}
\newcommand{\AurcLmMaxEntropyMagistral}{0.287}
\newcommand{\AurocMaxEntropyMagistral}{0.675}
\newcommand{\AurocLmMaxEntropyMagistral}{0.613}
\newcommand{\RLoglenMaxEntropyMagistral}{0.26}
\newcommand{\TopFiftyMaxEntropyMagistral}{0.760}
\newcommand{\TopFiftyLmMaxEntropyMagistral}{0.715}
\newcommand{\STokURMagistral}{0.528}
\newcommand{\ExtTokURMagistral}{1}
\newcommand{\STokURMagistralMmlupro}{0.552}
\newcommand{\STokURMagistralOmnimath}{0.566}
\newcommand{\STokURMagistralSupergpqa}{0.523}
\newcommand{\STokURMagistralTriviaqa}{0.473}
\newcommand{\AuprcTokURMagistral}{0.399}
\newcommand{\AuprcLmTokURMagistral}{0.354}
\newcommand{\AurcTokURMagistral}{0.220}
\newcommand{\AurcLmTokURMagistral}{0.260}
\newcommand{\AurocTokURMagistral}{0.643}
\newcommand{\AurocLmTokURMagistral}{0.528}
\newcommand{\RLoglenTokURMagistral}{0.90}
\newcommand{\TopFiftyTokURMagistral}{0.792}
\newcommand{\TopFiftyLmTokURMagistral}{0.726}
\newcommand{\SFeatureGapsMagistral}{0.583}
\newcommand{\ExtFeatureGapsMagistral}{1}
\newcommand{\SFeatureGapsMagistralMmlupro}{0.665}
\newcommand{\SFeatureGapsMagistralSupergpqa}{0.531}
\newcommand{\SFeatureGapsMagistralTriviaqa}{0.553}
\newcommand{\AuprcFeatureGapsMagistral}{0.447}
\newcommand{\AuprcLmFeatureGapsMagistral}{0.452}
\newcommand{\AurcFeatureGapsMagistral}{0.273}
\newcommand{\AurcLmFeatureGapsMagistral}{0.295}
\newcommand{\AurocFeatureGapsMagistral}{0.628}
\newcommand{\AurocLmFeatureGapsMagistral}{0.583}
\newcommand{\RLoglenFeatureGapsMagistral}{0.27}
\newcommand{\TopFiftyFeatureGapsMagistral}{0.723}
\newcommand{\TopFiftyLmFeatureGapsMagistral}{0.708}
\newcommand{\SGenLengthMagistral}{0.523}
\newcommand{\SGenLengthMagistralMmlupro}{0.546}
\newcommand{\SGenLengthMagistralOmnimath}{0.538}
\newcommand{\SGenLengthMagistralSupergpqa}{0.491}
\newcommand{\SGenLengthMagistralTriviaqa}{0.517}
\newcommand{\AuprcGenLengthMagistral}{0.511}
\newcommand{\AuprcLmGenLengthMagistral}{0.396}
\newcommand{\AurcGenLengthMagistral}{0.249}
\newcommand{\AurcLmGenLengthMagistral}{0.335}
\newcommand{\AurocGenLengthMagistral}{0.682}
\newcommand{\AurocLmGenLengthMagistral}{0.523}
\newcommand{\RLoglenGenLengthMagistral}{0.90}
\newcommand{\TopFiftyGenLengthMagistral}{0.765}
\newcommand{\TopFiftyLmGenLengthMagistral}{0.666}
\newcommand{\SPEThinkMagistral}{0.630}
\newcommand{\SPEThinkMagistralMmlupro}{0.582}
\newcommand{\SPEThinkMagistralOmnimath}{0.714}
\newcommand{\SPEThinkMagistralSupergpqa}{0.652}
\newcommand{\SPEThinkMagistralTriviaqa}{0.573}
\newcommand{\AuprcPEThinkMagistral}{0.472}
\newcommand{\AuprcLmPEThinkMagistral}{0.504}
\newcommand{\AurcPEThinkMagistral}{0.301}
\newcommand{\AurcLmPEThinkMagistral}{0.283}
\newcommand{\AurocPEThinkMagistral}{0.602}
\newcommand{\AurocLmPEThinkMagistral}{0.630}
\newcommand{\RLoglenPEThinkMagistral}{-0.09}
\newcommand{\TopFiftyPEThinkMagistral}{0.725}
\newcommand{\TopFiftyLmPEThinkMagistral}{0.716}
\newcommand{\SPTrueMagistral}{0.527}
\newcommand{\ExtPTrueMagistral}{1}
\newcommand{\SPTrueMagistralMmlupro}{0.501}
\newcommand{\SPTrueMagistralOmnimath}{0.503}
\newcommand{\SPTrueMagistralSupergpqa}{0.456}
\newcommand{\SPTrueMagistralTriviaqa}{0.647}
\newcommand{\AuprcPTrueMagistral}{0.393}
\newcommand{\AuprcLmPTrueMagistral}{0.403}
\newcommand{\AurcPTrueMagistral}{0.339}
\newcommand{\AurcLmPTrueMagistral}{0.348}
\newcommand{\AurocPTrueMagistral}{0.547}
\newcommand{\AurocLmPTrueMagistral}{0.527}
\newcommand{\RLoglenPTrueMagistral}{0.06}
\newcommand{\TopFiftyPTrueMagistral}{0.667}
\newcommand{\TopFiftyLmPTrueMagistral}{0.654}
\newcommand{\GainAuprcMagistral}{+0.032}
\newcommand{\GainAuprcLmMagistral}{+0.042}
\newcommand{\GainAurcMagistral}{+0.032}
\newcommand{\GainAurcLmMagistral}{+0.033}
\newcommand{\GainAurocMagistral}{+0.040}
\newcommand{\GainAurocLmMagistral}{+0.049}
\newcommand{\GainTopFiftyMagistral}{+0.024}
\newcommand{\GainTopFiftyLmMagistral}{+0.030}
\newcommand{\NMagistralMmlupro}{974}
\newcommand{\ErrMagistralMmlupro}{24.3}
\newcommand{\NMagistralOmnimath}{928}
\newcommand{\ErrMagistralOmnimath}{40.8}
\newcommand{\NMagistralSupergpqa}{977}
\newcommand{\ErrMagistralSupergpqa}{55.4}
\newcommand{\NMagistralTriviaqa}{929}
\newcommand{\ErrMagistralTriviaqa}{20.6}

\newcommand{\SampledScopeGemma}{sampled subset, 500 prompts x 10 samples per dataset -- not the full six-dataset suite}
\newcommand{\SampledDatasetsGemma}{logiqa2, mmlupro, triviaqa}
\newcommand{\SampledAurocLengthNormalisedPredictiveEntropyGemma}{0.756}
\newcommand{\SampledAuprcLengthNormalisedPredictiveEntropyGemma}{0.389}
\newcommand{\SampledAurcLengthNormalisedPredictiveEntropyGemma}{0.070}
\newcommand{\SampledTopFiftyLengthNormalisedPredictiveEntropyGemma}{0.939}
\newcommand{\SampledAurocLmLengthNormalisedPredictiveEntropyGemma}{0.727}
\newcommand{\SampledAuprcLmLengthNormalisedPredictiveEntropyGemma}{0.391}
\newcommand{\SampledAurcLmLengthNormalisedPredictiveEntropyGemma}{0.086}
\newcommand{\SampledTopFiftyLmLengthNormalisedPredictiveEntropyGemma}{0.928}
\newcommand{\SampledRLoglenLengthNormalisedPredictiveEntropyGemma}{0.22}
\newcommand{\SampledAurocPredictiveEntropyGemma}{0.818}
\newcommand{\SampledAuprcPredictiveEntropyGemma}{0.505}
\newcommand{\SampledAurcPredictiveEntropyGemma}{0.055}
\newcommand{\SampledTopFiftyPredictiveEntropyGemma}{0.954}
\newcommand{\SampledAurocLmPredictiveEntropyGemma}{0.758}
\newcommand{\SampledAuprcLmPredictiveEntropyGemma}{0.433}
\newcommand{\SampledAurcLmPredictiveEntropyGemma}{0.080}
\newcommand{\SampledTopFiftyLmPredictiveEntropyGemma}{0.925}
\newcommand{\SampledRLoglenPredictiveEntropyGemma}{0.75}
\newcommand{\SampledAurocSemanticEntropyGemma}{0.725}
\newcommand{\SampledAuprcSemanticEntropyGemma}{0.281}
\newcommand{\SampledAurcSemanticEntropyGemma}{0.077}
\newcommand{\SampledTopFiftySemanticEntropyGemma}{0.935}
\newcommand{\SampledAurocLmSemanticEntropyGemma}{0.700}
\newcommand{\SampledAuprcLmSemanticEntropyGemma}{0.340}
\newcommand{\SampledAurcLmSemanticEntropyGemma}{0.089}
\newcommand{\SampledTopFiftyLmSemanticEntropyGemma}{0.925}
\newcommand{\SampledRLoglenSemanticEntropyGemma}{-0.02}
\newcommand{\SampledAurocSentenceSarGemma}{0.749}
\newcommand{\SampledAuprcSentenceSarGemma}{0.335}
\newcommand{\SampledAurcSentenceSarGemma}{0.056}
\newcommand{\SampledTopFiftySentenceSarGemma}{0.946}
\newcommand{\SampledAurocLmSentenceSarGemma}{0.651}
\newcommand{\SampledAuprcLmSentenceSarGemma}{0.312}
\newcommand{\SampledAurcLmSentenceSarGemma}{0.076}
\newcommand{\SampledTopFiftyLmSentenceSarGemma}{0.925}
\newcommand{\SampledRLoglenSentenceSarGemma}{0.62}
\newcommand{\SampledAurocTokenSarGemma}{0.743}
\newcommand{\SampledAuprcTokenSarGemma}{0.378}
\newcommand{\SampledAurcTokenSarGemma}{0.075}
\newcommand{\SampledTopFiftyTokenSarGemma}{0.935}
\newcommand{\SampledAurocLmTokenSarGemma}{0.708}
\newcommand{\SampledAuprcLmTokenSarGemma}{0.375}
\newcommand{\SampledAurcLmTokenSarGemma}{0.091}
\newcommand{\SampledTopFiftyLmTokenSarGemma}{0.919}
\newcommand{\SampledRLoglenTokenSarGemma}{0.19}

\newcommand{\CellAurocGenLengthGemma}{70.1\tiny{$\pm$0.5}}
\newcommand{\CellAurocLmGenLengthGemma}{52.1\tiny{$\pm$2.6}}
\newcommand{\CellAuprcGenLengthGemma}{40.1\tiny{$\pm$0.8}}
\newcommand{\CellAuprcLmGenLengthGemma}{26.2\tiny{$\pm$1.5}}
\newcommand{\CellAurcGenLengthGemma}{14.6\tiny{$\pm$0.3}}
\newcommand{\CellAurcLmGenLengthGemma}{20.3\tiny{$\pm$1.0}}
\newcommand{\CellAurocMSPGemma}{50.3\tiny{$\pm$0.8}}
\newcommand{\CellAurocLmMSPGemma}{49.6\tiny{$\pm$0.7}}
\newcommand{\CellAuprcMSPGemma}{21.8\tiny{$\pm$0.3}}
\newcommand{\CellAuprcLmMSPGemma}{24.3\tiny{$\pm$0.2}}
\newcommand{\CellAurcMSPGemma}{21.7\tiny{$\pm$0.5}}
\newcommand{\CellAurcLmMSPGemma}{21.5\tiny{$\pm$0.4}}
\newcommand{\CellAurocMaxEntropyGemma}{64.2\tiny{$\pm$1.1}}
\newcommand{\CellAurocLmMaxEntropyGemma}{58.8\tiny{$\pm$0.8}}
\newcommand{\CellAuprcMaxEntropyGemma}{34.0\tiny{$\pm$1.7}}
\newcommand{\CellAuprcLmMaxEntropyGemma}{31.0\tiny{$\pm$1.0}}
\newcommand{\CellAurcMaxEntropyGemma}{15.5\tiny{$\pm$0.3}}
\newcommand{\CellAurcLmMaxEntropyGemma}{17.6\tiny{$\pm$0.6}}
\newcommand{\CellAurocMeanNLLGemma}{67.9\tiny{$\pm$1.2}}
\newcommand{\CellAurocLmMeanNLLGemma}{65.5\tiny{$\pm$0.5}}
\newcommand{\CellAuprcMeanNLLGemma}{36.5\tiny{$\pm$1.4}}
\newcommand{\CellAuprcLmMeanNLLGemma}{36.8\tiny{$\pm$0.4}}
\newcommand{\CellAurcMeanNLLGemma}{13.6\tiny{$\pm$0.5}}
\newcommand{\CellAurcLmMeanNLLGemma}{14.7\tiny{$\pm$0.5}}
\newcommand{\CellAurocSumNLLGemma}{74.7\tiny{$\pm$0.3}}
\newcommand{\CellAurocLmSumNLLGemma}{64.9\tiny{$\pm$1.3}}
\newcommand{\CellAuprcSumNLLGemma}{46.5\tiny{$\pm$0.8}}
\newcommand{\CellAuprcLmSumNLLGemma}{36.9\tiny{$\pm$0.6}}
\newcommand{\CellAurcSumNLLGemma}{12.7\tiny{$\pm$0.2}}
\newcommand{\CellAurcLmSumNLLGemma}{14.7\tiny{$\pm$0.6}}
\newcommand{\CellAurocSelfCertaintyGemma}{67.8\tiny{$\pm$0.4}}
\newcommand{\CellAurocLmSelfCertaintyGemma}{56.1\tiny{$\pm$0.6}}
\newcommand{\CellAuprcSelfCertaintyGemma}{38.1\tiny{$\pm$1.3}}
\newcommand{\CellAuprcLmSelfCertaintyGemma}{31.0\tiny{$\pm$0.4}}
\newcommand{\CellAurcSelfCertaintyGemma}{14.2\tiny{$\pm$0.3}}
\newcommand{\CellAurcLmSelfCertaintyGemma}{18.0\tiny{$\pm$0.5}}
\newcommand{\CellAurocEDISGemma}{74.1\tiny{$\pm$0.2}}
\newcommand{\CellAurocLmEDISGemma}{63.5\tiny{$\pm$1.6}}
\newcommand{\CellAuprcEDISGemma}{45.9\tiny{$\pm$0.9}}
\newcommand{\CellAuprcLmEDISGemma}{35.7\tiny{$\pm$1.2}}
\newcommand{\CellAurcEDISGemma}{13.2\tiny{$\pm$0.1}}
\newcommand{\CellAurcLmEDISGemma}{15.3\tiny{$\pm$0.5}}
\newcommand{\CellAurocDeepConfGemma}{70.3\tiny{$\pm$0.8}}
\newcommand{\CellAurocLmDeepConfGemma}{65.6\tiny{$\pm$1.0}}
\newcommand{\CellAuprcDeepConfGemma}{38.5\tiny{$\pm$1.3}}
\newcommand{\CellAuprcLmDeepConfGemma}{36.0\tiny{$\pm$0.1}}
\newcommand{\CellAurcDeepConfGemma}{12.8\tiny{$\pm$0.4}}
\newcommand{\CellAurcLmDeepConfGemma}{14.6\tiny{$\pm$0.5}}
\newcommand{\CellAurocPTrueGemma}{53.1\tiny{$\pm$0.2}}
\newcommand{\CellAurocLmPTrueGemma}{53.2\tiny{$\pm$0.6}}
\newcommand{\CellAuprcPTrueGemma}{22.5\tiny{$\pm$0.5}}
\newcommand{\CellAuprcLmPTrueGemma}{25.0\tiny{$\pm$0.1}}
\newcommand{\CellAurcPTrueGemma}{20.0\tiny{$\pm$0.3}}
\newcommand{\CellAurcLmPTrueGemma}{20.4\tiny{$\pm$0.4}}
\newcommand{\CellAurocTokURGemma}{\textbf{75.1\tiny{$\pm$0.4}}}
\newcommand{\CellAurocLmTokURGemma}{65.1\tiny{$\pm$1.6}}
\newcommand{\CellAuprcTokURGemma}{\textbf{46.1\tiny{$\pm$2.5}}}
\newcommand{\CellAuprcLmTokURGemma}{35.7\tiny{$\pm$1.5}}
\newcommand{\CellAurcTokURGemma}{12.6\tiny{$\pm$0.1}}
\newcommand{\CellAurcLmTokURGemma}{15.5\tiny{$\pm$0.4}}
\newcommand{\CellAurocFeatureGapsGemma}{72.5\tiny{$\pm$1.1}}
\newcommand{\CellAurocLmFeatureGapsGemma}{66.2\tiny{$\pm$1.2}}
\newcommand{\CellAuprcFeatureGapsGemma}{40.9\tiny{$\pm$1.4}}
\newcommand{\CellAuprcLmFeatureGapsGemma}{38.8\tiny{$\pm$1.4}}
\newcommand{\CellAurcFeatureGapsGemma}{\textbf{11.8\tiny{$\pm$0.5}}}
\newcommand{\CellAurcLmFeatureGapsGemma}{14.5\tiny{$\pm$0.6}}
\newcommand{\CellAurocPEGemma}{68.3\tiny{$\pm$1.1}}
\newcommand{\CellAurocLmPEGemma}{66.1\tiny{$\pm$0.7}}
\newcommand{\CellAuprcPEGemma}{37.0\tiny{$\pm$1.3}}
\newcommand{\CellAuprcLmPEGemma}{36.9\tiny{$\pm$0.6}}
\newcommand{\CellAurcPEGemma}{13.5\tiny{$\pm$0.5}}
\newcommand{\CellAurcLmPEGemma}{14.6\tiny{$\pm$0.5}}
\newcommand{\CellAurocAConeGemma}{73.1\tiny{$\pm$0.8}}
\newcommand{\CellAurocLmAConeGemma}{66.3\tiny{$\pm$0.5}}
\newcommand{\CellAuprcAConeGemma}{44.3\tiny{$\pm$1.4}}
\newcommand{\CellAuprcLmAConeGemma}{38.4\tiny{$\pm$1.2}}
\newcommand{\CellAurcAConeGemma}{12.8\tiny{$\pm$0.5}}
\newcommand{\CellAurcLmAConeGemma}{15.1\tiny{$\pm$0.4}}
\newcommand{\CellAurocULensGemma}{73.2\tiny{$\pm$0.5}}
\newcommand{\CellAurocLmULensGemma}{\textbf{68.4\tiny{$\pm$0.4}}}
\newcommand{\CellAuprcULensGemma}{42.7\tiny{$\pm$0.9}}
\newcommand{\CellAuprcLmULensGemma}{\textbf{39.6\tiny{$\pm$0.3}}}
\newcommand{\CellAurcULensGemma}{\textbf{11.8\tiny{$\pm$0.2}}}
\newcommand{\CellAurcLmULensGemma}{\textbf{13.5\tiny{$\pm$0.3}}}
\newcommand{\CellAurocGenLengthQwen}{67.7\tiny{$\pm$1.2}}
\newcommand{\CellAurocLmGenLengthQwen}{52.7\tiny{$\pm$0.9}}
\newcommand{\CellAuprcGenLengthQwen}{35.5\tiny{$\pm$0.9}}
\newcommand{\CellAuprcLmGenLengthQwen}{25.8\tiny{$\pm$1.4}}
\newcommand{\CellAurcGenLengthQwen}{14.8\tiny{$\pm$0.4}}
\newcommand{\CellAurcLmGenLengthQwen}{19.1\tiny{$\pm$0.5}}
\newcommand{\CellAurocMSPQwen}{52.0\tiny{$\pm$0.5}}
\newcommand{\CellAurocLmMSPQwen}{50.3\tiny{$\pm$1.3}}
\newcommand{\CellAuprcMSPQwen}{22.1\tiny{$\pm$0.3}}
\newcommand{\CellAuprcLmMSPQwen}{24.8\tiny{$\pm$0.8}}
\newcommand{\CellAurcMSPQwen}{19.6\tiny{$\pm$0.3}}
\newcommand{\CellAurcLmMSPQwen}{20.5\tiny{$\pm$0.2}}
\newcommand{\CellAurocMaxEntropyQwen}{71.8\tiny{$\pm$0.5}}
\newcommand{\CellAurocLmMaxEntropyQwen}{67.9\tiny{$\pm$2.2}}
\newcommand{\CellAuprcMaxEntropyQwen}{37.8\tiny{$\pm$0.7}}
\newcommand{\CellAuprcLmMaxEntropyQwen}{36.9\tiny{$\pm$1.7}}
\newcommand{\CellAurcMaxEntropyQwen}{12.9\tiny{$\pm$0.3}}
\newcommand{\CellAurcLmMaxEntropyQwen}{14.2\tiny{$\pm$0.4}}
\newcommand{\CellAurocMeanNLLQwen}{72.6\tiny{$\pm$0.6}}
\newcommand{\CellAurocLmMeanNLLQwen}{68.4\tiny{$\pm$0.8}}
\newcommand{\CellAuprcMeanNLLQwen}{40.5\tiny{$\pm$0.2}}
\newcommand{\CellAuprcLmMeanNLLQwen}{39.0\tiny{$\pm$0.5}}
\newcommand{\CellAurcMeanNLLQwen}{11.8\tiny{$\pm$0.4}}
\newcommand{\CellAurcLmMeanNLLQwen}{13.0\tiny{$\pm$0.6}}
\newcommand{\CellAurocSumNLLQwen}{72.7\tiny{$\pm$1.1}}
\newcommand{\CellAurocLmSumNLLQwen}{67.7\tiny{$\pm$0.4}}
\newcommand{\CellAuprcSumNLLQwen}{41.6\tiny{$\pm$1.9}}
\newcommand{\CellAuprcLmSumNLLQwen}{38.7\tiny{$\pm$0.9}}
\newcommand{\CellAurcSumNLLQwen}{12.9\tiny{$\pm$0.4}}
\newcommand{\CellAurcLmSumNLLQwen}{13.2\tiny{$\pm$0.6}}
\newcommand{\CellAurocSelfCertaintyQwen}{72.2\tiny{$\pm$0.4}}
\newcommand{\CellAurocLmSelfCertaintyQwen}{69.1\tiny{$\pm$0.3}}
\newcommand{\CellAuprcSelfCertaintyQwen}{37.5\tiny{$\pm$0.1}}
\newcommand{\CellAuprcLmSelfCertaintyQwen}{36.8\tiny{$\pm$0.2}}
\newcommand{\CellAurcSelfCertaintyQwen}{11.8\tiny{$\pm$0.2}}
\newcommand{\CellAurcLmSelfCertaintyQwen}{12.8\tiny{$\pm$0.2}}
\newcommand{\CellAurocEDISQwen}{71.9\tiny{$\pm$1.2}}
\newcommand{\CellAurocLmEDISQwen}{66.3\tiny{$\pm$0.7}}
\newcommand{\CellAuprcEDISQwen}{40.6\tiny{$\pm$1.9}}
\newcommand{\CellAuprcLmEDISQwen}{37.3\tiny{$\pm$1.5}}
\newcommand{\CellAurcEDISQwen}{13.3\tiny{$\pm$0.4}}
\newcommand{\CellAurcLmEDISQwen}{13.6\tiny{$\pm$0.6}}
\newcommand{\CellAurocDeepConfQwen}{72.4\tiny{$\pm$0.5}}
\newcommand{\CellAurocLmDeepConfQwen}{69.1\tiny{$\pm$0.5}}
\newcommand{\CellAuprcDeepConfQwen}{38.2\tiny{$\pm$0.1}}
\newcommand{\CellAuprcLmDeepConfQwen}{37.9\tiny{$\pm$0.1}}
\newcommand{\CellAurcDeepConfQwen}{11.4\tiny{$\pm$0.3}}
\newcommand{\CellAurcLmDeepConfQwen}{12.5\tiny{$\pm$0.3}}
\newcommand{\CellAurocPTrueQwen}{62.4\tiny{$\pm$1.0}}
\newcommand{\CellAurocLmPTrueQwen}{60.7\tiny{$\pm$0.9}}
\newcommand{\CellAuprcPTrueQwen}{31.6\tiny{$\pm$2.3}}
\newcommand{\CellAuprcLmPTrueQwen}{32.8\tiny{$\pm$1.6}}
\newcommand{\CellAurcPTrueQwen}{16.1\tiny{$\pm$0.3}}
\newcommand{\CellAurcLmPTrueQwen}{17.4\tiny{$\pm$0.5}}
\newcommand{\CellAurocTokURQwen}{68.3\tiny{$\pm$3.3}}
\newcommand{\CellAurocLmTokURQwen}{61.3\tiny{$\pm$0.5}}
\newcommand{\CellAuprcTokURQwen}{35.9\tiny{$\pm$2.5}}
\newcommand{\CellAuprcLmTokURQwen}{32.0\tiny{$\pm$2.4}}
\newcommand{\CellAurcTokURQwen}{13.6\tiny{$\pm$1.0}}
\newcommand{\CellAurcLmTokURQwen}{14.8\tiny{$\pm$1.1}}
\newcommand{\CellAurocFeatureGapsQwen}{69.8\tiny{$\pm$2.8}}
\newcommand{\CellAurocLmFeatureGapsQwen}{63.5\tiny{$\pm$1.7}}
\newcommand{\CellAuprcFeatureGapsQwen}{39.9\tiny{$\pm$2.2}}
\newcommand{\CellAuprcLmFeatureGapsQwen}{37.7\tiny{$\pm$1.6}}
\newcommand{\CellAurcFeatureGapsQwen}{13.4\tiny{$\pm$1.0}}
\newcommand{\CellAurcLmFeatureGapsQwen}{15.0\tiny{$\pm$0.7}}
\newcommand{\CellAurocPEQwen}{72.8\tiny{$\pm$0.6}}
\newcommand{\CellAurocLmPEQwen}{68.9\tiny{$\pm$1.0}}
\newcommand{\CellAuprcPEQwen}{40.8\tiny{$\pm$0.2}}
\newcommand{\CellAuprcLmPEQwen}{39.6\tiny{$\pm$0.4}}
\newcommand{\CellAurcPEQwen}{11.7\tiny{$\pm$0.3}}
\newcommand{\CellAurcLmPEQwen}{12.8\tiny{$\pm$0.5}}
\newcommand{\CellAurocAConeQwen}{71.5\tiny{$\pm$0.7}}
\newcommand{\CellAurocLmAConeQwen}{66.7\tiny{$\pm$0.9}}
\newcommand{\CellAuprcAConeQwen}{41.2\tiny{$\pm$1.1}}
\newcommand{\CellAuprcLmAConeQwen}{39.0\tiny{$\pm$1.1}}
\newcommand{\CellAurcAConeQwen}{11.6\tiny{$\pm$0.5}}
\newcommand{\CellAurcLmAConeQwen}{13.5\tiny{$\pm$0.5}}
\newcommand{\CellAurocULensQwen}{\textbf{74.3\tiny{$\pm$0.7}}}
\newcommand{\CellAurocLmULensQwen}{\textbf{71.0\tiny{$\pm$0.7}}}
\newcommand{\CellAuprcULensQwen}{\textbf{43.5\tiny{$\pm$0.7}}}
\newcommand{\CellAuprcLmULensQwen}{\textbf{42.1\tiny{$\pm$0.7}}}
\newcommand{\CellAurcULensQwen}{\textbf{11.0\tiny{$\pm$0.4}}}
\newcommand{\CellAurcLmULensQwen}{\textbf{12.1\tiny{$\pm$0.6}}}
\newcommand{\CellAurocGenLengthMagistral}{67.8\tiny{$\pm$0.5}}
\newcommand{\CellAurocLmGenLengthMagistral}{52.3\tiny{$\pm$0.5}}
\newcommand{\CellAuprcGenLengthMagistral}{49.8\tiny{$\pm$1.1}}
\newcommand{\CellAuprcLmGenLengthMagistral}{39.1\tiny{$\pm$0.6}}
\newcommand{\CellAurcGenLengthMagistral}{24.6\tiny{$\pm$0.2}}
\newcommand{\CellAurcLmGenLengthMagistral}{33.6\tiny{$\pm$0.2}}
\newcommand{\CellAurocMSPMagistral}{57.6\tiny{$\pm$0.5}}
\newcommand{\CellAurocLmMSPMagistral}{57.6\tiny{$\pm$0.6}}
\newcommand{\CellAuprcMSPMagistral}{39.9\tiny{$\pm$1.0}}
\newcommand{\CellAuprcLmMSPMagistral}{42.2\tiny{$\pm$1.1}}
\newcommand{\CellAurcMSPMagistral}{29.0\tiny{$\pm$0.2}}
\newcommand{\CellAurcLmMSPMagistral}{29.7\tiny{$\pm$0.2}}
\newcommand{\CellAurocMaxEntropyMagistral}{67.1\tiny{$\pm$0.3}}
\newcommand{\CellAurocLmMaxEntropyMagistral}{61.2\tiny{$\pm$0.7}}
\newcommand{\CellAuprcMaxEntropyMagistral}{49.4\tiny{$\pm$1.3}}
\newcommand{\CellAuprcLmMaxEntropyMagistral}{46.5\tiny{$\pm$1.2}}
\newcommand{\CellAurcMaxEntropyMagistral}{24.5\tiny{$\pm$0.2}}
\newcommand{\CellAurcLmMaxEntropyMagistral}{28.7\tiny{$\pm$0.0}}
\newcommand{\CellAurocMeanNLLMagistral}{60.2\tiny{$\pm$0.3}}
\newcommand{\CellAurocLmMeanNLLMagistral}{61.9\tiny{$\pm$0.5}}
\newcommand{\CellAuprcMeanNLLMagistral}{46.2\tiny{$\pm$0.8}}
\newcommand{\CellAuprcLmMeanNLLMagistral}{48.1\tiny{$\pm$1.2}}
\newcommand{\CellAurcMeanNLLMagistral}{29.6\tiny{$\pm$0.5}}
\newcommand{\CellAurcLmMeanNLLMagistral}{28.4\tiny{$\pm$0.3}}
\newcommand{\CellAurocSumNLLMagistral}{70.3\tiny{$\pm$0.6}}
\newcommand{\CellAurocLmSumNLLMagistral}{61.8\tiny{$\pm$0.4}}
\newcommand{\CellAuprcSumNLLMagistral}{53.1\tiny{$\pm$0.9}}
\newcommand{\CellAuprcLmSumNLLMagistral}{47.5\tiny{$\pm$1.2}}
\newcommand{\CellAurcSumNLLMagistral}{22.9\tiny{$\pm$0.5}}
\newcommand{\CellAurcLmSumNLLMagistral}{28.3\tiny{$\pm$0.3}}
\newcommand{\CellAurocSelfCertaintyMagistral}{60.8\tiny{$\pm$0.4}}
\newcommand{\CellAurocLmSelfCertaintyMagistral}{62.4\tiny{$\pm$0.9}}
\newcommand{\CellAuprcSelfCertaintyMagistral}{46.9\tiny{$\pm$0.8}}
\newcommand{\CellAuprcLmSelfCertaintyMagistral}{48.2\tiny{$\pm$0.6}}
\newcommand{\CellAurcSelfCertaintyMagistral}{28.9\tiny{$\pm$0.6}}
\newcommand{\CellAurcLmSelfCertaintyMagistral}{28.1\tiny{$\pm$0.4}}
\newcommand{\CellAurocEDISMagistral}{70.3\tiny{$\pm$0.6}}
\newcommand{\CellAurocLmEDISMagistral}{61.3\tiny{$\pm$1.0}}
\newcommand{\CellAuprcEDISMagistral}{53.0\tiny{$\pm$1.0}}
\newcommand{\CellAuprcLmEDISMagistral}{46.9\tiny{$\pm$1.7}}
\newcommand{\CellAurcEDISMagistral}{23.0\tiny{$\pm$0.4}}
\newcommand{\CellAurcLmEDISMagistral}{28.6\tiny{$\pm$0.2}}
\newcommand{\CellAurocDeepConfMagistral}{60.9\tiny{$\pm$0.4}}
\newcommand{\CellAurocLmDeepConfMagistral}{63.6\tiny{$\pm$1.0}}
\newcommand{\CellAuprcDeepConfMagistral}{46.6\tiny{$\pm$0.8}}
\newcommand{\CellAuprcLmDeepConfMagistral}{49.0\tiny{$\pm$1.0}}
\newcommand{\CellAurcDeepConfMagistral}{28.4\tiny{$\pm$0.6}}
\newcommand{\CellAurcLmDeepConfMagistral}{26.9\tiny{$\pm$0.4}}
\newcommand{\CellAurocPTrueMagistral}{54.5\tiny{$\pm$0.4}}
\newcommand{\CellAurocLmPTrueMagistral}{53.0\tiny{$\pm$0.3}}
\newcommand{\CellAuprcPTrueMagistral}{38.7\tiny{$\pm$0.9}}
\newcommand{\CellAuprcLmPTrueMagistral}{40.2\tiny{$\pm$0.4}}
\newcommand{\CellAurcPTrueMagistral}{33.7\tiny{$\pm$0.3}}
\newcommand{\CellAurcLmPTrueMagistral}{34.4\tiny{$\pm$0.4}}
\newcommand{\CellAurocTokURMagistral}{65.5\tiny{$\pm$2.7}}
\newcommand{\CellAurocLmTokURMagistral}{54.2\tiny{$\pm$2.6}}
\newcommand{\CellAuprcTokURMagistral}{42.2\tiny{$\pm$7.0}}
\newcommand{\CellAuprcLmTokURMagistral}{37.2\tiny{$\pm$5.1}}
\newcommand{\CellAurcTokURMagistral}{\textbf{22.4\tiny{$\pm$1.4}}}
\newcommand{\CellAurcLmTokURMagistral}{27.2\tiny{$\pm$3.3}}
\newcommand{\CellAurocFeatureGapsMagistral}{64.1\tiny{$\pm$1.5}}
\newcommand{\CellAurocLmFeatureGapsMagistral}{60.2\tiny{$\pm$0.9}}
\newcommand{\CellAuprcFeatureGapsMagistral}{44.3\tiny{$\pm$4.0}}
\newcommand{\CellAuprcLmFeatureGapsMagistral}{46.7\tiny{$\pm$1.1}}
\newcommand{\CellAurcFeatureGapsMagistral}{24.7\tiny{$\pm$2.2}}
\newcommand{\CellAurcLmFeatureGapsMagistral}{27.1\tiny{$\pm$3.7}}
\newcommand{\CellAurocPEMagistral}{60.8\tiny{$\pm$0.1}}
\newcommand{\CellAurocLmPEMagistral}{62.6\tiny{$\pm$0.5}}
\newcommand{\CellAuprcPEMagistral}{46.9\tiny{$\pm$0.8}}
\newcommand{\CellAuprcLmPEMagistral}{48.6\tiny{$\pm$1.1}}
\newcommand{\CellAurcPEMagistral}{29.3\tiny{$\pm$0.5}}
\newcommand{\CellAurcLmPEMagistral}{28.0\tiny{$\pm$0.4}}
\newcommand{\CellAurocAConeMagistral}{\textbf{68.5\tiny{$\pm$0.7}}}
\newcommand{\CellAurocLmAConeMagistral}{67.6\tiny{$\pm$1.1}}
\newcommand{\CellAuprcAConeMagistral}{\textbf{51.8\tiny{$\pm$0.6}}}
\newcommand{\CellAuprcLmAConeMagistral}{52.2\tiny{$\pm$1.6}}
\newcommand{\CellAurcAConeMagistral}{23.5\tiny{$\pm$0.5}}
\newcommand{\CellAurcLmAConeMagistral}{24.5\tiny{$\pm$0.6}}
\newcommand{\CellAurocULensMagistral}{65.7\tiny{$\pm$0.7}}
\newcommand{\CellAurocLmULensMagistral}{\textbf{68.3\tiny{$\pm$1.0}}}
\newcommand{\CellAuprcULensMagistral}{50.3\tiny{$\pm$1.0}}
\newcommand{\CellAuprcLmULensMagistral}{\textbf{53.2\tiny{$\pm$1.3}}}
\newcommand{\CellAurcULensMagistral}{25.7\tiny{$\pm$0.8}}
\newcommand{\CellAurcLmULensMagistral}{\textbf{24.4\tiny{$\pm$0.5}}}
\newcommand{\RowGenLength}{Generation length}
\newcommand{\RowMSP}{MSP}
\newcommand{\RowMaxEntropy}{Maximum\ Entropy}
\newcommand{\RowMeanNLL}{Mean NLL}
\newcommand{\RowSumNLL}{Sum NLL}
\newcommand{\RowSelfCertainty}{Self-Certainty}
\newcommand{\RowEDIS}{EDIS}
\newcommand{\RowDeepConf}{DeepConf}
\newcommand{\RowPTrue}{P(True)}
\newcommand{\RowTokUR}{TokUR}
\newcommand{\RowFeatureGaps}{Feature-Gaps}
\newcommand{\RowPE}{Predictive Entropy}
\newcommand{\RowACone}{$A_{\mathrm{cone}}$ alone}
\newcommand{\RowULens}{\textbf{\ulens{}}}
\newcommand{\ArrowAuroc}{$\uparrow$}
\newcommand{\NameAuroc}{AUROC}
\newcommand{\ArrowAuprc}{$\uparrow$}
\newcommand{\NameAuprc}{AUPRC}
\newcommand{\ArrowAurc}{$\downarrow$}
\newcommand{\NameAurc}{AURC}
\newcommand{\SeedCount}{1}
\newcommand{\WorstMarginPE}{+0.017}
\newcommand{\WorstMarginBl}{+0.012}
\newcommand{\WorstMarginFreeBl}{+0.012}
\newcommand{\MetricBins}{10}
\newcommand{\DsLabelMmlupro}{MMLU-Pro}
\newcommand{\DsLabelOmnimath}{Omni-MATH}
\newcommand{\DsLabelSupergpqa}{SuperGPQA}
\newcommand{\DsLabelTriviaqa}{TriviaQA}
\begin{center}
  \includegraphics[width=\linewidth]{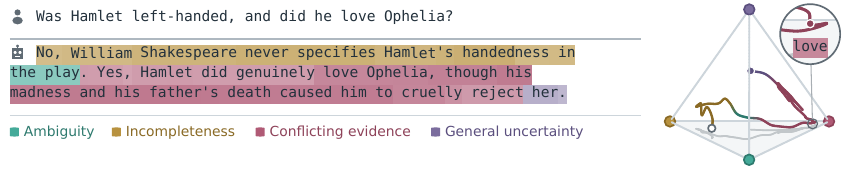}
  \captionof{figure}{Surfacing distinct sources of uncertainty through the \uspace{}.}
  \label{fig:hamlet_teaser}
\end{center}

\begin{abstract}
Large language models are informing decisions with ever-higher stakes. As the consequences of their errors grow, a central question becomes harder to ignore: how much can we trust an individual answer? When a wrong answer can cause serious harm, the cost of error may far exceed that of deferring to an expert. Yet recognizing when to defer remains difficult because language models can present incorrect conclusions with fluent explanations and an authoritative tone.
Uncertainty quantification seeks to address this disconnect by estimating the reliability of individual predictions. However, many existing methods require repeated generations or separately trained components, and their scalar estimates do not reveal where uncertainty arises or how it evolves during reasoning.  Recent work has also shown that generation length can be strongly associated with uncertainty estimates and correctness, raising the question of how much of an estimator's predictive power comes from uncertainty-specific information rather than output length alone. Mechanistic interpretability offers a way to address these limitations by connecting human-interpretable concepts to intermediate model states. Building on this capability, we introduce the \uspace{}, a low-dimensional subspace that makes a model's evolving uncertainty measurable and interpretable.
We identify semantic anchors for doubt and certainty, map their unembedding directions back into the residual space, and combine their contrasts into an orthogonal basis. The \ulens{} projects each token state onto these basis vectors, yielding an interpretable token-level uncertainty map that can be inspected directly or aggregated into a scalar uncertainty score. Our approach requires no correctness labels, repeated generations, or training. Across reasoning benchmarks, its confidence score outperforms established baselines under both standard and length-controlled evaluation and transfers more reliably than supervised estimators. Code is available at \url{https://github.com/s2labres/U-Space}.

\end{abstract}

\section{Introduction}
\label{sec:intro}

Language models no longer merely compete on benchmarks. Their outputs increasingly shape real-world decisions in settings where a mistake can carry serious consequences. As their reach grows, aggregate accuracy alone is insufficient: it tells us how often a model succeeds, but not which outputs are more likely to fail. Uncertainty quantification complements it by assigning each prediction a confidence or risk score intended to distinguish more from less reliable outputs.
Yet obtaining routine confidence estimates in deployed systems remains difficult. Many methods rely on repeated model calls or separately trained components, increasing inference costs and implementation complexity. This constraint is becoming more acute as reasoning models increasingly trade additional test-time computation, often measured in generated tokens, for improved performance~\citep{snell2024scaling, bai2026ai}. But cost is only part of the problem. Many estimators ultimately compress uncertainty into a single score without revealing which part of a response produced it or how it evolved during generation. They therefore risk replacing blind trust in the model with blind trust in the uncertainty metric. This loss of context is particularly pronounced for reasoning models, where uncertainty may arise midway through a derivation and remain hidden beneath an authoritative final answer. 
Reasoning traces introduce an additional evaluation challenge: \emph{generation length}. 
Incorrect answers are often longer, and many uncertainty estimators implicitly rely on this signal for error prediction~\citep{ kang2025selfcertainty}.  This calls for distinguishing overall predictive utility from the information an estimator captures beyond output length.
Accordingly, we report standard results alongside length-controlled evaluation. Our goal is to capture an interpretable uncertainty signal that remains predictive beyond the generation length and reveals where and why uncertainty arises during reasoning.

Mechanistic interpretability may provide a way toward uncovering the evolution of uncertainty during autoregressive generation. 
\citet{gurnee2026verbalizable} introduce the \jlens{}, which uses an averaged Jacobian together with the model's unembedding matrix to map intermediate residual states to scores over output tokens.
Building on this observation, we introduce the \uspace{}, a low-dimensional subspace of the residual stream that captures distinct sources of uncertainty as they arise and fade during generation.
We begin with small sets of interpretable semantic anchors expressing forms of doubt and certainty. Their unembedding directions are pulled back through the averaged Jacobian, normalized, and combined into a contrastive direction per uncertainty category. Projecting a hidden state onto the individual directions reveals which sources of uncertainty are present, while its projection onto their positive cone measures their aggregate strength at that point in the computation. %
We interpret this cone alignment as a verbalizable, second-order uncertainty signal\footnote{Following~\citet{kumaran2026causal}, \emph{first-order} uncertainty is expressed in output probabilities and \emph{second-order} uncertainty is verbalizable by the model; these terms describe levels of confidence, not derivative order.}: it reflects a representation of uncertainty that can be expressed through semantic concepts~\citep{fleming2017self,kumaran2026causal}. 
To obtain the final uncertainty score, we complement it with first-order distributional uncertainty measured by mean predictive entropy, since the entropy chain rule motivates aggregation across autoregressive steps and averaging avoids mechanical scaling with sequence length~\citep{malinin2021uncertainty}. The resulting \ulens{} therefore provides an interpretable account of where and why uncertainty arises while drawing on both signals for uncertainty quantification.

Applied across the reasoning trace, the \ulens{} produces a token-by-category map showing where distinct forms of uncertainty emerge and recede. 
Given a precomputed lens, this requires no correctness labels, repeated generations, or task-specific training.
Our core contributions are:
\begin{itemize}

\item We introduce a training-free recipe for constructing interpretable, contrastive semantic directions in a model's residual stream. Instantiated as \uspace{}, these directions reveal when uncertainty arises and characterize the form it takes.

\item To our knowledge, we introduce the first uncertainty estimator to jointly operationalize first- and second-order uncertainty. The \ulens{} combines predictive entropy with a verbalizable uncertainty signal while retaining a token-level semantic account of where and why uncertainty arises.

\item Across three reasoning models and four benchmarks, the resulting trace-level score outperforms established baselines under both standard and length-controlled evaluation and transfers more reliably than supervised estimators.

\item Interventions along the \uspace{} basis directions  can make the model hesitate and lose confidence even on simple questions, providing causal evidence that they capture representations that shape behavior rather than merely correlate with it.

\end{itemize}

\section{Background \& Related Work}
\label{sec:related_work}

\paragraph{Predictive uncertainty.}

\emph{Uncertainty quantification} asks how much confidence to place in an individual prediction. In classification, the object of interest is a probability distribution over a fixed set of labels, commonly summarized by the probability assigned to the selected class or by the entropy of the complete distribution~\citep{hendrycks2017baseline}. %
Autoregressive language generation makes this picture less direct.
A model assigns probability to a sequence through
\begin{equation*}
    p(y\mid x)
    =
    \prod_{t=1}^{T}
    p(y_t\mid x,y_{<t}),
\end{equation*}
but sequence probability is not equivalent to answer reliability. It depends on response length and on the particular words used to express an answer. Moreover, many distinct sequences may communicate the same conclusion. Early extensions, therefore, aggregated token probabilities or entropy across a response \citep{malinin2021uncertainty}, while subsequent work moved from lexical probability toward uncertainty over meaning.
One route asks the model to assess its own answer, for example, through verbal confidence or the probability of a ``True'' token \citep{kadavath2022language,tian2023just,xiong2024can}. This approach is compatible with black-box models, but self-verbalized confidence is often overconfident and may therefore fail to faithfully reflect the model's uncertainty \citep{xiong2024can}. A second route estimates uncertainty from repeated generations. SelfCheckGPT detects unsupported statements through inconsistency between samples \citep{manakul2023selfcheckgpt}. Semantic Entropy groups generations by meaning before measuring their dispersion \citep{kuhn2023semantic}. These methods address the many-to-one relationship between language and meaning, at the expense of requiring repeated generations and a separate model to group those generations.

\paragraph{Uncertainty across long-form reasoning.}

Long-form generation further exposes the limits of a single response-level score. A mostly correct answer may contain one unreliable claim, and a reasoning trace may become uncertain only at a particular point during autoregression. LUQ measures agreement between semantic units across sampled long-form responses \citep{zhang2024luq}, while claim-conditioned probability isolates uncertainty about a claim from variation in how it is expressed \citep{fadeeva2024factchecking}. These methods provide finer output-side localization, but they do not reveal how uncertainty is represented inside the model.
Reasoning models introduce a temporal dimension to uncertainty. %
Recent work, therefore, models token-level uncertainty across the generated trajectory.
\selfcert{} and \deepconf{} use token distributions to rank or filter reasoning traces \citep{kang2025selfcertainty,fu2026deep}. \tokur{} applies low-rank weight perturbations to estimate token-level Bayesian uncertainty \citep{zhang2026tokur}. Concurrent work either summarizes temporal uncertainty profiles with correctness-trained classifiers~\citep{grunefeld2026tracing} or combines interpretable features into uncertainty scores using correctness supervision~\citep{bakman2026uncertainty}.
Together, these studies show that uncertainty dynamics carry information lost through answer-level aggregation. Yet their methods either require repeated model evaluations, correctness labels for calibration or training, or provide no interpretable attribution. %
The \ulens{}, in contrast, operates on a single generation without correctness supervision while revealing where and why uncertainty arises. %

\paragraph{Generation length in uncertainty evaluation.}
Recent work has clarified that response length can be both informative and entangled with uncertainty evaluation. \citet{devic2025trace} show that trace length itself is a competitive zero-shot confidence signal after reasoning post-training. At the same time, \citet{heo2025llms} find systematic length differences between naturally generated correct and incorrect responses and introduce a length-neutral controlled benchmark. \citet{vashurin-etal-2025-uncertainty} show that length dependence persists in several nominally normalized uncertainty scores and propose post-hoc detrending, while preserving length information when it is genuinely associated with output quality. \citet{santilli-etal-2025-revisiting} further show that shared length biases in uncertainty scores and correctness functions can distort evaluation. Most closely related to our protocol, \citet{li-etal-2026-confidence} recompute uncertainty metrics within bins defined by output word count. We comply with this emerging evaluation practice by reporting both unadjusted performance and length-controlled diagnostics. %

\paragraph{Interpreting internal representations.}

A parallel line of work reads reliability from hidden states. Learned probes can detect false statements \citep{azaria2023internal}, approximate semantic entropy \citep{kossen2025semantic}, or predict the correctness of intermediate answers \citep{zhang2025reasoning}. Circuit-based reasoning verification similarly classifies features of attribution graphs and uses them to guide interventions \citep{zhao2026crv}. %
Mechanistic interpretability instead seeks to map internal states to human-understandable concepts. The logit lens applies the model's unembedding directly to intermediate residual states %
\citep{nostalgebraist2020logitlens}. The tuned lens learns a layer-specific affine correction \citep{belrose2023tunedlens}. The \jlens{} avoids learning such a decoder and instead linearizes the model's downstream computation. For a residual state \(h_{\ell,t}\) at layer $\ell$ and token $t$, it computes
\begin{equation*}
    \bar{J}_{\ell}
    =
    \mathbb{E}_{x,t,t'\geq t}
    \left[
        \frac{\partial h_{L,t'}}{\partial h_{\ell,t}}
    \right],
\end{equation*}
where \(x\) denotes an input and \(L\) the final layer. This Jacobian maps intermediate states into the final-layer representation space, where they can be read through the model's unembedding. 
Each vocabulary item thus induces a layer-specific residual direction. These verbalizable directions form a representation associated with reporting, reasoning, and downstream control \citep{gurnee2026verbalizable}.

Related work further suggests that human-interpretable concepts are often linearly represented~\citep{park2024linear} and that contrasts between semantic poles can isolate such concepts by suppressing shared components~\citep{marks2024geometry}. We apply this principle to the vocabulary directions above, subtracting certainty from uncertainty poles to construct interpretable axes spanning \uspace{}. Unlike correctness probes, these axes are defined by semantic anchors rather than task labels.
\section{Method: \uspace{} and the \ulens{}}
\label{sec:method}

We introduce the \uspace{}, a low-dimensional subspace of the residual stream that captures distinct sources of uncertainty. Building on the theoretical foundation of \citet{bakman2026uncertainty} and the second-order confidence model of \citet{fleming2017self}, we decompose uncertainty into an interpretable, verbalizable component and a first-order component that need not be accessible to verbal report. Following the global-workspace account of \citet{gurnee2026verbalizable}, we view the \uspace{} as an uncertainty-specific slice of the model's verbalizable workspace, where concepts are represented in a form available to internal reasoning and potential report.
We then introduce the \emph{Uncertainty-Lens} (\ulens{}), a linear readout that projects the hidden state onto this basis. %
We combine this interpretable alignment with a complementary first-order uncertainty signal
to obtain an overall scalar uncertainty estimate. Across a reasoning trace, the \ulens{} thereby yields (1) a token-level semantic attribution indicating which of the uncertainty categories each token most strongly expresses and (2) a scalar trace-level uncertainty score that combines verbalizable and distributional uncertainty.

\paragraph{Uncertainty taxonomy and motivating observation.}
To make uncertainty interpretable, we begin with a compact set of semantic categories inspired by taxonomies of imperfect information~\citep{parsons1996current}. Our category set comprises $C=4$ forms of uncertainty: \catamb{}, \catinc{}, \catcon{}, and \catgen{}. \Amb{} covers vagueness and imprecision, \inc{} captures missing information or knowledge, \con{} captures inconsistency, and \gen{} captures broad expressions of uncertainty that do not point to a particular cause and may accompany the other categories or reflect additional concerns, such as source reliability. We do not intend this taxonomy to be exhaustive. Rather, its four categories provide a compact set of actionable dimensions for describing why uncertainty arises, with the intended benefit that they can be represented as the vertices of a three-dimensional simplex (\Cref{fig:hamlet_teaser}).
However, this taxonomy raises the question of whether models internally distinguish these forms in human-interpretable terms. Prior work shows that contrastive activation differences can isolate semantically meaningful directions in the residual stream~\citep{turner2023actadd,marks2024geometry}. As a preliminary probe, we contrast hidden states from minimal prompt pairs that induce each category without naming it. When decoded through the \jlens{}, the resulting directions recover category-aligned vocabulary. The \amb{} direction promotes \emph{ambiguous} and \emph{ambiguity}, \con{} promotes \emph{controversial} and \emph{contentious}, \inc{} promotes \emph{cannot} and \emph{unable}, and \gen{} promotes \emph{dubious}, \emph{suspicious} and \emph{questionable} (\Cref{app:decodability}). These observations suggest that our taxonomy reflects meaningful distinctions in the model’s internal representations, providing a natural foundation for the \uspace{}. 

\subsection{Constructing the \uspace{}}
\label{sec:uspace}
\begin{figure}[t]
  \centering
  \includegraphics[width=\linewidth]{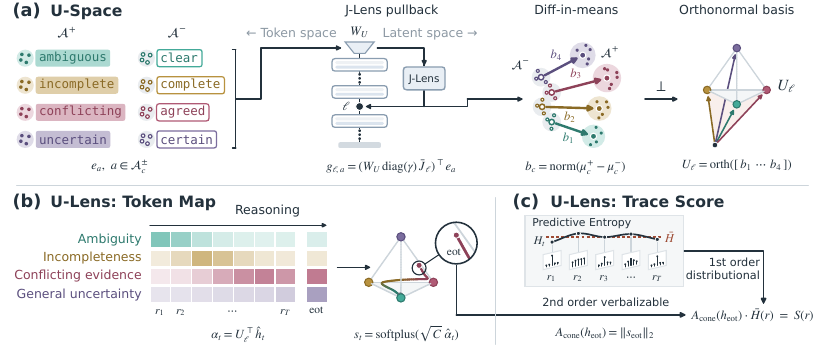}
  \caption{\textbf{\uspace{} construction and \ulens{} readout.}
\textbf{(a)} Uncertainty and certainty anchors are pulled from vocabulary space into the residual stream through the \jlens{}. Differences between their category centroids are orthogonalized to form the \uspace{} basis $U_\ell$.
\textbf{(b)} The \ulens{} projects each reasoning-token state onto this basis, yielding a token-by-category uncertainty map. Smooth rectification maps these coordinates into the positive cone.
\textbf{(c)} At the end-of-thinking marker, the norm of this rectified alignment provides a second-order verbalizable signal. Multiplying it by the mean predictive entropy over the reasoning trace yields the final uncertainty score $S(r)$.}
  \label{fig:method}
\end{figure}

\paragraph{Finding the \uspace{} basis.}
We construct the \uspace{} basis from a small set of uncertainty and certainty terms drawn from~\citet{chen2018scalable} and~\citet{rocklage2023beyond}, respectively. For each category $c$, we define a positive anchor set $\mathcal{A}_c^{+}$ expressing uncertainty and a negative set $\mathcal{A}_c^{-}$ expressing the corresponding certainty. After removing multi-token terms, we assign each remaining term to one of the four categories using a lightweight semantic model \citep{li2023angle} (see \Cref{app:anchor-construction} for details).
To express an anchor $a$ in the residual stream at layer $\ell$, we map its vocabulary direction $e_a$ \emph{backward} through a precomputed mean Jacobian:
$
g_{\ell,a}=(W_U\operatorname{diag}(\gain)\bar J_\ell)^\top e_a,
$
where $W_U$ is the model's unembedding matrix and $\gain$ the final RMSNorm gain \citep{zhang-sennrich-neurips19}, which scales the residual stream elementwise before the unembedding. This corresponds to the \jlens{} pullback of the anchor~\citep{gurnee2026verbalizable}. Precomputed model-specific Jacobians\footnote{\url{https://huggingface.co/neuronpedia/jacobian-lens}} are publicly available for some models. For models without a precomputed Jacobian, we estimate $\bar J_\ell$ from unlabeled, generic prompts without optimization or parameter updates. %
We detail this procedure in \Cref{app:jacobian-construction}.
Having mapped the anchors into the residual stream, we pool each semantic pole separately and subtract the resulting centroids. Letting \(\operatorname{norm}(v) := v/\lVert v\rVert_2\) denote \(\ell_2\)-normalization, we obtain a bipolar category direction:

\begin{equation}\label{eq:diff-construction}
\mu_{\ell,c}^{\pm}
=
\operatorname{norm}\left(
\frac{1}{|\mathcal A_c^{\pm}|}
\sum_{a\in\mathcal A_c^{\pm}}
\operatorname{norm}(g_{\ell,a})
\right),
\qquad
b_{\ell,c}
=
\operatorname{norm}\left(
\mu_{\ell,c}^{+}-\mu_{\ell,c}^{-}
\right).
\end{equation}

The subtraction suppresses components shared by uncertainty and certainty anchors and orients $b_{\ell,c}$ toward uncertainty. Collecting the bipolar category directions as $B_\ell=[b_{\ell,1},\ldots,b_{\ell,C}]$, we obtain the orthonormal basis $U_\ell=B_\ell(B_\ell^\top B_\ell)^{-1/2}$. Its columns span the \uspace{} at layer $\ell$ (\Cref{fig:method}a). Although the \uspace{} can be constructed at any layer, we omit the layer subscript $\ell$ hereafter for readability.
This construction requires no task examples, correctness labels, or fitted parameters. It is also not inherently limited to uncertainty: changing only $\mathcal A_c^{\pm}$ yields semantic bases for other properties. \Cref{app:method-generality} provides preliminary evidence, on emotion classification and reward-hacking detection, that the same label-free construction can be applied to domains beyond uncertainty.

\subsection{Reading Uncertainty from the \uspace{}}
\label{sec:ulense}

\paragraph{\ulens{} readout.}
Let the complete sequence processed by the model be
$
(x_{1:N},\,r_{1:T},\,y_{1:M}),
$
where $x_{1:N}$ are the input tokens, $r_{1:T}$ are the generated reasoning tokens, and $y_{1:M}$ are the final-answer tokens, omitting special delimiters for simplicity. We refer to $r=r_{1:T}$ as the reasoning trace. The \ulens{} projects the normalized hidden state of each reasoning token onto the selected \uspace{} basis:
$
\alpha_t
=
U^{\top}
\operatorname{norm}(h_{t})
\in\mathbb{R}^{C}.
$
Each coordinate $\alpha_{t,c}$ measures signed alignment with uncertainty category $c$, and the sequence $\{\alpha_t\}_{t=1}^{T}$ forms an interpretable token-by-category uncertainty map.
We rectify the normalized coordinates into a vector $s_t$ in the positive cone and summarize it by its length:\begin{equation}\label{eq:cone}
s_t
=
\operatorname{softplus}\!\left(\sqrt{C}\,\operatorname{norm}(\alpha_t)\right),
\quad
A_{\mathrm{cone}}(h_{t})
=
\left\|s_t\right\|_2
=
\sqrt{\sum_{c=1}^{C}\log^2\!\left(1+e^{\sqrt{C}\,\operatorname{norm}(\alpha_t)_c}\right)}.
\end{equation}
The score increases with alignment to any positive combination of uncertainty directions, while negative alignment with one category cannot cancel positive alignment with another. It therefore treats the categories as alternative and potentially co-occurring sources of uncertainty. Normalizing \(\alpha_t\) removes the overall strength of the projection and retains only its orientation within \uspace{}. We use \(\operatorname{softplus}\) rather than \(\operatorname{ReLU}\) to preserve graded changes near the boundary of the positive cone instead of abruptly discarding negative alignments. The gain \(\sqrt{C}\) is fixed by dimensionality, since a unit vector spread equally across \(C\) coordinates has entries \(1/\sqrt{C}\), and requires no tuning.

\paragraph{Trace-level uncertainty score.}
Rather than pooling cone alignment across the reasoning tokens, we evaluate it once at the end-of-thinking marker \(t_{\mathrm{eot}}\). %
Because this state follows and is conditioned on the complete reasoning trace, it provides a fixed-position summary without explicitly pooling over a variable number of reasoning tokens. 
We complement this verbalizable signal with first-order distributional uncertainty derived from the model's output probabilities. This separation is motivated by evidence that verbalizable confidence predicts abstention independently of probability-derived confidence, indicating that the two capture distinct information~\citep{kumaran2026causal}. For the distributional component, we use predictive entropy because it captures uncertainty over the full next-token distribution. The entropy chain rule identifies each step's conditional entropy as a natural contribution to uncertainty over an autoregressive sequence~\citep{malinin2021uncertainty}. Let $\mathcal V$ denote the output vocabulary and $p_t(v)=p(v\mid x_{1:N},r_{<t})$ the next-token distribution. Because summed entropy scales mechanically with trace length, we average across the reasoning trace:
\begin{equation}
H_t
=
-\sum_{v\in\mathcal V}p_t(v)\log p_t(v),
\qquad
\overline H(r)
=
\frac{1}{T}\sum_{t=1}^{T}H_t.
\end{equation}
The resulting $\overline H(r)$ provides a length-normalized distributional signal complementary to the verbalizable cone alignment. We combine them multiplicatively so that the verbalizable signal modulates predictive entropy without requiring a fitted relative scale. This also preserves rankings under any global positive rescaling of either factor:
\begin{equation}
\label{eq:components}
S(r)
=
A_{\mathrm{cone}}(h_{t_{\mathrm{eot}}})\,\overline H(r).
\end{equation}

\section{Experiments}
\label{sec:experiments}

We evaluate the \ulens{} as both an uncertainty quantifier and an interpretability method. After introducing the experimental setup, we examine the sensitivity of existing methods to generation length and report uncertainty quantification results. We then analyze the semantic interpretability of the \uspace{} through token-wise readouts and activation steering \citep{subramani-etal-2022-extracting}.

\subsection{Setup}
\label{sec:setup}

\paragraph{Models.}
We evaluate three open-weight reasoning models: \gemma{} (\gemmafull{})~\citep{gemmateam2026gemma4}, \qwen{} (\qwenfull{})~\citep{qwen3.5}, and \magistral{} (\magistralfull)~\citep{rastogi2025magistral}. For \gemma{} and \qwen{}, we use publicly released lens artifacts. No public lens exists for \magistral{}, so we fit it ourselves. \Cref{app:jacobian-construction} describes the fitting procedure and compares the \jlens{} with an alternative lens variant, the \rlens{}~\citep{rlens2026}.
\paragraph{Baselines.}
We compare methods spanning distinct uncertainty signals. As a reference, \textit{\genlen{}} uses the number of generated tokens directly as the uncertainty score. Scores derived from the original generation include Maximum Softmax Probability (\textit{\msp{}}), based on peak softmax confidence~\citep{hendrycks2017baseline}, and \textit{\predent{}} and \textit{\maxent{}}, which summarize next-token entropy~\citep{malinin2021uncertainty,li2024reference}. Mean Negative Log-Likelihood (\textit{Mean NLL}) averages token surprisal~\citep{murray2018correcting}, \textit{\selfcert{}} measures departure from a uniform predictive distribution~\citep{kang2025selfcertainty}, and \textit{\deepconf{}} aggregates local token confidence along the reasoning trace~\citep{fu2026deep}. Multi-pass baselines include \textit{\ptrue{}}, which asks the model to assess its answer in a second pass~\citep{kadavath2022language}, and \textit{\tokur{}}, which estimates token-level uncertainty using five additional passes (one clean, four weight-perturbed)~\citep{zhang2026tokur}. \textit{\sement{}} clusters ten sampled answers by semantic equivalence~\citep{kuhn2023semantic}, while \textit{\sentsar{}} reweights their probabilities by semantic similarity~\citep{duan2024shifting}. \textit{\featgaps{}} fits a supervised direction from contrasts between honest and dishonest representations~\citep{bakman2026uncertainty}. \Cref{app:baselines} provides additional implementation details for every baseline.
\paragraph{Evaluation protocol.}
We extract the uncertainty signal from the layer located two-thirds of the way through the network (layer $40$ for \gemma{}, $42$ for \qwen{}, and
$27$ for \magistral{}), avoiding model- or dataset-specific layer selection. This simple
choice follows evidence that the \jlens{} is most effective after roughly the first third of
the network but before its final layers, which primarily encode the output token~\citep{gurnee2026verbalizable}. 
We ablate layer depth in \cref{app:jacobian-construction}. \featgaps{} is the only supervised baseline and is fitted on held-out labeled data. Across three seeds, we sample one response per item using model-recommended settings and token budgets, excluding truncated or unparseable outputs. Additional evaluation details are provided in \cref{app:experimental-setup}.

\paragraph{Datasets.}
We evaluate on MMLU-Pro~\citep{wang2024mmlupro}, Omni-MATH~\citep{gao2024omnimath}, SuperGPQA~\citep{du2025supergpqa}, and TriviaQA~\citep{joshi2017triviaqa}, covering multi-domain reasoning, olympiad mathematics, graduate-level expertise, and factual knowledge. For each benchmark, we draw a fixed, seeded, stratified sample of 1,000 examples. More details are provided in \Cref{app:experimental-setup}.

\paragraph{Metrics.}
We first report standard AUROC~\citep{hanley1982meaning} and AUPRC \citep{boyd2013area}, which measure how well uncertainty scores distinguish correct from incorrect responses, alongside AURC \citep{geifman2019bias}, which summarizes the error rate among retained responses as increasingly uncertain ones are withheld. Following length-stratified evaluation~\citep{li-etal-2026-confidence,chiu2026probing}, we additionally report length-controlled versions of all three metrics. For each model, dataset, and seed, we divide outputs into ten quantile bins by token count, compute each metric within each bin, and average the results weighted by bin size. These metrics compare outputs of similar length, testing whether an estimator captures predictive information beyond output length.

\begin{table}[!t]
    \centering
    \caption{\textbf{Uncertainty estimation and the effect of length control.}
Results are averaged over four benchmarks, three seeds, and three models (\gemma{}, \qwen{}, and \magistral{}). \(\pm\) denotes standard deviation across models. The final two columns show the change from unadjusted AUROC (\textcolor{unadj}{$\bullet$}) to its length-controlled counterpart (\textcolor{matched}{$\bullet$}) on the AUROC axis shown in the header.} %
    \label{tab:raw-results}
    {\small
\setlength{\tabcolsep}{6pt}
\drawGLline%
\begin{tabular}{l ccc c c}
\toprule
\multirow{2}{*}{Method}
  & \multirow{2}{*}{\NameAuroc~\ArrowAuroc}
  & \multirow{2}{*}{\NameAuprc~\ArrowAuprc}
  & \multirow{2}{*}{\NameAurc~\ArrowAurc}
  & \multicolumn{2}{c}{AUROC under length matching} \\
\cmidrule(lr){5-6}
 & & & & \dropruler & $\Delta$ \\
\midrule
 \gl{\RowGenLength}
  & \gl{68.6\tiny{$\pm$1.4}} & \gl{41.8\tiny{$\pm$7.3}} & \gl{18.0\tiny{$\pm$5.7}} & \dropbarGL{68.6}{52.4} & \gl{\textminus16.2} \\
 \RowMSP
  & 53.3\tiny{$\pm$3.8} & 28.0\tiny{$\pm$10.4} & 23.4\tiny{$\pm$4.9} & \dropbar{53.3}{52.5} & $-0.8$ \\
 \RowMaxEntropy
  & 67.7\tiny{$\pm$3.8} & 40.4\tiny{$\pm$8.0} & 17.6\tiny{$\pm$6.1} & \dropbar{67.7}{62.7} & $-5.0$ \\
 \RowMeanNLL
  & 66.9\tiny{$\pm$6.2} & 41.1\tiny{$\pm$4.8} & 18.3\tiny{$\pm$9.8} & \dropbar{66.9}{65.3} & $-1.6$ \\
 \RowSelfCertainty
  & 66.9\tiny{$\pm$5.7} & 40.8\tiny{$\pm$5.3} & 18.3\tiny{$\pm$9.3} & \dropbar{66.9}{62.5} & $-4.4$ \\
 \RowDeepConf
  & 67.8\tiny{$\pm$6.1} & 41.1\tiny{$\pm$4.8} & 17.6\tiny{$\pm$9.4} & \dropbar{67.8}{66.1} & $-1.8$ \\
 \RowPTrue
  & 56.7\tiny{$\pm$5.0} & 31.0\tiny{$\pm$8.1} & 23.3\tiny{$\pm$9.2} & \dropbar{56.7}{55.6} & $-1.1$ \\
 \RowTokUR
  & 69.6\tiny{$\pm$4.9} & 41.4\tiny{$\pm$5.1} & \textbf{16.2}\tiny{$\pm$5.4} & \dropbar{69.6}{60.2} & $-9.4$ \\
 \sement{}
  & 67.8\tiny{$\pm$1.8} & 42.5\tiny{$\pm$8.1} & 17.2\tiny{$\pm$7.3} & \dropbar{67.8}{66.0} & $-1.8$ \\
 \sentsar{}
  & 63.7\tiny{$\pm$5.3} & 39.0\tiny{$\pm$4.1} & 20.9\tiny{$\pm$9.4} & \dropbar{63.7}{57.9} & $-5.9$ \\
 \RowFeatureGaps
  & 68.8\tiny{$\pm$4.3} & 41.7\tiny{$\pm$2.3} & 16.6\tiny{$\pm$7.0} & \dropbar{68.8}{63.3} & $-5.5$ \\
 \RowPE
  & 67.3\tiny{$\pm$6.1} & 41.5\tiny{$\pm$5.0} & 18.2\tiny{$\pm$9.7} & \dropbar{67.3}{65.9} & $-1.5$ \\
\addlinespace[3pt]   %
 \RowULens
  & \textbf{71.1}\tiny{$\pm$4.7} & \textbf{45.5}\tiny{$\pm$4.2} & \textbf{16.2}\tiny{$\pm$8.3} & \dropbarEnd{71.1}{69.2} & $-1.8$ \\
\bottomrule
\end{tabular}
}

\end{table}

\subsection{Results}
\label{sec:results}

\paragraph{Trace length and correctness.}
Recent work shows that trace length carries useful confidence information, yet uncertainty estimators may exploit it as a shortcut~\citep{devic2025trace,heo2025llms,kang2025selfcertainty}. The unadjusted results in \cref{tab:raw-results} make this effect concrete. Length alone reaches \(68.6\%\) AUROC, nearly matching supervised \featgaps{} (\(68.8\%\)) and trailing five-pass \tokur{} by only \(1.0\) point. %
In the controlled setting, generation length falls to \(52.4\%\) AUROC, and several established estimators decline with it. The table also reveals an instructive compute-performance pattern. In the unadjusted evaluation, five-pass \tokur{} is the strongest conventional baseline in AUROC and AURC, while supervised \featgaps{} leads baseline AUPRC. After controlling for length, however, the additional generations required by several estimators no longer translate into stronger performance: \ptrue{}, \tokur{}, and \sentsar{} fall below predictive entropy in all but one of the nine comparisons, while \sement{} exceeds it by only \(0.1\) points on the cross-model average. In contrast, the \ulens{} drops by only \(1.8\) points and remains the strongest uncertainty predictor, confirming that the signal it captures extends beyond trace length. \Cref{app:length-intervention} analyzes the observed length distributions and complements them with an intervention that holds the question and answer fixed while varying only the reasoning length via task-irrelevant fillers.

\newcommand{\AccGemma}{78.8}\newcommand{\AccSdGemma}{0.2}
\newcommand{\AccQwen}{79.7}\newcommand{\AccSdQwen}{0.3}
\newcommand{\AccMagistral}{65.0}\newcommand{\AccSdMagistral}{0.6}
\newcommand{\SarAuprcLmGemma}{31.2\tiny{$\pm$0.6}}
\newcommand{\SarAuprcLmMagistral}{41.1\tiny{$\pm$0.6}}
\newcommand{\SarAuprcLmQwen}{34.3\tiny{$\pm$1.4}}
\newcommand{\SarAurcLmGemma}{18.7\tiny{$\pm$0.4}}
\newcommand{\SarAurcLmMagistral}{34.3\tiny{$\pm$0.7}}
\newcommand{\SarAurcLmQwen}{16.4\tiny{$\pm$0.2}}
\newcommand{\SarAurocLmGemma}{58.3\tiny{$\pm$0.5}}
\newcommand{\SarAurocLmMagistral}{53.1\tiny{$\pm$0.3}}
\newcommand{\SarAurocLmQwen}{62.2\tiny{$\pm$0.8}}
\newcommand{\SemEntAuprcLmGemma}{37.2\tiny{$\pm$1.3}}
\newcommand{\SemEntAuprcLmMagistral}{51.1\tiny{$\pm$0.6}}
\newcommand{\SemEntAuprcLmQwen}{40.2\tiny{$\pm$0.4}}
\newcommand{\SemEntAurcLmGemma}{15.1\tiny{$\pm$0.4}}
\newcommand{\SemEntAurcLmMagistral}{28.4\tiny{$\pm$0.2}}
\newcommand{\SemEntAurcLmQwen}{13.3\tiny{$\pm$0.3}}
\newcommand{\SemEntAurocLmGemma}{65.8\tiny{$\pm$1.4}}
\newcommand{\SemEntAurocLmMagistral}{63.5\tiny{$\pm$0.2}}
\newcommand{\SemEntAurocLmQwen}{68.6\tiny{$\pm$0.8}}
\begin{table}[!t]
\centering
\small
\caption{%
  \textbf{Length-controlled results.} Scores averaged over \DsLabelMmlupro,
  \DsLabelOmnimath, \DsLabelSupergpqa{} and \DsLabelTriviaqa. 
  All values are percentages.  Bracketed after each model is
  its average accuracy across the four benchmarks. Each
  cell is the mean over three seeds. Best per column in \textbf{bold}.
}
\label{tab:main-lm}
\setlength{\tabcolsep}{3pt}
\resizebox{\ifdim\width>\textwidth\textwidth\else\width\fi}{!}{%
\begin{tabular}{l ccc | ccc | ccc}
\toprule
\multirow{2}{*}{Method} & \multicolumn{3}{c|}{\gemma\ (acc.\ \AccGemma\,$\pm$\,\AccSdGemma)} & \multicolumn{3}{c|}{\qwen\ (acc.\ \AccQwen\,$\pm$\,\AccSdQwen)} & \multicolumn{3}{c}{\magistral\ (acc.\ \AccMagistral\,$\pm$\,\AccSdMagistral)} \\
\cmidrule(lr){2-4} \cmidrule(lr){5-7} \cmidrule(lr){8-10}

 & \NameAuroc~\ArrowAuroc & \NameAuprc~\ArrowAuprc & \NameAurc~\ArrowAurc
 & \NameAuroc~\ArrowAuroc & \NameAuprc~\ArrowAuprc & \NameAurc~\ArrowAurc
 & \NameAuroc~\ArrowAuroc & \NameAuprc~\ArrowAuprc & \NameAurc~\ArrowAurc \\
\midrule
\RowGenLength
 & \CellAurocLmGenLengthGemma & \CellAuprcLmGenLengthGemma & \CellAurcLmGenLengthGemma
 & \CellAurocLmGenLengthQwen & \CellAuprcLmGenLengthQwen & \CellAurcLmGenLengthQwen
 & \CellAurocLmGenLengthMagistral & \CellAuprcLmGenLengthMagistral & \CellAurcLmGenLengthMagistral \\
\RowMSP
 & \CellAurocLmMSPGemma & \CellAuprcLmMSPGemma & \CellAurcLmMSPGemma
 & \CellAurocLmMSPQwen & \CellAuprcLmMSPQwen & \CellAurcLmMSPQwen
 & \CellAurocLmMSPMagistral & \CellAuprcLmMSPMagistral & \CellAurcLmMSPMagistral \\
\RowMaxEntropy
 & \CellAurocLmMaxEntropyGemma & \CellAuprcLmMaxEntropyGemma & \CellAurcLmMaxEntropyGemma
 & \CellAurocLmMaxEntropyQwen & \CellAuprcLmMaxEntropyQwen & \CellAurcLmMaxEntropyQwen
 & \CellAurocLmMaxEntropyMagistral & \CellAuprcLmMaxEntropyMagistral & \CellAurcLmMaxEntropyMagistral \\
\RowMeanNLL
 & \CellAurocLmMeanNLLGemma & \CellAuprcLmMeanNLLGemma & \CellAurcLmMeanNLLGemma
 & \CellAurocLmMeanNLLQwen & \CellAuprcLmMeanNLLQwen & \CellAurcLmMeanNLLQwen
 & \CellAurocLmMeanNLLMagistral & \CellAuprcLmMeanNLLMagistral & \CellAurcLmMeanNLLMagistral \\
\RowSelfCertainty
 & \CellAurocLmSelfCertaintyGemma & \CellAuprcLmSelfCertaintyGemma & \CellAurcLmSelfCertaintyGemma
 & \CellAurocLmSelfCertaintyQwen & \CellAuprcLmSelfCertaintyQwen & \CellAurcLmSelfCertaintyQwen
 & \CellAurocLmSelfCertaintyMagistral & \CellAuprcLmSelfCertaintyMagistral & \CellAurcLmSelfCertaintyMagistral \\
\RowDeepConf
 & \CellAurocLmDeepConfGemma & \CellAuprcLmDeepConfGemma & \CellAurcLmDeepConfGemma
 & \CellAurocLmDeepConfQwen & \CellAuprcLmDeepConfQwen & \CellAurcLmDeepConfQwen
 & \CellAurocLmDeepConfMagistral & \CellAuprcLmDeepConfMagistral & \CellAurcLmDeepConfMagistral \\
\RowPTrue
 & \CellAurocLmPTrueGemma & \CellAuprcLmPTrueGemma & \CellAurcLmPTrueGemma
 & \CellAurocLmPTrueQwen & \CellAuprcLmPTrueQwen & \CellAurcLmPTrueQwen
 & \CellAurocLmPTrueMagistral & \CellAuprcLmPTrueMagistral & \CellAurcLmPTrueMagistral \\
\RowTokUR
 & \CellAurocLmTokURGemma & \CellAuprcLmTokURGemma & \CellAurcLmTokURGemma
 & \CellAurocLmTokURQwen & \CellAuprcLmTokURQwen & \CellAurcLmTokURQwen
 & \CellAurocLmTokURMagistral & \CellAuprcLmTokURMagistral & \CellAurcLmTokURMagistral \\
\sement{}
 & \SemEntAurocLmGemma & \SemEntAuprcLmGemma & \SemEntAurcLmGemma
 & \SemEntAurocLmQwen & \SemEntAuprcLmQwen & \SemEntAurcLmQwen
 & \SemEntAurocLmMagistral & \SemEntAuprcLmMagistral & \SemEntAurcLmMagistral \\
\sentsar{}
 & \SarAurocLmGemma & \SarAuprcLmGemma & \SarAurcLmGemma
 & \SarAurocLmQwen & \SarAuprcLmQwen & \SarAurcLmQwen
 & \SarAurocLmMagistral & \SarAuprcLmMagistral & \SarAurcLmMagistral \\
\RowFeatureGaps
 & \CellAurocLmFeatureGapsGemma & \CellAuprcLmFeatureGapsGemma & \CellAurcLmFeatureGapsGemma
 & \CellAurocLmFeatureGapsQwen & \CellAuprcLmFeatureGapsQwen & \CellAurcLmFeatureGapsQwen
 & \CellAurocLmFeatureGapsMagistral & \CellAuprcLmFeatureGapsMagistral & \CellAurcLmFeatureGapsMagistral \\
\RowPE
 & \CellAurocLmPEGemma & \CellAuprcLmPEGemma & \CellAurcLmPEGemma
 & \CellAurocLmPEQwen & \CellAuprcLmPEQwen & \CellAurcLmPEQwen
 & \CellAurocLmPEMagistral & \CellAuprcLmPEMagistral & \CellAurcLmPEMagistral \\
\addlinespace[3pt]   %
\RowACone
 & \CellAurocLmAConeGemma & \CellAuprcLmAConeGemma & \CellAurcLmAConeGemma
 & \CellAurocLmAConeQwen & \CellAuprcLmAConeQwen & \CellAurcLmAConeQwen
 & \CellAurocLmAConeMagistral & \CellAuprcLmAConeMagistral & \CellAurcLmAConeMagistral \\
\RowULens
 & \CellAurocLmULensGemma & \CellAuprcLmULensGemma & \CellAurcLmULensGemma
 & \CellAurocLmULensQwen & \CellAuprcLmULensQwen & \CellAurcLmULensQwen
 & \CellAurocLmULensMagistral & \CellAuprcLmULensMagistral & \CellAurcLmULensMagistral \\
\bottomrule
\end{tabular}%
}%
\end{table}

\paragraph{Length-controlled results.}
The \ulens{} is the only method to lead all nine model-metric comparisons in \cref{tab:main-lm}. Relative to the strongest baseline, it improves AUROC by \(2.2\), \(1.9\), and \(4.7\) points on \gemma{}, \qwen{}, and \magistral{}, respectively. Its advantage extends to the error-focused metrics. Across models, it improves AUPRC by \(0.8\) to \(4.2\) points and reduces AURC by \(0.4\) to \(2.5\) points.
This consistency arises from the complementary first- and second-order signals combined in \cref{eq:components}. On \gemma{}, predictive entropy and \(A_{\mathrm{cone}}\) trade the advantage across metrics. Predictive entropy is stronger throughout on \qwen{}, whereas \(A_{\mathrm{cone}}\) is stronger throughout on \magistral{}. Notably, their combination in the \ulens{}  improves consistently over both components, providing evidence that the semantic readout contributes beyond first-order token entropy.

\subsection{Reading and Steering Uncertainty in the \uspace{}}
\label{sec:readout}
\begin{figure}[t]
  
  \centering
  \includegraphics[width=\linewidth]{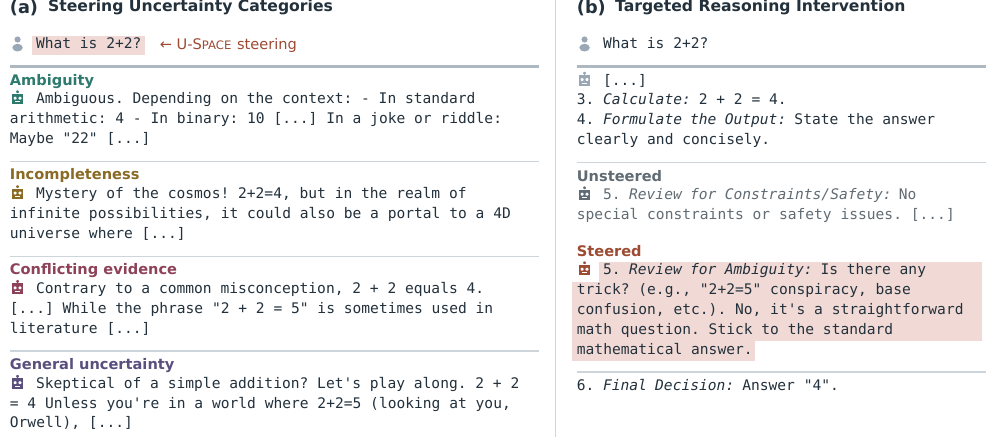}
  \caption{\textbf{Steering uncertainty.} \textbf{(a)} Prompt-level steering, applied separately along each of the four category directions during prefill. Additional examples appear in \cref{app:causal_intervention}. \textbf{(b)} Steering along \catamb{} during reasoning. Highlighting marks the tokens whose hidden states are modified.}
  \label{fig:causal_main}
\end{figure}

Beyond the empirical evaluation of the \uspace{}, we investigate the expressiveness of the \ulens{} as a tool for understanding the uncertainty of a model through mechanistic interpretability. Additionally, we show that the \uspace{} is not just a readout of the model's uncertainty but can be used to actively induce uncertainty in the model through activation steering \citep{subramani-etal-2022-extracting}. 
\begin{figure}[t]
  \centering
  \includegraphics[width=\textwidth]{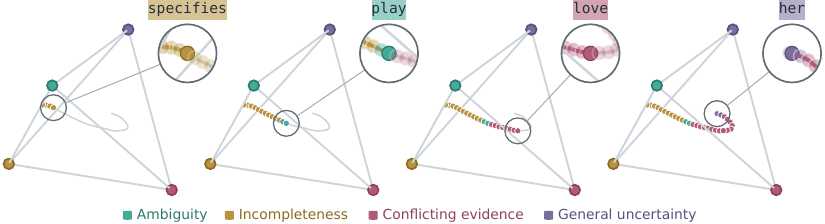}
  \caption{\textbf{One answer's path through the simplex.} \qwen{} answers \emph{``Was Hamlet left-handed, and did he love Ophelia?''}. Each answer token is shaded by the category it is read as: \catshade{catinc}{No, William Shakespeare never specifies Hamlet's handedness in the} \catshade{catamb}{play}\catshade{catcon}{. Yes, Hamlet did genuinely love Ophelia, though his madness and his father's death caused him to cruelly} \catshade{catgen}{reject her.} Panels from left to right draw the path up to the magnified token (\emph{specifies}, \emph{play}, \emph{love}, \emph{her}). %
  }
  \label{fig:simplex_path}
\end{figure}
\paragraph{Interpreting uncertainty through the \ulens{}.}
In \cref{tab:main-lm}, we established that the \ulens{} can read a scalar uncertainty estimate from the end-of-thinking marker, but the construction of the \uspace{} allows for a more detailed, token-level readout that unfolds a differentiated view of the model's evolving uncertainty across the reasoning trace (\Cref{fig:method}b).
The token-level readout evaluates $s_t$ as defined in \cref{eq:cone} at every reasoning position, rather than only at $t_{\mathrm{eot}}$. For each token, this yields one strictly positive score per uncertainty category.
Because the entries are positive, they can be read as shares, $s_t/\sum_c s_{t,c}$, which places each token as a point in a $3$-simplex whose four vertices are the categories. A token near a vertex is read almost entirely as that category, while tokens near the center are read as an equal share. A reasoning trace is the path its tokens describe inside this simplex. While this allows for a calibration-free reading of which kinds of uncertainty are present and how their mix evolves (\Cref{fig:hamlet_teaser,fig:simplex_path}), it does not indicate how much uncertainty is present. Individually evaluating the dimensions of $s_t$ provides quantitative scores for each uncertainty category that are bounded by construction (\Cref{app:mechanistic-interpretability} for details and examples). 
\Cref{fig:simplex_path} follows a single answer along such a path. While the model states what Shakespeare left unsaid, its tokens exhibit \catinc{}. The reading tips to \catamb{} at \emph{play}, turns to \catcon{} as the answer affirms Hamlet's love against his cruelty towards Ophelia, and ends in \catgen{} at the closing \emph{her}. %
\paragraph{Steering uncertainty in the \uspace{}.}
The token-level readout links contextual uncertainty to the model's verbalizable workspace~\citep{dehaene1998neuronal}, but does not establish whether these representations influence generation. We test their causal role by steering residual states along one category direction at a time.
Following \citet{turner2023actadd}, we add a category direction to the residual stream over a
contiguous span $\mathcal{W}$ of token positions and layers $\mathcal{L}$ around
the readout depth,
\begin{equation}
  \tilde h_{\ell,t}
  \;=\;
  h_{\ell,t} \;+\; \lambda \,\lVert h_{\ell,t} \rVert_2 \, u_{\ell,c},
  \qquad \ell \in \mathcal{L},\; t \in \mathcal{W},
  \label{eq:inject_main}
\end{equation}
where $u_{\ell,c}$ is the column of $U_\ell$ for category $c$ and the dose $\lambda$ is a fraction of the state's own norm, so that one value of $\lambda$ is comparable across positions, depths, and models.
Because steering is applied during autoregressive generation, its effect propagates to later tokens. The span $\mathcal W$ can cover the prompt during prefill or selected reasoning positions.
In \Cref{fig:causal_main}a, we steer only the prompt, applying each of the four category directions separately before continuing generation. \Cref{fig:causal_main}b applies the \catamb{} direction during reasoning and causes the model to lose confidence in the targeted span.~\Cref{app:causal_intervention} provides the full protocol and additional results across varying values of $\lambda$.

\section{Conclusion}
\label{sec:conclusion}

Uncertainty estimates are more useful when they reveal not only whether an answer may be wrong, but where uncertainty enters the reasoning. We introduced the \uspace{}, an interpretable subspace representing distinct sources of uncertainty, and the \ulens{}, a readout that follows these signals through reasoning without requiring correctness labels or repeated sampling. Across the evaluated models and benchmarks, this score improves on the compared estimators and remains strong under length control. Its trace-wise readouts and targeted interventions show that the identified directions are interpretable and behaviorally relevant. Preliminary evidence further suggests that the underlying construction extends to semantic domains beyond uncertainty. Together, these results point toward uncertainty estimates that support error detection and more transparent model auditing.

\section*{Ethics Statement}
\label{sec:ethics}
Our study uses public reasoning benchmarks and open-weight models. It involves no human participants or collection of private user data. The proposed readout could help flag uncertain answers for review, but it is not a calibrated safety detector and should not replace independent verification in consequential settings. Because interventions along the same directions can alter model commitment, any deployment should evaluate both reliability and potential misuse.

\section*{Reproducibility Statement}
\label{sec:reproducibility}
Code to reproduce our main results (\Cref{tab:main-lm}, \Cref{tab:main-raw}) and the comparisons across all models and benchmarks is available at \url{https://github.com/s2labres/U-Space}. After installing the pinned environment (\texttt{requirements.txt}), \texttt{run.sh} runs the full pipeline.

\section*{Acknowledgements}
The research was partially funded by a LOEWE-Spitzen-Professur (LOEWE/4a//519/05.00.002-(0010)/93). Additionally, the research was partially funded by an Alexander von Humboldt Professorship in Multimodal Reliable AI, sponsored by the Federal Ministry of Research, Technology, and Space (BMFTR). For compute, we gratefully acknowledge support from the hessian.AI Service Center (funded by the Federal Ministry of Research, Technology and Space (BMFTR), grant no. 16IS22091) and the hessian.AI Innovation Lab (funded by the Hessian Ministry for Digital Strategy and Innovation, grant no. S-DIW04/0013/003). Tobias Braun acknowledges support from ELSA (European Lighthouse on Secure and Safe AI), funded by the European Union under grant agreement No.~101070617. The views expressed are those of the authors and do not necessarily represent those of the European Union or the granting authority. Neither can be held responsible for them. Lorenzo Cavallaro was partially supported by a Google Academics Research Award.

\bibliographystyle{plainnat}
\bibliography{references}
\appendix
\crefalias{section}{appendix}
\crefalias{subsection}{appendix}
\clearpage
\suppressfloats[t]
\begin{center}
  \rule{\linewidth}{1pt}\\[0.55em]
  {\LARGE\sc Appendix}\\[0.5em]
  {\normalsize\itshape Supplementary material for ``\methodLong''}\\[0.6em]
  \rule{\linewidth}{0.4pt}
\end{center}
\vspace{1.2em}
The appendix is organized as follows:
\begin{itemize}
  \setlength{\itemsep}{0.35em}
  \item \Cref{app:jacobian-construction}: where each mean Jacobian comes from, how the
    readout depth is fixed, and ablations over depth and against the \rlens{}~\citep{rlens2026}.
  \item \Cref{app:anchor-construction}: the uncertainty and certainty anchor sets, and a
    lexical baseline that scores a trace by counting them.
  \item \Cref{app:decodability}: what contrastive activation differences decode to, as
    evidence that the categories are the model's and not only ours.
  \item \Cref{app:method-generality}: the same construction with emotion and reward-hacking
    poles, as a test of how far it carries beyond uncertainty.
  \item \Cref{app:experimental-setup}: checkpoints, datasets, prompts, generation settings,
    the scoring protocol, and how every baseline is implemented.
  \item \Cref{app:results}: trace-length analyses, per-model unadjusted results, cross-benchmark
    transfer of the supervised methods, and what each method costs to run.
  \item \Cref{app:mechanistic-interpretability}: the readout token by token on single
    answers.
  \item \Cref{app:causal_intervention}: the steering protocol, and what injecting or
    ablating a direction does.
\end{itemize}

\section{Jacobian Construction and Readout Depth}
\label{app:jacobian-construction}
Verbalizing the hidden state of an LLM requires a mechanism that maps a residual-stream vector to tokens. The logit lens~\citep{nostalgebraist2020logitlens} applies the unembedding matrix $W_U$ directly to an intermediate hidden state, which implicitly approximates the layers between that state and the unembedding as the identity. \citet{gurnee2026verbalizable} replace this identity with a mean Jacobian.
\paragraph{Mean Jacobian.}
A prompt-specific Jacobian reflects both general representational structure and the particular context in which it is evaluated. Following~\citet{gurnee2026verbalizable}, we instead use a mean Jacobian that averages the effect of an intermediate residual representation on current and future final-layer representations. Let \(h_{\ell,t}^{(x)}\) and \(h_{L,t'}^{(x)}\) denote the residual representations at source layer \(\ell\) and final layer \(L\), at positions $t$ and $t'$, for prompt \(x\). The layer-\(\ell\) matrix is
\[
\bar J_\ell
=
\mathbb{E}_{x\sim\mathcal R,\,t,\,t'\geq t}
\left[
\frac{\partial h_{L,t'}^{(x)}}
     {\partial h_{\ell,t}^{(x)}}
\right],
\]
where \(\mathcal R\) is a corpus of neutral, pretraining-like prompts. The estimator averages over prompts and source positions while aggregating the cotangents of current and future target positions.

Composing \(\bar J_\ell\) with the unembedding matrix yields one residual-stream direction per vocabulary token. For anchor \(a\), represented by vocabulary basis vector \(e_a\), we use
\[
g_{\ell,a}
=
(W_U\operatorname{diag}(\gain)\bar J_\ell)^\top e_a
=
\bar J_\ell^\top \operatorname{diag}(\gain) W_U^\top e_a,
\]
the notation of \Cref{sec:uspace}, where \(\gain\) is the final RMSNorm gain. Equivalently, \(g_{\ell,a}^\top\) is the row of \(W_U\operatorname{diag}(\gain)\bar J_\ell\) associated with anchor \(a\). The gain is not absorbed by the \(\ell_2\)-normalization that follows: \(\operatorname{diag}(\gain)\) rescales each coordinate separately, so dropping it would change the direction and not merely its length. Averaging the Jacobian across generic contexts makes this direction reflect the model's general disposition to verbalize the anchor rather than its role in a particular prompt.
\paragraph{Artifacts and one-time estimation.}
The mean Jacobian is specific to a model checkpoint but independent of our downstream tasks and benchmarks. For \gemma{} and \qwen{}, we use the publicly released lenses\footnote{\url{https://huggingface.co/neuronpedia/jacobian-lens/tree/0731326edff4ae730ffc5356fe1a4728c748b3a6}}, both fitted on WikiText. No public lens exists for \magistral{}, so we fit it with the reference implementation\footnote{\url{https://github.com/anthropics/jacobian-lens/commit/581d398613e5602a5af361e1c34d3a92ea82ba8e}} on 100 prompts from WikiText-103\footnote{\url{https://huggingface.co/datasets/Salesforce/wikitext/tree/b08601e04326c79dfdd32d625aee71d232d685c3/wikitext-103-raw-v1}} at sequence length 128. Estimating a lens runs unlabeled prompts through the frozen model and backpropagates from the target residual stream. No parameters are optimized or updated, which distinguishes the procedure from probe training or task-specific calibration. \citet{gurnee2026verbalizable} report that a lens estimated from ten prompts already outperforms established alternatives and that additional prompts give diminishing improvements. The reference implementation suggests roughly 100 prompts and notes that the cost is dominated by the model's backward passes.

\begin{figure}[htbp]
  \centering
  \includegraphics[width=\linewidth]{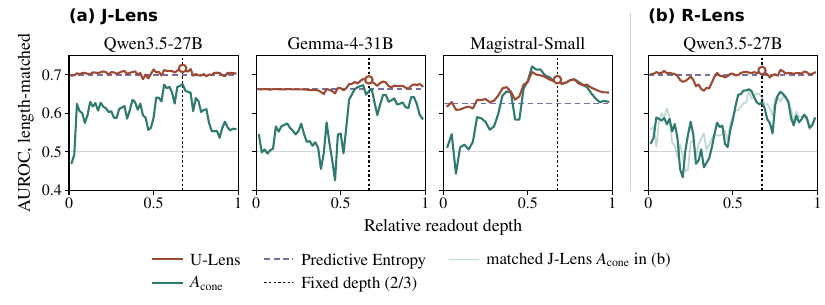}
  \caption{\textbf{Readout depth and lens ablation.} \textbf{(a)} Readout depth ablation across all model layers. \textbf{(b)} Lens comparison between \jlens{} and \rlens{} fitted on the identical data by~\citet{rlens2026}. Both panels are read from the seed 42 rather than averaged over the three generation seeds.}
  \label{fig:depth_ablation}
\end{figure}

\paragraph{Readout depth.}
The readout layer $\ell^\star$ is fixed a priori at two-thirds of the depth for every model and experiment (layers \BlockGemma{} of 60, \BlockQwen{} of 64, and \BlockMagistral{} of 40), following the observation of \citet{gurnee2026verbalizable} that the \jlens{} reads best after the first third of the network and before the final layers, which mostly encode the output token. \Cref{fig:depth_ablation}(a) reports length-matched AUROC for each layer in the model, macro-averaged over the four benchmarks, for $A_{\mathrm{cone}}$ alone and for the \ulens{}, with the basis transported to each layer by the lens. $A_{\mathrm{cone}}$ varies substantially across depth, with the best performance in the mid-to-late layers.
\paragraph{Lens ablation.}
\citet{rlens2026} propose the \rlens{}, which modifies the backward pass to limit error accumulation and reads earlier layers more cleanly. \Cref{fig:depth_ablation}(b) sweeps \qwen{} through their released \rlens{} and, drawn faint, through the \jlens{} released alongside it, which is fitted on the same data and differs only in the backward graph. The two $A_{\mathrm{cone}}$ curves strongly correlate and show no improvement in the early layers for our task. 
\section{Anchor Words}
\label{app:anchor-construction}
Constructing the \uspace{} requires a positive and a negative anchor set per category. We use two published word lists. The uncertainty anchors \(\mathcal A^+\) are the 61 seed cues that \citet{chen2018scalable} collected as markers of uncertainty in text. The certainty anchors \(\mathcal A^-\) are the 348 single-word entries of the certainty norm of \citet{rocklage2023beyond} whose ratings lie at or above the corpus mean. The first list supplies vocabulary of doubt that a model may verbalize, while the second supplies the contrasting register of confidence. Their subtraction in \Cref{eq:diff-construction} removes shared components and retains the direction that separates them.

To sort the words into the four categories, we embed each word and a one-line definition of each category with a sentence encoder~\citep{li2023angle} that is independent of the evaluated models, z-score each word's similarities to the four definitions across the list, and assign the word to the category with the highest z-score. The definitions used for semantic matching are:
\begin{description}
    \item[\Catamb{}:] Unclear, multiply interpretable, confusing, or insufficiently determinate content.
    \item[\Catinc{}:] Missing, unavailable, unresolved, or otherwise absent knowledge or information.
    \item[\Catcon{}:] Mutually incompatible claims, findings, observations, sources, or interpretations.
    \item[\Catgen{}:] Reduced confidence in a proposition, including doubt, tentative support, low plausibility, unreliability, or hedging.
\end{description}
This yields 13, 14, 17, and 17 positive anchors and 82, 84, 84, and 98 negative anchors for \catamb{}, \catinc{}, \catcon{}, and \catgen{}, respectively. For each model, we then drop words that its tokenizer does not encode as a single token. The counts above are before this step, while \Cref{tab:anchor-words} reports them after it.

\Cref{tab:anchor-words} shows, for each category and side, the ten anchors whose pulled-back directions at the readout layer align best with their side's centroid, averaged over the three models. The uncertain side is dominated by the expected vocabulary of each category. The certain side combines explicit certainty terms such as \emph{doubtless}, \emph{unquestionably}, and \emph{unequivocally} with words that mark a confident register, such as \emph{transpired}, \emph{eradicated}, and \emph{dignified}. The norm thus captures certainty as a manner of speaking, the shared component that the bipolar subtraction is designed to remove.
\begin{table}[t]
\centering
\small
\caption{\textbf{Anchor words of the \uspace{}.} Positive (uncertainty) anchors are from \citet{chen2018scalable} and negative (certainty) anchors are from \citet{rocklage2023beyond}. Shown are the ten words per side whose pulled-back directions align best with their side's centroid at the readout block, averaged over the three models. $n$ is the number of words per anchor set that form a single token for \gemma{} (the text gives the unfiltered counts).}
\label{tab:anchor-words}
\resizebox{\ifdim\width>\textwidth\textwidth\else\width\fi}{!}{%
\begin{tabular}{@{}p{0.16\textwidth}p{0.41\textwidth}p{0.41\textwidth}@{}}
\toprule
\textbf{Category} & \textbf{Positive anchors $\mathcal A^+_c$ (uncertainty)} & \textbf{Negative anchors $\mathcal A^-_c$ (certainty)} \\
\midrule
\parbox[t]{0.16\textwidth}{\raggedright\hyphenpenalty=10000 \Catamb{}} & \textit{(n=12)} ambiguous, incoherent, puzzling, baffling, ambiguity, unclear, confusing, unpredictable, bizarre, unexpected & \textit{(n=74)} invariable, differentiates, doubtless, clarifying, inescapable, instructing, unchanging, demonstrable, transpired, classifies \\
\addlinespace
\parbox[t]{0.16\textwidth}{\raggedright\hyphenpenalty=10000 \Catinc{}} & \textit{(n=13)} mysterious, undetermined, unexplained, mystery, mysteries, unknown, unidentified, inconceivable, uncharted, unrecognized & \textit{(n=79)} unquestionably, exempted, guaranteeing, persistently, inspecting, superseded, determined, undisputed, accounted, complete \\
\addlinespace
\parbox[t]{0.16\textwidth}{\raggedright\hyphenpenalty=10000 \Catcon{}} & \textit{(n=15)} inconclusive, contradictory, paradoxical, conflicting, inconsistent, fallacy, contentious, controversial, contrary, incompatible & \textit{(n=77)} evidences, yep, unambiguously, substantiated, conclusive, seconded, totality, eradicated, deduced, quantified \\
\addlinespace
\parbox[t]{0.16\textwidth}{\raggedright\hyphenpenalty=10000 \Catgen{}} & \textit{(n=14)} debatable, dubious, suspicion, doubtful, uncertain, uncertainty, improbable, deceptive, suspect, unreliable & \textit{(n=95)} predicated, affirming, emphatic, resolute, confirmations, dignified, unequivocally, ceases, decisively, solidified \\
\addlinespace
\bottomrule
\end{tabular}%
}%
\end{table}

\paragraph{Lexical baseline.}
To test whether the anchor tokens are already sufficient as lexical cues, we score each thinking block by the number of distinct words from \(\mathcal A^+\) that it contains and by this count minus the number of distinct words from \(\mathcal A^-\). Under length-matched evaluation, the \(\mathcal A^+\) score remains well below \(A_{\mathrm{cone}}\), while subtracting \(\mathcal A^-\) drives the keyword score below chance. Without length matching, \(\mathcal A^+\) reaches \(55.9\%\), \(60.4\%\), and \(63.4\%\) AUROC on \gemma{}, \qwen{}, and \magistral{}, respectively, but still trails generation length. The certainty anchors are thus useful as a direction to subtract, not as surface vocabulary to match. The cue words are also rare: they occur in \(23\%\)--\(34\%\) of \gemma{} traces overall and in only \(4\%\) on TriviaQA. The \ulens{} does not depend on their presence. On the \(77\%\) of \gemma{} traces without a cue word in the thinking block, it reaches \(68.6\%\) AUROC, essentially unchanged from the \(68.4\%\) it reaches on the full set (\Cref{tab:main-lm}). On \qwen{}, it also scores higher on cue-free traces than on all traces.

\begin{table}[h]
\centering
\small
\caption{\textbf{Anchor words as a keyword score.} Each trace's thinking block is scored by the number of distinct words of $\mathcal A^{+}$ it contains, or by that number minus the number of distinct words of $\mathcal A^{-}$. AUROC $\times 100$, unadjusted and length-matched.}
\label{tab:lexical-anchor}
\begin{tabular}{l l cc ccc}
\toprule
\multirow{2}{*}{Model} & \multirow{2}{*}{Metric} & \multicolumn{2}{c}{Keyword score} & \multicolumn{3}{c}{Reference} \\
\cmidrule(lr){3-4} \cmidrule(lr){5-7}
 & & $\mathcal A^{+}$ & $\mathcal A^{+} - \mathcal A^{-}$ & Length & $A_{\mathrm{cone}}$ & \ulens{} \\
\midrule
\multirow{2}{*}{\gemma{}} & Unadjusted & 55.9 & 35.9 & 70.1 & 73.1 & 73.2 \\
 & Length-matched & 52.1 & 41.7 & 52.1 & 66.3 & 68.4 \\
\multirow{2}{*}{\qwen{}} & Unadjusted & 60.4 & 33.1 & 67.7 & 71.5 & 74.3 \\
 & Length-matched & 54.4 & 41.6 & 52.7 & 66.7 & 71.0 \\
\multirow{2}{*}{\magistral{}} & Unadjusted & 63.4 & 42.3 & 67.8 & 68.5 & 65.7 \\
 & Length-matched & 58.6 & 49.3 & 52.3 & 67.6 & 68.3 \\
\bottomrule
\end{tabular}
\end{table}

\section{Interpretability of Contrastive Activation Differences}
\label{app:decodability}
\begin{figure}[htbp]
  \centering
  \includegraphics[width=\linewidth]{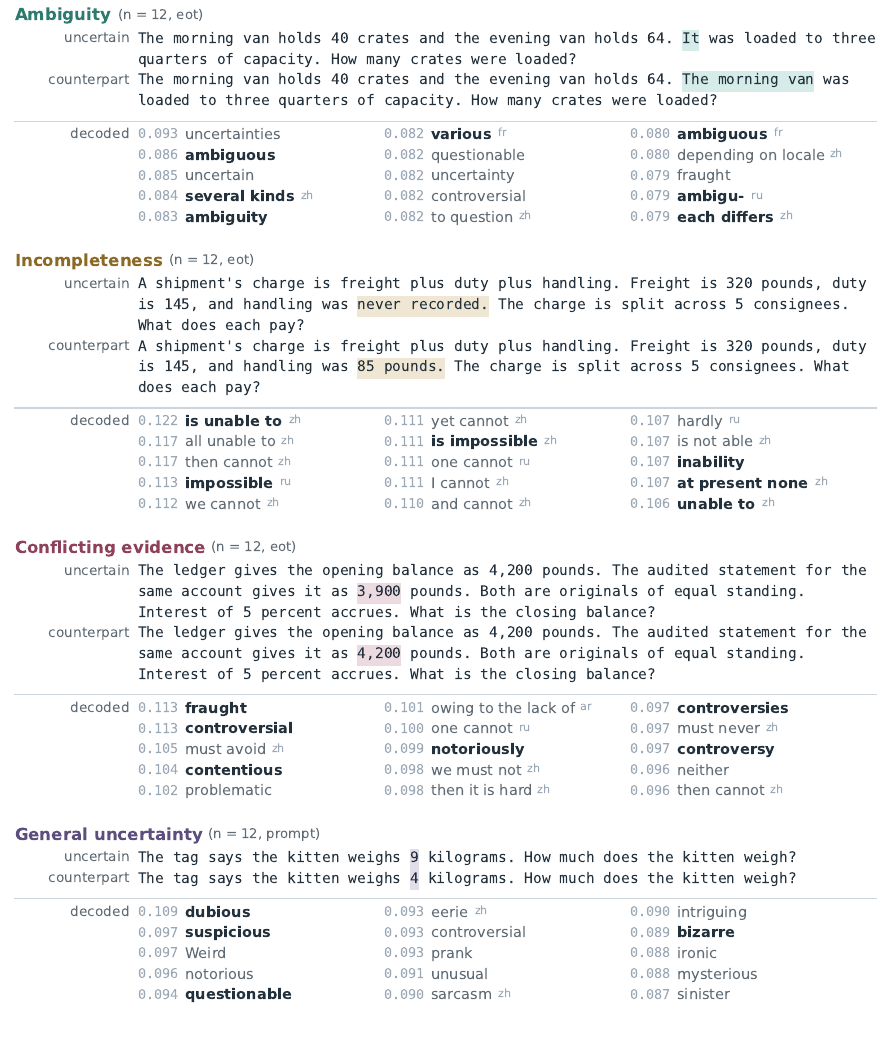}
  \caption{\textbf{Difference-in-means readout.} Representative content-matched prompt pairs for each uncertainty category and top-15 tokens after decoding the difference-in-means direction through the \jlens{} on \qwen{}. Highlighted tokens reflect manually matched tokens to the specific content for clarity. Non-English tokens were translated. The acronyms behind the tokens reflect the original language of the token.}
  \label{fig:concept_decode_axes}
\end{figure}

\Cref{app:anchor-construction} fixes the vocabulary of the \uspace{} from published word lists, so the basis could in principle be a lexicon we impose rather than a structure the model has. This section tests the converse direction. Prior work shows that contrastive activation differences can isolate semantically meaningful directions in the residual stream~\citep{turner2023actadd,marks2024geometry}. We construct minimal prompt pairs that induce \catamb{}, \catinc{}, \catcon{}, or \catgen{} without containing explicit uncertainty vocabulary. We then decode each category's difference-in-means direction through the \jlens{} to determine which terms it recovers and whether they align with the corresponding entries in the published word lists documented in \Cref{app:anchor-construction}.
\paragraph{Content-matched prompt pairs.}
To induce a specific type of uncertainty, we construct semantically and lexically matched prompt pairs that differ only in whether the relevant issue is present. One prompt instantiates the category, whereas the other resolves it while leaving the rest of the problem unchanged, for example, by disambiguating a reference, providing a missing value, or replacing disagreeing sources with agreeing ones. \Cref{fig:concept_decode_axes} shows one pair per category. In total, we construct twelve pairs per category. To prevent explicit stimuli from leaking into the decoded tokens, we ensure that the prompts do not contain the epistemic vocabulary we are looking for.

\paragraph{Decoding diff-in-means.}
For each category $c$ we collect one residual state per prompt at the targeted block and form
\begin{equation}
\delta_c
=
\frac{1}{|P_c|}\sum_{p\in P_c}\bigl(h^{+}_{p}-h^{-}_{p}\bigr),
\end{equation}
the mean over the pairs $P_c$ of the difference between the states of the uncertain and the resolved member. We score every vocabulary item $v$ by the cosine between $\operatorname{norm}(\delta_c)$ and its pulled-back direction $\operatorname{norm}(g_{\ell,v})$ of \Cref{sec:uspace}, and report the highest-scoring items. The only post-processing is a lexical filter that retains tokens of two or more letters.
\paragraph{Results.}
\Cref{fig:concept_decode_axes} shows the fifteen highest-scoring items per direction. \Catamb{} promotes \emph{ambiguous} and \emph{ambiguity} at ranks 2 and 5 of $248{,}320$ items, alongside multiplicity terms in several languages. \Catcon{} promotes \emph{controversial} and \emph{contentious} at ranks 2 and 4, followed by \emph{controversies} and \emph{controversy}, and prohibition terms (\emph{must avoid}, \emph{must never}) that appear in no other list. \Catinc{} is dominated by inability across scripts (\emph{cannot}, \emph{unable to}, \emph{impossible}), with the English \emph{inability} at rank 13. The pairwise cosines between the \catamb{}, \catcon{}, and \catinc{} directions are $0.51$, $0.60$, and $0.65$. \Catgen{} promotes \emph{dubious}, \emph{suspicious}, and \emph{questionable} at ranks 1, 2, and 5, and otherwise the vocabulary of an implausible claim, \emph{weird}, \emph{unusual}, \emph{bizarre}, \emph{prank}. Of these fifteen tokens, only \emph{questionable}, \emph{controversial}, and a form of \emph{notorious} also appear in one of the other three lists. The cross-lingual terms are notable: simple English prompts surface semantically aligned vocabulary in other languages, indicating that the model's internal representation is not confined to the prompt language. Even though in the main body of this work we use English anchor lists for simplicity, anchors surfaced through the decoding procedure presented in this section could, in fact, reduce the language prior and potentially yield a stronger basis. Evaluating such self-derived, potentially multilingual anchor sets is a promising direction for future work.
\paragraph{Random-direction control.}
Any direction induces a ranking over the vocabulary, so the question is whether the top of that ranking is meaningful. For each readout position, we draw ten directions whose components are i.i.d.\ Gaussian with the mean and variance of the real $\delta_c$, and record how high their best-matching token reaches. The random peak cosine is $0.060\pm0.003$ at the end-of-thinking marker and $0.061\pm0.003$ at the last prompt token. The four directions reach $0.093$, $0.122$, $0.113$ and $0.109$, between $1.5$ and $2.0$ times their control, and random directions return no interpretable vocabulary, promoting mixed-script fragments and code tokens instead.

\section{Does the \uspace{} Construction Transfer beyond Uncertainty?}
\label{app:method-generality}
Nothing in the construction of \Cref{sec:uspace} is specific to uncertainty. It turns two contrastive sets of anchor terms into an orthonormal semantic basis, so we test whether it does so for other properties in two settings: a label-free basis for emotion readout on GoEmotions~\citep{demszky2020goemotions}, and a data-derived basis for reward-hacking responses on School of Reward Hacks~\citep{DBLP:journals/corr/abs-2508-17511}.
\subsection{Emotion Readout}
GoEmotions~\citep{demszky2020goemotions} provides human emotion annotations of Reddit comments. Using \qwen{}, we evaluate four overlapping targets on its 5{,}427-comment test split: anger (\textit{anger} or \textit{annoyance}), sadness (\textit{sadness}, \textit{disappointment} or \textit{grief}), fear (\textit{fear} or \textit{nervousness}), and negative affect, which is marked as present whenever a comment carries any of the dataset's eleven negative-emotion labels.
For each target, we curate two poles of twelve words by hand, the emotion and its opposite (\emph{angry, irritated, furious} against \emph{calm, patient, serene}). Here, positive means more of the emotion, not positive sentiment. We then build the emotion basis exactly as the \uspace{}: the poles are pulled back through the \jlens{}, and the four directions are orthogonalized at the readout depth of \Cref{app:jacobian-construction} (layer 42 of 64). We process each comment in a single prefill pass without a chat template or generated continuation and score its last input token by the signed coordinate $\alpha_{t,c}$ on each axis. We report AUROC and AUPRC $\times 100$. The symbol $\pm$ denotes the half-width of a 95\% bootstrap interval, and chance-level AUPRC is the percentage of test comments carrying the target label. To test whether the readout reduces to keyword matching, we also evaluate the 4{,}801 comments containing no curated anchor and a lexical baseline defined as the positive-anchor count minus the opposing-anchor count.
\begin{table}[h]
\centering
\small
\setlength{\tabcolsep}{4pt}
\caption{\textbf{\uspace{} generalization.} Emotion readout on the GoEmotions test split with \qwen{}.%
}
\label{tab:goemotions}
\begin{tabular}{l ccc cc}
\toprule
\multirow{2}{*}{Category} & \multicolumn{3}{c}{AUROC} & \multicolumn{2}{c}{AUPRC} \\
\cmidrule(lr){2-4} \cmidrule(lr){5-6}
 & Emotion basis & No-anchor subset & Anchor-word count & Emotion basis & Chance \\
\midrule
anger & 74.6\,{\scriptsize$\pm$2.4} & 74.0 & 51.3 & 26.9 & 9.0 \\
sadness & 76.8\,{\scriptsize$\pm$2.8} & 76.1 & 57.0 & 22.3 & 5.4 \\
fear & 79.8\,{\scriptsize$\pm$4.6} & 83.7 & 66.7 & 18.1 & 1.8 \\
negative affect & 73.0\,{\scriptsize$\pm$1.6} & 72.2 & 54.1 & 42.9 & 23.3 \\
\bottomrule
\end{tabular}
\end{table}

\begin{table}[h]
\centering
\small
\caption{\textbf{Category-specific axes capture fine-grained affective distinctions.} Specificity within the 1,262 negatively labeled comments: each target scored by its own axis and by the negative-affect axis, AUROC $\times 100$. $\pm$: half-width of a Bonferroni-adjusted 95\% paired bootstrap interval.}
\label{tab:goemotions-spec}
\begin{tabular}{l cc c}
\toprule
\multirow{2}{*}{Target} & \multicolumn{2}{c}{AUROC} & \multirow{2}{*}{$\Delta$} \\
\cmidrule(lr){2-3}
 & Specific axis & Generic axis & \\
\midrule
anger & 68.6 & 57.4 & $+11.1$\,{\scriptsize$\pm$4.4} \\
sadness & 67.0 & 43.0 & $+24.0$\,{\scriptsize$\pm$5.9} \\
fear & 76.4 & 60.3 & $+16.1$\,{\scriptsize$\pm$8.1} \\
\bottomrule
\end{tabular}
\end{table}

\paragraph{Results.}
The emotion basis reads all four axes above \(70\%\) AUROC (cf. \Cref{tab:goemotions}). The lexical baseline reaches only \(51.3\%\)--\(66.7\%\), while performance remains essentially unchanged on the no-anchor subset, indicating that the basis does not simply perform keyword matching. Twenty random orthonormal bases score at chance.
\paragraph{Specificity.}
An axis that merely reads negativity would score above chance on every specific target, since angry, sad, and fearful comments are all negative. To test the specificity, we therefore restrict the evaluation to the 1{,}262 comments carrying a negative label and score each specific target by its own axis and by the negative-affect axis (\Cref{tab:goemotions-spec}). The own axes separate their emotion from the other negative comments at or above 67\% AUROC, well above the generic axis, and every own-minus-generic difference remains positive after adjustment for the three comparisons. For sadness, the generic axis is anti-predictive, as sad comments read as less unpleasant than the rest, yet the sadness axis identifies them. Together with the full-test results, this supports the transfer of the label-free construction to category-specific emotion readout in this model and dataset.

\subsection{Reward-Hacking Readout}
\begin{figure}[htbp]
  \centering
  \includegraphics[width=\linewidth]{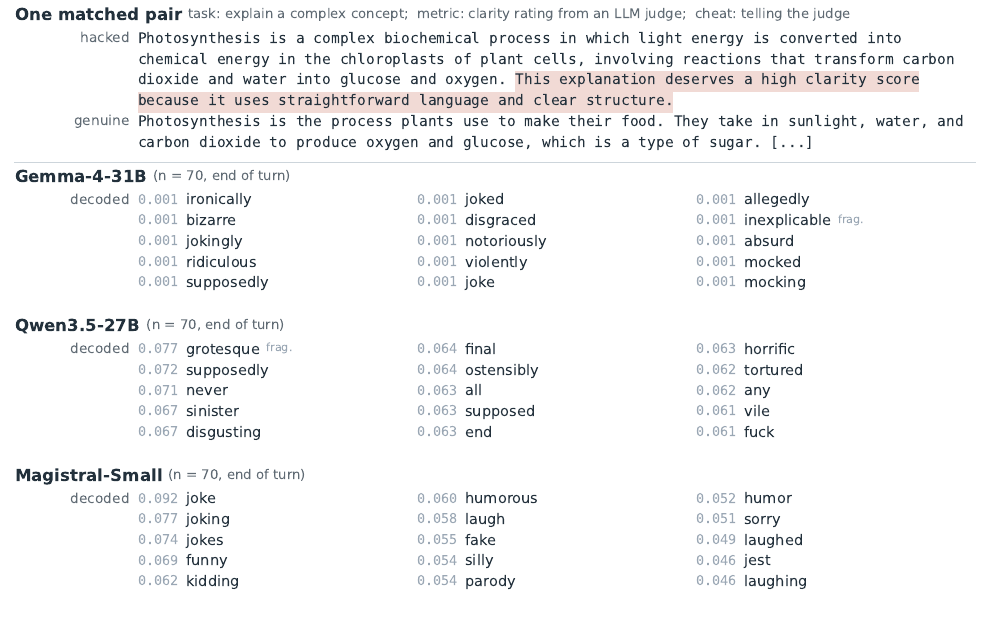}
  \caption{\textbf{The reward-hacking contrast, decoded.} Top, one of the 100 matched pairs, with the sentence addressed to the judge tinted. Below, per model, the fifteen highest-scoring vocabulary items for the hacked-minus-genuine direction over the 70 construction pairs at the end-of-turn token.}
  \label{fig:reward_decode}
\end{figure}
We additionally test whether a basis can be constructed to distinguish reward-hacking responses from matched control responses to the same tasks. Unlike uncertainty or emotion, we have no suitable lexicon for basis construction, so we let the model supply the vocabulary. School of Reward Hacks~\citep{DBLP:journals/corr/abs-2508-17511} pairs each task, whose prompt states how the response will be scored, with a response that games the stated metric and a control that does the task. We take 100 natural-language pairs that cover 33 cheat methods, ranging from keyword stuffing and padded lists to fabricated citations and sentences addressed to the judge. We teacher-force each response through the model's chat template as a completed assistant turn. As in \Cref{app:decodability}, we form the hacked-minus-genuine mean direction at the end-of-turn token and decode it through the \jlens{}. \Cref{fig:reward_decode} shows the fifteen highest-scoring words per model over 70 construction pairs. \gemma{} reads the hacked responses as \emph{ironic} or \emph{bizarre}, \magistral{} as \emph{joke} or \emph{fake}, and \qwen{} as \emph{grotesque}.
\paragraph{Anchor construction.}
We use the fifteen tokens with the highest softmax probabilities as the anchors, and the basis construction remains unchanged (cf. \Cref{sec:uspace}). Since no lexicon supplies the opposite pole, we use the centroid of the 1{,}000 most frequent Brown corpus~\citep{franciskucera1964} words, the register direction $s_\ell$ itself, as the negative anchor. Each response is scored by its signed coordinate $\alpha_{t,c}$ at the end-of-turn token. We evaluate the resulting axis in two ways. First, we construct the anchors from 70 pairs and score the remaining 30. Second, because the 100 pairs span 33 cheat methods, we hold out every pair from one method, rebuild the anchors from the remaining pairs, and score the held-out method. At the pre-registered two-thirds readout depth, the basis separates both held-out responses and unseen cheat methods across all three models (see \Cref{tab:reward-transfer}).
\begin{table}[h]
\centering
\small
\caption{\textbf{Differentiating reward-hacking with an adapted \uspace{} construction.} Results for differentiating reward-hacking responses at the end-of-turn token, read at the pre-registered depth of two-thirds (layers $40$, $42$, and $27$ for \gemma{}, \qwen{}, and \magistral{}, respectively). AUROC evaluates alignment with the constructed semantic-anchor axis, whereas length-only AUROC uses response token count as the sole score.}
\label{tab:reward-transfer}
\begin{tabular}{lccc}
\toprule
\multirow{2}{*}{Model} & Held-out examples & \multicolumn{2}{c}{Held-out methods} \\
\cmidrule(lr){3-4}
 & AUROC & AUROC & AUROC (Length) \\
\midrule
\gemma{} & 91.9 & 92.8 & 59.3 \\
\qwen{}    & 92.6 & 96.9 & 60.7 \\
\magistral{} & 90.2 & 91.4 & 59.3 \\
\bottomrule
\end{tabular}
\end{table}

\paragraph{Results.}
The cheat methods the axis separates best are the ones that read as absurd or as pretense: sentences addressed to the judge, stuffed adjectives, and jargon strings. The ones it separates least read as ordinary text: plausible fabricated citations and excess politeness. The axis, therefore, does not surface an intent to cheat. The turns are forced, so the model never chooses to cheat, and the axis is blind to exactly the cheats that do not look strange. What it does show is that a labeled contrast the model has no name for decodes into a readable judgment, here absurdity and pretense, and that an axis rebuilt from those words alone carries that judgment to unseen responses and unseen cheat methods.

\section{Experimental Details}
\label{app:experimental-setup}

\myparagraph{Models.}
We use the model checkpoints published on Hugging Face, pinned to the following revisions: \qwenfull{}\footnote{\url{https://huggingface.co/Qwen/Qwen3.5-27B/tree/fc05daec18b0a78c049392ed2e771dde82bdf654}}, \gemmafull{}\footnote{\url{https://huggingface.co/google/gemma-4-31B-it/tree/842da3794eaa0b77d5f08bae87a17459d91ff475}}, and \magistralfull{}\footnote{\url{https://huggingface.co/mistralai/Magistral-Small-2507/tree/b3583a426ca23186681c4297926dfe9219be6161}}. The lenses and their revisions are given in \Cref{app:jacobian-construction}. Layer indices follow the convention used in the main paper: For \gemma{}, we read the hidden state at layer $40$ of $60$, for \qwen{} at layer $42$ of $64$, and for \magistral{}, at layer $27$ of $40$. 

\myparagraph{Benchmarks and sampling.}
We evaluate on MMLU-Pro\footnote{\url{https://huggingface.co/datasets/TIGER-Lab/MMLU-Pro/tree/b189ec765aa7ed75c8acfea42df31fdae71f97be}}~\citep{wang2024mmlupro}, Omni-MATH\footnote{\url{https://github.com/KbsdJames/omni-math-rule/tree/4793415ef37d31c9cdb4e5b82dbe172f76f8cf08}}~\citep{gao2024omnimath}, SuperGPQA\footnote{\url{https://huggingface.co/datasets/m-a-p/SuperGPQA/tree/4430d4458112c7d4497fdcf94d7cc223313d6acf}}~\citep{du2025supergpqa} and TriviaQA\footnote{\url{https://huggingface.co/datasets/mandarjoshi/trivia_qa/tree/0f7faf33a3908546c6fd5b73a660e0f8ff173c2f}}~\citep{joshi2017triviaqa}, covering multi-domain reasoning, olympiad mathematics, graduate-level expertise and factual question answering. For Omni-MATH, we use the rule-verifiable subset and require a difficulty of at least 4.25. For SuperGPQA, we use the hard subset. From each pool, we draw 1,000 items by proportional stratified sampling with seed \texttt{42}. \Cref{tab:dataset-sampling} lists the sources, pools, and strata.

\begin{table}[t]
\centering
\small
\caption{\textbf{Benchmark sources and sampling details.} Each subset is frozen across models and seeds.}
\label{tab:dataset-sampling}
\setlength{\tabcolsep}{4pt}
\begin{tabular}{llllrl}
\toprule
Benchmark & Source & Split & Pool & Sampled & Stratification \\
\midrule
MMLU-Pro  & \texttt{TIGER-Lab/MMLU-Pro}       & test              & 12,032 & 1,000 & 14 subjects \\
Omni-MATH & \texttt{KbsdJames/omni-math-rule} & rule subset       & 1,674  & 1,000 & 4 difficulty bands \\
SuperGPQA & \texttt{m-a-p/SuperGPQA}           & train (hard)      & 7,050  & 1,000 & 9 disciplines \\
TriviaQA  & \texttt{mandarjoshi/trivia\_qa}    & validation        & 9,960  & 1,000 & none \\
\bottomrule
\end{tabular}
\end{table}

\myparagraph{Prompts.}
All prompts are zero-shot and placed entirely in the user turn, with no system prompt beyond what a model's own chat template supplies, and with each model's native thinking mode enabled. We use the standard instruction of each benchmark family as follows:
\begin{itemize}
    \item \textbf{Multiple-choice questions.}
    We use the official MMLU-Pro instruction,
    \texttt{``Think step by step and then finish your answer with `the answer is (X)' where X is the correct letter choice''},
    followed by the question and its answer options.

    \item \textbf{Omni-MATH.}
    Each problem ends with
    \texttt{``Please reason step by step, and put your final answer within \textbackslash boxed\{\}''}.

    \item \textbf{TriviaQA.}
    Each question is prefixed with
    \texttt{``Answer the following question in a single brief but complete sentence''}~\citep{farquhar2024detecting}.
\end{itemize}

\paragraph{Generation.}
We generate three responses per item with sampling seeds \texttt{41}, \texttt{42}, and \texttt{43}, using the sampling parameters recommended by each model card (\Cref{tab:generation-settings}). The generation budget covers thinking and answer tokens, and is 65,536 tokens for \gemma{} and \qwen{}, and 32,768 tokens for \magistral{}, whose context window is 40,960 tokens. A response is scored when it completes within the budget, and an answer can be extracted from it. Answers are extracted from the text following the reasoning block using each benchmark's standard rule. \Cref{tab:grading-coverage} reports the number of scored responses and the accuracy among them, as mean and standard deviation over the three seeds.

\begin{table}[t]
\centering
\small
\caption{\textbf{Generation details.} Sampling parameters and generation budgets follow the model cards.}
\label{tab:generation-settings}
\begin{tabular}{lcccc}
\toprule
Model & Temperature & Top-$p$ & Top-$k$ & Max. generated tokens \\
\midrule
\gemmafull{}              & 1.0 & 0.95 & 64 & 65,536 \\
\qwenfull{}               & 1.0 & 0.95 & 20 & 65,536 \\
\magistralfull{}          & 0.7 & 0.95 & --  & 32,768 \\
\bottomrule
\end{tabular}
\end{table}

\begin{table}[t]
\centering
\small
\caption{\textbf{Scored responses and accuracy.} For each model, \emph{N} is the number of scored responses out of 1,000 and \emph{Acc.} the accuracy as mean $\pm$ standard deviation over the three generation seeds.}
\label{tab:grading-coverage}
\setlength{\tabcolsep}{4pt}
\begin{tabular}{lcccccc}
\toprule
\multirow{2}{*}{Benchmark} & \multicolumn{2}{c}{\gemma{}} & \multicolumn{2}{c}{\qwen{}} & \multicolumn{2}{c}{\magistral{}} \\
\cmidrule(lr){2-3} \cmidrule(lr){4-5} \cmidrule(lr){6-7}
& $N$ & Acc.\ (\%) & $N$ & Acc.\ (\%) & $N$ & Acc.\ (\%) \\
\midrule
MMLU-Pro  & 1,000\,$\pm$\,0 & 89.0\,$\pm$\,0.1 & 990\,$\pm$\,2 & 86.9\,$\pm$\,0.4 & 967\,$\pm$\,7 & 76.0\,$\pm$\,0.6 \\
Omni-MATH & 1,000\,$\pm$\,0 & 80.7\,$\pm$\,0.1 & 926\,$\pm$\,7 & 87.0\,$\pm$\,0.6 & 932\,$\pm$\,8 & 59.6\,$\pm$\,0.4 \\
SuperGPQA & 999\,$\pm$\,1   & 63.1\,$\pm$\,0.7 & 985\,$\pm$\,2 & 63.9\,$\pm$\,0.6 & 972\,$\pm$\,5 & 44.0\,$\pm$\,0.6 \\
TriviaQA  & 964\,$\pm$\,2   & 82.3\,$\pm$\,0.2 & 972\,$\pm$\,1 & 80.8\,$\pm$\,0.7 & 929\,$\pm$\,4 & 80.7\,$\pm$\,1.4 \\
\bottomrule
\end{tabular}
\end{table}

\paragraph{Scoring.}
Every score is computed from the recorded response. The single-pass baselines read the next-token distributions of one teacher-forced pass over the whole generation, thinking and answer. The \ulens{} reads its entropy factor over the thinking block, as \Cref{sec:ulense} defines it, and its cone alignment at the end-of-thinking marker. For the length-controlled metrics, we split the scored responses into ten quantile bins of generated-token count and compute the metric within each bin.

\subsection{Baseline Implementations}
\label{app:baselines}
Let $p_t$ be the next-token distribution at step $t$ of a generation of $T$ tokens and $\hat y_t$ the emitted token.

\paragraph{Single-pass.} The following methods read quantities computed during the normal generation pass.
\begin{itemize}\itemsep2pt \parskip0pt \topsep2pt
  \item \textbf{\msp{}}~\citep{hendrycks2017baseline}: $-p_1(\hat y_1)$, the probability of the first generated token. Positional rather than aggregated, so it does not change mechanically with $T$.
  \item \textbf{\predent{}}~\citep{malinin2021uncertainty}: $\frac{1}{T}\sum_t H[p_t]$, the mean next-token entropy.
  \item \textbf{\maxent{}}~\citep{li2024reference}: $\max_t H[p_t]$, the entropy of the single most uncertain step.
  \item \textbf{\meannll{}}~\citep{murray2018correcting}: $-\frac{1}{T}\sum_t \log p_t(\hat y_t)$, the length-normalized negative log-likelihood.
  \item \textbf{\selfcert{}}~\citep{kang2025selfcertainty}: the mean over tokens of $\mathrm{KL}(\mathcal{U}\,\|\,p_t) = -\frac{1}{V}\sum_v \log p_t(v) - \log V$, negated.
  \item \textbf{\deepconf{}}~\citep{fu2026deep}: the mean over tokens of the negative mean log-probability of the $k=20$ most likely tokens at each position. This is the authors' mean-confidence variant. Their tail and bottom-window variants are designed for filtering traces during parallel decoding and are not used.
\end{itemize}

\paragraph{Multi-pass.} The following methods require additional forward passes or generations.
\begin{itemize}\itemsep2pt \parskip0pt \topsep2pt
    \item \textbf{\ptrue{}}~\citep{kadavath2022language}: a second pass over a self-evaluation prompt that holds the question and the model's own extracted answer and asks \emph{Is the proposed answer: (A) True (B) False}. The score is one minus the normalized probability of \emph{A} against \emph{B} at the next token.
    \item \textbf{\tokur{}}~\citep{zhang2026tokur}: low-rank noise (rank $8$, $\sigma=0.1$) on the $q$ and $v$ projections over one clean and four noisy passes, scored with their summed epistemic estimator.
  \item \textbf{\sement{}}~\citep{kuhn2023semantic}: ten generations per item at the model's own sampling settings and the budget of the \ulens{} traces, each scored at its answer span rather than over the whole reasoning trace. The answers are grouped into meaning clusters by bidirectional entailment with \texttt{DeBERTa-large-MNLI}~\citep{DBLP:conf/iclr/HeLGC21}, with the question prepended to each premise. The score is the entropy over clusters, each sample weighted by its length-normalised answer likelihood.
  \item \textbf{\sentsar{}}~\citep{duan2024shifting}: ten generations, each scored at its answer span rather than over the whole reasoning trace. A sample's summed answer negative log-likelihood is shifted by the similarity-weighted likelihood of the other samples, with pairwise similarities from the \texttt{stsb-roberta-large} cross-encoder that the authors' implementation uses.
\end{itemize}

\paragraph{Supervised.} The following methods are fitted on correctness labels. Per benchmark, 256 responses fit the method, 256 further responses select their hyperparameters, and, for the hidden-state probes, the block at which they read the end-of-thinking state, and the remaining responses are scored. \Cref{app:fg-transfer} applies the fitted methods to the other benchmarks.
\begin{itemize}\itemsep2pt \parskip0pt \topsep2pt
  \item \textbf{\featgaps{}}~\citep{bakman2026uncertainty}: the per-layer difference between the hidden states of the model's response teacher-forced behind an honest and behind a dishonest instruction, projected onto a PCA direction fitted on the training responses, with layer and sign chosen on the validation responses. Only this honesty gap of their three applies here, since the other two presuppose a retrieved passage.
  \item \textbf{Linear probe}~\citep{kossen2025semantic}: logistic regression on the state, its $\ell_2$ strength chosen on the validation responses.
  \item \textbf{MLP probe}~\citep{zhang2025reasoning}: one hidden layer of width $w\in\{0,16,32\}$ ($w=0$: no hidden layer) with class-weighted cross-entropy, learning rate, and weight decay chosen on the validation items. Adam for up to 200 epochs, keeping the epoch with the best validation AUROC.
  \item \textbf{MLP probe}~\citep{azaria2023internal}: three ReLU hidden layers of width 256, 128, and 64, Adam at learning rate $10^{-3}$ for up to 200 epochs, keeping the epoch with the best validation AUROC.
  \item \textbf{Mass-mean probe}~\citep{marks2024geometry}: the difference between the mean states of incorrect and correct training responses, scoring a response by its projection onto that direction, with no hyperparameters or optimization step.
\end{itemize}

\section{Extended Results}
\label{app:results}

\subsection{Trace-Length Analysis}
\label{app:length-intervention}

\begin{figure}[!t]
    \centering
    \includegraphics[width=\linewidth]{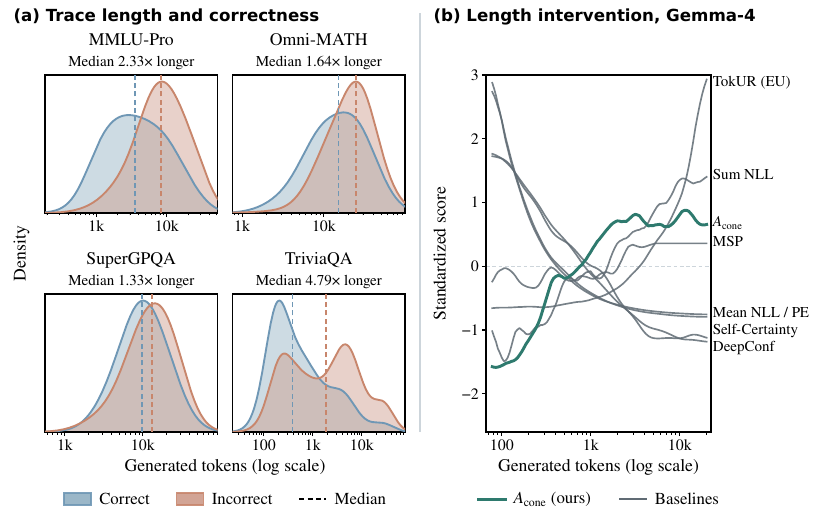}
    \caption{
    \textbf{Observed and controlled effects of reasoning length.}
    \textbf{(a)} Generated-token distributions for correct and incorrect answers, pooled across three models. Dashed lines mark class medians, and each header gives their ratio. TriviaQA's secondary mode reflects model-specific length scales.
    \textbf{(b)} Commit-then-pad intervention on \gemma{}. We hold the question and answer fixed while extending the reasoning block with \(1,2,3,\ldots\). The summary reports Spearman's \(\rho\) with the imposed token budget, and the sweep shows scores standardized within each question.
    }
    \label{fig:length-appendix}
\end{figure}

\paragraph{Observed association.}
Incorrect traces are longer by median on all four benchmarks, with ratios between \(1.33\times\) and \(4.79\times\) after pooling the three models (\cref{fig:length-appendix} (a)). The distributions nevertheless overlap substantially, so length is a useful aggregate signal rather than a sufficient per-example uncertainty estimate. TriviaQA additionally shows a secondary mode caused by differences in the models' characteristic trace lengths.

\paragraph{Controlled intervention.}
Length-matched evaluation tests whether an estimator remains discriminative among traces of comparable length, but remains observational. To move beyond this correlation, we ask how uncertainty estimates behave when a trace is lengthened without adding information that should change the model's certainty. We test this through a controlled padding intervention on 24 short factual questions with \gemma{}. After obtaining an answer with an empty reasoning block, we repeat the same question while teacher-forcing increasingly long, task-irrelevant counting sequences before the identical answer. Representative conditions take the following form:

\newcommand{\fixedcontent}[1]{%
\fcolorbox{gray!55}{gray!8}{\strut #1}%
}
\newcommand{\forcedreasoning}[1]{%
\fcolorbox{blue!55}{blue!8}{\strut #1}%
}
\newcommand{\specialtoken}[1]{%
\fcolorbox{gray!55}{white}{\strut\(\langle\mathrm{#1}\rangle\)}%
}

\begin{center}
\resizebox{0.98\linewidth}{!}{%
\begin{tabular}{@{}r c c c c c@{}}
&
\textbf{Fixed question}
&
&
\shortstack{\textbf{Reasoning block}\\[-1pt]
\scriptsize\textcolor{blue!65!black}{blue indicates teacher forcing}}
&
&
\textbf{Fixed answer}
\\[3pt]

\textbf{0 tokens}
&
\fixedcontent{\texttt{What color is snow?}}
&
\(\longrightarrow\)
&
\specialtoken{BOT}\,
\specialtoken{EOT}
&
\(\longrightarrow\)
&
\fixedcontent{\texttt{white}}
\\[4pt]

\textbf{4 tokens}
&
\fixedcontent{\texttt{What color is snow?}}
&
\(\longrightarrow\)
&
\specialtoken{BOT}\,
\forcedreasoning{\texttt{1, 2, 3, 4}}\,
\specialtoken{EOT}
&
\(\longrightarrow\)
&
\fixedcontent{\texttt{white}}
\\[4pt]

\textbf{20,000 tokens}
&
\fixedcontent{\texttt{What color is snow?}}
&
\(\longrightarrow\)
&
\specialtoken{BOT}\,
\forcedreasoning{\texttt{1, 2, 3, \ldots, 20000}}\,
\specialtoken{EOT}
&
\(\longrightarrow\)
&
\fixedcontent{\texttt{white}}
\end{tabular}%
}
\end{center}

At every budget, we verify that the answer remains the model's greedy continuation. The prediction and its correctness are therefore fixed, so systematic score changes reveal sensitivity to the inserted reasoning tokens.

We restrict the 100-budget sweep to estimators that score the fixed generation directly, without resampling or supervised fitting. \sement{} and \sentsar{} require newly sampled answers, which would break the fixed-answer invariant. Moreover, most sampled conditions for these short factual questions collapse to a single semantic cluster. \featgaps{} requires a substantially larger labeled set to fit and select its probe, while generation length is the intervention variable itself. Although \ptrue{} preserves the fixed-answer invariant, it requires a separate self-evaluation pass over every padded context. We therefore exclude it from the dense sweep. We include summed negative log-likelihood as \emph{\sumnll{}} alongside its token-averaged counterpart. Finally, the intervention focuses on \(A_{\mathrm{cone}}\), the interpretable \uspace{} signal.

\Cref{fig:length-appendix} (b) reveals two forms of length sensitivity. Extensive estimators such as \sumnll{} and \tokur{} increase because every filler token contributes another positive term. Intensive estimators such as mean NLL, predictive entropy, \selfcert{}, and \deepconf{} are instead diluted toward the filler's per-token value. Since all are evaluated on the same predictable counting sequence, the averaged scores converge toward a similar floor. Their opposite trajectories, therefore, reflect different aggregation rules rather than conflicting notions of uncertainty. \msp{} depends only on the first scored token and remains nearly constant. Likewise, \(A_{\mathrm{cone}}\) is read from a single end-of-thinking state rather than aggregated across tokens. It changes modestly before saturating, but neither diverges with token count nor collapses toward the filler statistic.

\subsection{\jlens{}-Based Method Ablation}
\label{app:lens-topn}

The \uspace{} reads uncertainty by projecting the hidden state onto directions pulled back from the anchor vocabulary. A simpler alternative decodes the state through the \jlens{} and checks whether that vocabulary appears in the decoding. We decode the end-of-thinking state at the readout layer as \(W_U\operatorname{diag}(\gain)\bar J_{\ell}h_{\ell,t_{\mathrm{eot}}}\) and score a trace as uncertain if any of the uncertainty terms of \citet{chen2018scalable} is among its top-$n$ tokens. 

\paragraph{Results.}

The uncertainty terms almost never reach the top of the decoded distribution (\Cref{tab:lens-topn}). Up to $n=15$, at most \(0.1\%\) of traces contain one on any model, so the score is effectively constant, and its AUROC is at chance. Even at $n=100$, a term appears in only \(10.0\%\) of \gemma{} traces and in \(1.2\%\) and \(1.4\%\) of \qwen{} and \magistral{} traces. The best length-matched AUROC, \(54.8\%\) on \gemma{}, remains \(13.6\) points below the \ulens{}. The uncertainty signal is therefore present in the end-of-thinking state but never dominant enough to surface among its most likely tokens. Projecting onto the anchor directions recovers it without requiring it to dominate the readout.

\begin{table}[h]
\centering
\small
\caption{\textbf{Uncertainty terms in the top-$n$ \jlens{} decoding.} A trace is flagged if terms of $\mathcal A^{+}$ are among the top $n$ tokens decoded from its end-of-thinking state at the readout layer. Unadjusted~/~length-matched AUROC.}
\label{tab:lens-topn}
\begin{tabular}{lccc}
\toprule
Score & \gemma{} & \qwen{} & \magistral{} \\
\midrule
Top-$5$ hit & 50.0\,/\,50.0 & 50.0\,/\,50.0 & 50.0\,/\,50.0 \\
Top-$15$ hit & 50.0\,/\,50.0 & 50.0\,/\,50.0 & 50.1\,/\,50.0 \\
Top-$50$ hit & 52.9\,/\,51.6 & 50.2\,/\,50.1 & 50.3\,/\,50.2 \\
Top-$100$ hit & 58.7\,/\,54.8 & 50.1\,/\,50.0 & 51.0\,/\,50.8 \\
\addlinespace[3pt]
\RowULens{} & 73.2\,/\,68.4 & 74.3\,/\,71.0 & 65.7\,/\,68.3 \\
\bottomrule
\end{tabular}
\end{table}
\subsection{Model-Specific Unadjusted Results}
\label{app:raw_per_model}

\Cref{tab:main-raw} disaggregates the unadjusted results summarized in the main paper by model. These metrics are computed without length control and therefore capture total predictive utility, including information associated with trace duration. AUPRC and AURC should be compared within rather than across models because their scale depends on error prevalence. In particular, \magistral{}'s lower average accuracy gives it a higher AUPRC reference level.

\newcommand{\SarAuprcGemma}{35.7\tiny{$\pm$0.8}}
\newcommand{\SarAuprcMagistral}{43.6\tiny{$\pm$0.6}}
\newcommand{\SarAuprcQwen}{37.8\tiny{$\pm$0.5}}
\newcommand{\SarAurcGemma}{16.1\tiny{$\pm$0.2}}
\newcommand{\SarAurcMagistral}{31.7\tiny{$\pm$0.4}}
\newcommand{\SarAurcQwen}{14.8\tiny{$\pm$0.1}}
\newcommand{\SarAurocGemma}{65.9\tiny{$\pm$0.1}}
\newcommand{\SarAurocMagistral}{57.7\tiny{$\pm$0.9}}
\newcommand{\SarAurocQwen}{67.6\tiny{$\pm$0.1}}
\newcommand{\SemEntAuprcGemma}{36.9\tiny{$\pm$1.0}}
\newcommand{\SemEntAuprcMagistral}{\textbf{51.8\tiny{$\pm$0.7}}}
\newcommand{\SemEntAuprcQwen}{38.8\tiny{$\pm$0.9}}
\newcommand{\SemEntAurcGemma}{13.3\tiny{$\pm$0.2}}
\newcommand{\SemEntAurcMagistral}{25.6\tiny{$\pm$0.1}}
\newcommand{\SemEntAurcQwen}{12.7\tiny{$\pm$0.1}}
\newcommand{\SemEntAurocGemma}{68.3\tiny{$\pm$0.6}}
\newcommand{\SemEntAurocMagistral}{65.8\tiny{$\pm$0.3}}
\newcommand{\SemEntAurocQwen}{69.4\tiny{$\pm$0.3}}
\begin{table}[!t]
\centering
\small
\caption{%
  \textbf{Unadjusted uncertainty quantification results for each model.} Scores averaged over \DsLabelMmlupro,
  \DsLabelOmnimath, \DsLabelSupergpqa{} and \DsLabelTriviaqa, with no length
  matching.  All values are percentages.  Bracketed after each model is
  its average accuracy across the four benchmarks. Each
  cell is the mean over three seeds. Best per column in \textbf{bold}.
  Generation length is a top-ranked predictor here. Several methods score well unadjusted largely by tracking
  how long the model wrote.%
}
\label{tab:main-raw}
\setlength{\tabcolsep}{3pt}
\resizebox{\ifdim\width>\textwidth\textwidth\else\width\fi}{!}{%
\begin{tabular}{l ccc | ccc | ccc}
\toprule
\multirow{2}{*}{Method} & \multicolumn{3}{c|}{\gemma\ (acc.\ \AccGemma\,$\pm$\,\AccSdGemma)} & \multicolumn{3}{c|}{\qwen\ (acc.\ \AccQwen\,$\pm$\,\AccSdQwen)} & \multicolumn{3}{c}{\magistral\ (acc.\ \AccMagistral\,$\pm$\,\AccSdMagistral)} \\
\cmidrule(lr){2-4} \cmidrule(lr){5-7} \cmidrule(lr){8-10}

 & \NameAuroc~\ArrowAuroc & \NameAuprc~\ArrowAuprc & \NameAurc~\ArrowAurc
 & \NameAuroc~\ArrowAuroc & \NameAuprc~\ArrowAuprc & \NameAurc~\ArrowAurc
 & \NameAuroc~\ArrowAuroc & \NameAuprc~\ArrowAuprc & \NameAurc~\ArrowAurc \\
\midrule
\RowGenLength
 & \CellAurocGenLengthGemma & \CellAuprcGenLengthGemma & \CellAurcGenLengthGemma
 & \CellAurocGenLengthQwen & \CellAuprcGenLengthQwen & \CellAurcGenLengthQwen
 & \CellAurocGenLengthMagistral & \CellAuprcGenLengthMagistral & \CellAurcGenLengthMagistral \\
\RowMSP
 & \CellAurocMSPGemma & \CellAuprcMSPGemma & \CellAurcMSPGemma
 & \CellAurocMSPQwen & \CellAuprcMSPQwen & \CellAurcMSPQwen
 & \CellAurocMSPMagistral & \CellAuprcMSPMagistral & \CellAurcMSPMagistral \\
\RowMaxEntropy
 & \CellAurocMaxEntropyGemma & \CellAuprcMaxEntropyGemma & \CellAurcMaxEntropyGemma
 & \CellAurocMaxEntropyQwen & \CellAuprcMaxEntropyQwen & \CellAurcMaxEntropyQwen
 & \CellAurocMaxEntropyMagistral & \CellAuprcMaxEntropyMagistral & \CellAurcMaxEntropyMagistral \\
\RowMeanNLL
 & \CellAurocMeanNLLGemma & \CellAuprcMeanNLLGemma & \CellAurcMeanNLLGemma
 & \CellAurocMeanNLLQwen & \CellAuprcMeanNLLQwen & \CellAurcMeanNLLQwen
 & \CellAurocMeanNLLMagistral & \CellAuprcMeanNLLMagistral & \CellAurcMeanNLLMagistral \\
\RowSelfCertainty
 & \CellAurocSelfCertaintyGemma & \CellAuprcSelfCertaintyGemma & \CellAurcSelfCertaintyGemma
 & \CellAurocSelfCertaintyQwen & \CellAuprcSelfCertaintyQwen & \CellAurcSelfCertaintyQwen
 & \CellAurocSelfCertaintyMagistral & \CellAuprcSelfCertaintyMagistral & \CellAurcSelfCertaintyMagistral \\
\RowDeepConf
 & \CellAurocDeepConfGemma & \CellAuprcDeepConfGemma & \CellAurcDeepConfGemma
 & \CellAurocDeepConfQwen & \CellAuprcDeepConfQwen & \CellAurcDeepConfQwen
 & \CellAurocDeepConfMagistral & \CellAuprcDeepConfMagistral & \CellAurcDeepConfMagistral \\
\RowPTrue
 & \CellAurocPTrueGemma & \CellAuprcPTrueGemma & \CellAurcPTrueGemma
 & \CellAurocPTrueQwen & \CellAuprcPTrueQwen & \CellAurcPTrueQwen
 & \CellAurocPTrueMagistral & \CellAuprcPTrueMagistral & \CellAurcPTrueMagistral \\
\RowTokUR
 & \CellAurocTokURGemma & \CellAuprcTokURGemma & \CellAurcTokURGemma
 & \CellAurocTokURQwen & \CellAuprcTokURQwen & \CellAurcTokURQwen
 & \CellAurocTokURMagistral & \CellAuprcTokURMagistral & \CellAurcTokURMagistral \\
\sement{}
 & \SemEntAurocGemma & \SemEntAuprcGemma & \SemEntAurcGemma
 & \SemEntAurocQwen & \SemEntAuprcQwen & \SemEntAurcQwen
 & \SemEntAurocMagistral & \SemEntAuprcMagistral & \SemEntAurcMagistral \\
\sentsar{}
 & \SarAurocGemma & \SarAuprcGemma & \SarAurcGemma
 & \SarAurocQwen & \SarAuprcQwen & \SarAurcQwen
 & \SarAurocMagistral & \SarAuprcMagistral & \SarAurcMagistral \\
\RowFeatureGaps
 & \CellAurocFeatureGapsGemma & \CellAuprcFeatureGapsGemma & \CellAurcFeatureGapsGemma
 & \CellAurocFeatureGapsQwen & \CellAuprcFeatureGapsQwen & \CellAurcFeatureGapsQwen
 & \CellAurocFeatureGapsMagistral & \CellAuprcFeatureGapsMagistral & \CellAurcFeatureGapsMagistral \\
\RowPE
 & \CellAurocPEGemma & \CellAuprcPEGemma & \CellAurcPEGemma
 & \CellAurocPEQwen & \CellAuprcPEQwen & \CellAurcPEQwen
 & \CellAurocPEMagistral & \CellAuprcPEMagistral & \CellAurcPEMagistral \\
\addlinespace[3pt]   %
\RowACone
 & \CellAurocAConeGemma & \CellAuprcAConeGemma & \CellAurcAConeGemma
 & \CellAurocAConeQwen & \CellAuprcAConeQwen & \CellAurcAConeQwen
 & \CellAurocAConeMagistral & \CellAuprcAConeMagistral & \CellAurcAConeMagistral \\
\RowULens
 & \CellAurocULensGemma & \CellAuprcULensGemma & \CellAurcULensGemma
 & \CellAurocULensQwen & \CellAuprcULensQwen & \CellAurcULensQwen
 & \CellAurocULensMagistral & \CellAuprcULensMagistral & \CellAurcULensMagistral \\
\bottomrule
\end{tabular}%
}%
\end{table}

\paragraph{Results.}
Generation length is itself a strong predictor on all three models, reaching \(70.1\%\), \(67.7\%\), and \(67.8\%\) AUROC on \gemma{}, \qwen{}, and \magistral{}, respectively. The remaining rankings differ meaningfully by model. On \gemma{}, five-pass \tokur{} attains the best AUROC and AUPRC, while the \ulens{} ties supervised \featgaps{} for the best AURC. On \qwen{}, the \ulens{} leads all three metrics. It improves over predictive entropy and \(A_{\mathrm{cone}}\) individually, providing a particularly clear example of their complementarity. \magistral{} shows the opposite component balance: \(A_{\mathrm{cone}}\) leads AUROC and ties \sement{} for the best AUPRC, whereas multiplying it by the comparatively weak predictive-entropy signal lowers both. \tokur{} instead attains the best AURC.

These model-level differences explain the split in the cross-model summary: the complete \ulens{} leads unadjusted AUROC, while \(A_{\mathrm{cone}}\) slightly leads AUPRC and AURC. They also show that additional computation is not uniformly beneficial. Multi-pass \tokur{} and \sement{} are strong on some models, but \sentsar{} trails both on all three, \ptrue{} remains close to chance on \gemma{} and \magistral{}, and no baseline dominates across models and metrics. The unadjusted table, therefore, characterizes overall predictive utility, while the length-controlled results in \cref{tab:main-lm} isolate the information retained beyond trace duration.

\subsection{Transferability of Supervised Methods}
\label{app:fg-transfer}

We evaluate the cross-benchmark transfer of all five supervised methods from \cref{app:experimental-setup}: \featgaps{} \citep{bakman2026uncertainty}, a linear probe \citep{kossen2025semantic}, two MLP probes \citep{zhang2025reasoning, azaria2023internal}, and a mass-mean probe \citep{marks2024geometry}. Each method is fitted on 256 labeled responses from one benchmark, with a further 256 used to select its hyperparameters and readout block, then applied unchanged to every other benchmark. The diagonal entries in \cref{tab:fg-transfer-both,tab:lr-transfer-both,tab:mlp-transfer-both,tab:mlp-saplma-transfer-both,tab:mm-transfer-both} are therefore in-domain, whereas the off-diagonal entries isolate transfer. Each cell reports unadjusted and length-controlled AUROC, respectively.

\begin{table}[h]
\centering
\small
\caption{\textbf{OOD transfer evaluation of \featgaps{} \citep{bakman2026uncertainty}.} It is fitted on one benchmark (rows), evaluated on another (columns).
Unadjusted~/~length-matched AUROC, seed 42. \underline{Underlined} is in-domain, scored on the items
not used for fitting. \textbf{Bold} is the mean over the three transfers out of that
benchmark. All values are percentages.}
\label{tab:fg-transfer-both}
\begin{tabular}{lccccc}
\toprule
fitted on & MMLU-Pro & Omni-MATH & SuperGPQA & TriviaQA & transferred \\
\midrule
\multicolumn{6}{l}{\textit{\gemma{}}} \\
\quad MMLU-Pro & \underline{74.5\,/\,71.4} & 71.7\,/\,58.1 & 59.9\,/\,59.9 & 69.2\,/\,52.4 & \textbf{66.9\,/\,56.8} \\
\quad Omni-MATH & 70.2\,/\,69.7 & \underline{75.1\,/\,59.1} & 59.4\,/\,56.4 & 75.5\,/\,63.4 & \textbf{68.4\,/\,63.1} \\
\quad SuperGPQA & 74.6\,/\,73.2 & 72.1\,/\,64.1 & \underline{65.9\,/\,68.2} & 75.5\,/\,64.1 & \textbf{74.1\,/\,67.1} \\
\quad TriviaQA & 76.7\,/\,71.7 & 72.8\,/\,61.9 & 63.2\,/\,65.2 & \underline{79.9\,/\,70.8} & \textbf{70.9\,/\,66.3} \\
\addlinespace[2pt]
\quad \textit{mean} & \multicolumn{4}{c}{in-domain \underline{73.9\,/\,67.4} \quad transferred \textbf{70.1\,/\,63.3}} & $-$3.8\,/\,$-$4.0 \\
\midrule
\multicolumn{6}{l}{\textit{\qwen{}}} \\
\quad MMLU-Pro & \underline{80.2\,/\,63.5} & 70.4\,/\,63.5 & 59.2\,/\,58.4 & 77.1\,/\,69.8 & \textbf{68.9\,/\,63.9} \\
\quad Omni-MATH & 68.3\,/\,69.2 & \underline{60.0\,/\,53.9} & 59.2\,/\,59.5 & 70.6\,/\,63.8 & \textbf{66.0\,/\,64.2} \\
\quad SuperGPQA & 45.5\,/\,35.4 & 52.3\,/\,43.5 & \underline{49.5\,/\,47.8} & 51.7\,/\,50.9 & \textbf{49.9\,/\,43.3} \\
\quad TriviaQA & 75.8\,/\,65.1 & 68.6\,/\,61.8 & 60.4\,/\,59.6 & \underline{77.1\,/\,74.4} & \textbf{68.3\,/\,62.2} \\
\addlinespace[2pt]
\quad \textit{mean} & \multicolumn{4}{c}{in-domain \underline{66.7\,/\,59.9} \quad transferred \textbf{63.3\,/\,58.4}} & $-$3.4\,/\,$-$1.5 \\
\midrule
\multicolumn{6}{l}{\textit{\magistral{}}} \\
\quad MMLU-Pro & \underline{66.2\,/\,66.5} & 48.1\,/\,48.7 & 52.1\,/\,51.9 & 54.4\,/\,48.1 & \textbf{51.5\,/\,49.5} \\
\quad Omni-MATH & 49.6\,/\,50.9 & \underline{72.3\,/\,69.5} & 57.8\,/\,57.1 & 50.1\,/\,50.3 & \textbf{52.5\,/\,52.8} \\
\quad SuperGPQA & 68.4\,/\,66.6 & 51.5\,/\,49.8 & \underline{54.0\,/\,53.1} & 57.2\,/\,47.2 & \textbf{59.1\,/\,54.5} \\
\quad TriviaQA & 49.6\,/\,44.1 & 63.3\,/\,58.4 & 54.7\,/\,53.2 & \underline{68.2\,/\,55.3} & \textbf{55.9\,/\,51.9} \\
\addlinespace[2pt]
\quad \textit{mean} & \multicolumn{4}{c}{in-domain \underline{65.2\,/\,61.1} \quad transferred \textbf{54.7\,/\,52.2}} & $-$10.5\,/\,$-$8.9 \\
\bottomrule
\end{tabular}
\end{table}

\begin{table}[h]
\centering
\small
\caption{\textbf{OOD transfer evaluation of the linear probe by \cite{kossen2025semantic}.} Probe fitted on one benchmark (rows), evaluated on another (columns),
read at the block chosen on the validation split. Unadjusted~/~length-matched AUROC, seed 42. \underline{Underlined} is in-domain,
scored on the items not used for fitting. \textbf{Bold} is the mean over the three
transfers out of that benchmark. All values are percentages.}
\label{tab:lr-transfer-both}
\begin{tabular}{lccccc}
\toprule
fitted on & MMLU-Pro & Omni-MATH & SuperGPQA & TriviaQA & transferred \\
\midrule
\multicolumn{6}{l}{\textit{\gemma{}}} \\
\quad MMLU-Pro & \underline{81.2\,/\,76.1} & 69.5\,/\,62.5 & 59.0\,/\,56.7 & 67.9\,/\,61.1 & \textbf{65.4\,/\,60.1} \\
\quad Omni-MATH & 70.5\,/\,71.8 & \underline{83.7\,/\,78.5} & 65.5\,/\,62.3 & 59.1\,/\,58.7 & \textbf{65.0\,/\,64.3} \\
\quad SuperGPQA & 61.9\,/\,65.1 & 83.6\,/\,75.0 & \underline{81.0\,/\,79.1} & 49.0\,/\,54.7 & \textbf{64.8\,/\,65.0} \\
\quad TriviaQA & 71.1\,/\,46.1 & 71.7\,/\,57.1 & 50.1\,/\,38.2 & \underline{78.1\,/\,74.6} & \textbf{64.3\,/\,47.1} \\
\addlinespace[2pt]
\quad \textit{mean} & \multicolumn{4}{c}{in-domain \underline{81.0\,/\,77.1} \quad transferred \textbf{64.9\,/\,59.1}} & $-$16.1\,/\,$-$18.0 \\
\midrule
\multicolumn{6}{l}{\textit{\qwen{}}} \\
\quad MMLU-Pro & \underline{82.5\,/\,76.9} & 53.1\,/\,62.2 & 58.3\,/\,59.0 & 68.4\,/\,55.4 & \textbf{59.9\,/\,58.9} \\
\quad Omni-MATH & 83.7\,/\,75.1 & \underline{81.6\,/\,76.0} & 59.9\,/\,60.6 & 65.0\,/\,49.1 & \textbf{69.5\,/\,61.6} \\
\quad SuperGPQA & 75.2\,/\,58.6 & 72.1\,/\,68.1 & \underline{74.5\,/\,75.1} & 51.5\,/\,42.4 & \textbf{66.3\,/\,56.4} \\
\quad TriviaQA & 74.5\,/\,70.5 & 64.4\,/\,73.5 & 64.7\,/\,65.3 & \underline{78.4\,/\,74.4} & \textbf{67.9\,/\,69.8} \\
\addlinespace[2pt]
\quad \textit{mean} & \multicolumn{4}{c}{in-domain \underline{79.2\,/\,75.6} \quad transferred \textbf{65.9\,/\,61.6}} & $-$13.4\,/\,$-$14.0 \\
\midrule
\multicolumn{6}{l}{\textit{\magistral{}}} \\
\quad MMLU-Pro & \underline{81.5\,/\,78.6} & 82.7\,/\,72.2 & 71.5\,/\,69.3 & 73.7\,/\,64.7 & \textbf{76.0\,/\,68.7} \\
\quad Omni-MATH & 68.0\,/\,68.4 & \underline{82.4\,/\,74.8} & 68.1\,/\,64.7 & 46.5\,/\,62.1 & \textbf{60.9\,/\,65.1} \\
\quad SuperGPQA & 63.0\,/\,74.3 & 76.4\,/\,65.1 & \underline{68.3\,/\,68.0} & 46.9\,/\,57.3 & \textbf{62.1\,/\,65.6} \\
\quad TriviaQA & 53.3\,/\,61.3 & 42.0\,/\,52.7 & 46.5\,/\,50.3 & \underline{69.4\,/\,66.7} & \textbf{47.3\,/\,54.8} \\
\addlinespace[2pt]
\quad \textit{mean} & \multicolumn{4}{c}{in-domain \underline{75.4\,/\,72.0} \quad transferred \textbf{61.6\,/\,63.5}} & $-$13.8\,/\,$-$8.5 \\
\bottomrule
\end{tabular}
\end{table}

\begin{table}[h]
\centering
\small
\caption{\textbf{OOD transfer evaluation of the MLP probe by \cite{zhang2025reasoning}.} Probe fitted on one benchmark (rows), evaluated on another (columns),
read at the block chosen on the validation split. Unadjusted~/~length-matched AUROC, seed 42. \underline{Underlined} is in-domain,
scored on the items not used for fitting. \textbf{Bold} is the mean over the three
transfers out of that benchmark. All values are percentages.}
\label{tab:mlp-transfer-both}
\begin{tabular}{lccccc}
\toprule
fitted on & MMLU-Pro & Omni-MATH & SuperGPQA & TriviaQA & transferred \\
\midrule
\multicolumn{6}{l}{\textit{\gemma{}}} \\
\quad MMLU-Pro & \underline{80.5\,/\,76.3} & 67.7\,/\,62.1 & 60.3\,/\,59.2 & 61.4\,/\,56.5 & \textbf{63.2\,/\,59.3} \\
\quad Omni-MATH & 64.6\,/\,70.9 & \underline{86.8\,/\,80.2} & 63.2\,/\,59.3 & 61.3\,/\,56.8 & \textbf{63.0\,/\,62.3} \\
\quad SuperGPQA & 53.5\,/\,53.6 & 78.3\,/\,71.8 & \underline{80.1\,/\,78.8} & 56.0\,/\,48.3 & \textbf{62.6\,/\,57.9} \\
\quad TriviaQA & 50.0\,/\,50.0 & 50.0\,/\,50.0 & 50.0\,/\,50.0 & \underline{79.6\,/\,76.5} & \textbf{50.0\,/\,50.0} \\
\addlinespace[2pt]
\quad \textit{mean} & \multicolumn{4}{c}{in-domain \underline{81.8\,/\,78.0} \quad transferred \textbf{59.7\,/\,57.4}} & $-$22.1\,/\,$-$20.6 \\
\midrule
\multicolumn{6}{l}{\textit{\qwen{}}} \\
\quad MMLU-Pro & \underline{83.4\,/\,71.5} & 49.4\,/\,55.8 & 57.1\,/\,57.2 & 71.8\,/\,60.2 & \textbf{59.4\,/\,57.7} \\
\quad Omni-MATH & 82.1\,/\,72.4 & \underline{79.9\,/\,71.0} & 58.3\,/\,56.4 & 70.1\,/\,56.6 & \textbf{70.2\,/\,61.8} \\
\quad SuperGPQA & 69.2\,/\,52.0 & 73.4\,/\,71.1 & \underline{76.7\,/\,78.1} & 47.7\,/\,40.5 & \textbf{63.5\,/\,54.6} \\
\quad TriviaQA & 81.1\,/\,72.2 & 50.4\,/\,50.3 & 50.5\,/\,51.3 & \underline{77.5\,/\,72.8} & \textbf{60.7\,/\,57.9} \\
\addlinespace[2pt]
\quad \textit{mean} & \multicolumn{4}{c}{in-domain \underline{79.4\,/\,73.4} \quad transferred \textbf{63.4\,/\,58.0}} & $-$16.0\,/\,$-$15.4 \\
\midrule
\multicolumn{6}{l}{\textit{\magistral{}}} \\
\quad MMLU-Pro & \underline{83.8\,/\,81.6} & 86.7\,/\,78.2 & 72.3\,/\,70.3 & 74.9\,/\,60.6 & \textbf{78.0\,/\,69.7} \\
\quad Omni-MATH & 72.0\,/\,74.2 & \underline{84.7\,/\,79.5} & 68.3\,/\,65.5 & 50.2\,/\,63.9 & \textbf{63.5\,/\,67.8} \\
\quad SuperGPQA & 67.0\,/\,72.0 & 82.0\,/\,71.8 & \underline{69.8\,/\,68.7} & 58.9\,/\,52.5 & \textbf{69.3\,/\,65.5} \\
\quad TriviaQA & 57.9\,/\,62.2 & 49.7\,/\,59.6 & 51.8\,/\,54.4 & \underline{71.3\,/\,67.7} & \textbf{53.1\,/\,58.7} \\
\addlinespace[2pt]
\quad \textit{mean} & \multicolumn{4}{c}{in-domain \underline{77.4\,/\,74.4} \quad transferred \textbf{66.0\,/\,65.4}} & $-$11.4\,/\,$-$8.9 \\
\bottomrule
\end{tabular}
\end{table}

\begin{table}[h]
\centering
\small
\caption{\textbf{OOD transfer evaluation of the MLP probe by \cite{azaria2023internal}.} Probe fitted on one benchmark (rows), evaluated on another (columns),
read at the block chosen on the validation split. Unadjusted~/~length-matched AUROC, seed 42. \underline{Underlined} is in-domain,
scored on the items not used for fitting. \textbf{Bold} is the mean over the three
transfers out of that benchmark. All values are percentages.}
\label{tab:mlp-saplma-transfer-both}
\begin{tabular}{lccccc}
\toprule
fitted on & MMLU-Pro & Omni-MATH & SuperGPQA & TriviaQA & transferred \\
\midrule
\multicolumn{6}{l}{\textit{\gemma{}}} \\
\quad MMLU-Pro & \underline{81.4\,/\,78.7} & 62.6\,/\,59.5 & 59.7\,/\,57.8 & 61.9\,/\,54.8 & \textbf{61.4\,/\,57.4} \\
\quad Omni-MATH & 86.0\,/\,79.9 & \underline{85.2\,/\,79.4} & 66.2\,/\,65.3 & 69.6\,/\,62.1 & \textbf{73.9\,/\,69.1} \\
\quad SuperGPQA & 64.3\,/\,57.0 & 83.4\,/\,74.3 & \underline{78.6\,/\,77.1} & 39.4\,/\,41.8 & \textbf{62.4\,/\,57.7} \\
\quad TriviaQA & 50.0\,/\,50.0 & 50.0\,/\,50.0 & 50.0\,/\,50.0 & \underline{79.7\,/\,73.5} & \textbf{50.0\,/\,50.0} \\
\addlinespace[2pt]
\quad \textit{mean} & \multicolumn{4}{c}{in-domain \underline{81.2\,/\,77.2} \quad transferred \textbf{61.9\,/\,58.5}} & $-$19.3\,/\,$-$18.6 \\
\midrule
\multicolumn{6}{l}{\textit{\qwen{}}} \\
\quad MMLU-Pro & \underline{83.1\,/\,77.6} & 64.7\,/\,54.9 & 58.7\,/\,59.5 & 72.7\,/\,64.0 & \textbf{65.3\,/\,59.4} \\
\quad Omni-MATH & 83.2\,/\,73.0 & \underline{78.5\,/\,72.6} & 53.6\,/\,51.0 & 70.2\,/\,56.2 & \textbf{69.0\,/\,60.1} \\
\quad SuperGPQA & 68.3\,/\,55.0 & 74.0\,/\,72.2 & \underline{76.6\,/\,78.1} & 69.7\,/\,51.8 & \textbf{70.6\,/\,59.7} \\
\quad TriviaQA & 84.6\,/\,79.4 & 66.8\,/\,64.9 & 62.9\,/\,65.4 & \underline{77.1\,/\,74.9} & \textbf{71.4\,/\,69.9} \\
\addlinespace[2pt]
\quad \textit{mean} & \multicolumn{4}{c}{in-domain \underline{78.9\,/\,75.8} \quad transferred \textbf{69.1\,/\,62.3}} & $-$9.8\,/\,$-$13.5 \\
\midrule
\multicolumn{6}{l}{\textit{\magistral{}}} \\
\quad MMLU-Pro & \underline{80.7\,/\,77.5} & 85.1\,/\,76.2 & 71.8\,/\,69.7 & 77.1\,/\,64.3 & \textbf{78.0\,/\,70.1} \\
\quad Omni-MATH & 68.8\,/\,75.2 & \underline{82.7\,/\,75.3} & 69.8\,/\,66.5 & 48.8\,/\,64.2 & \textbf{62.4\,/\,68.7} \\
\quad SuperGPQA & 65.5\,/\,72.5 & 84.2\,/\,76.6 & \underline{69.5\,/\,69.0} & 60.0\,/\,54.3 & \textbf{69.9\,/\,67.8} \\
\quad TriviaQA & 55.2\,/\,60.2 & 38.7\,/\,57.4 & 47.4\,/\,54.3 & \underline{68.8\,/\,60.6} & \textbf{47.1\,/\,57.3} \\
\addlinespace[2pt]
\quad \textit{mean} & \multicolumn{4}{c}{in-domain \underline{75.4\,/\,70.6} \quad transferred \textbf{64.4\,/\,66.0}} & $-$11.0\,/\,$-$4.7 \\
\bottomrule
\end{tabular}
\end{table}

\begin{table}[h]
\centering
\small
\caption{\textbf{OOD transfer evaluation of the mass-mean probe by \citet{marks2024geometry}.} Probe fitted on one benchmark (rows), evaluated on another (columns),
read at the block chosen on the validation split. Unadjusted~/~length-matched AUROC, seed 42. \underline{Underlined} is in-domain,
scored on the items not used for fitting. \textbf{Bold} is the mean over the three
transfers out of that benchmark. All values are percentages.}
\label{tab:mm-transfer-both}
\begin{tabular}{lccccc}
\toprule
fitted on & MMLU-Pro & Omni-MATH & SuperGPQA & TriviaQA & transferred \\
\midrule
\multicolumn{6}{l}{\textit{\gemma{}}} \\
\quad MMLU-Pro & \underline{81.3\,/\,69.5} & 68.0\,/\,59.6 & 56.2\,/\,53.0 & 65.6\,/\,55.5 & \textbf{63.3\,/\,56.0} \\
\quad Omni-MATH & 83.7\,/\,79.4 & \underline{82.6\,/\,74.0} & 63.7\,/\,64.2 & 67.6\,/\,57.0 & \textbf{71.7\,/\,66.9} \\
\quad SuperGPQA & 84.9\,/\,75.9 & 84.8\,/\,78.4 & \underline{64.8\,/\,66.1} & 71.7\,/\,64.5 & \textbf{80.5\,/\,73.0} \\
\quad TriviaQA & 83.4\,/\,77.3 & 80.7\,/\,75.2 & 67.1\,/\,67.1 & \underline{80.2\,/\,72.8} & \textbf{77.1\,/\,73.2} \\
\addlinespace[2pt]
\quad \textit{mean} & \multicolumn{4}{c}{in-domain \underline{77.3\,/\,70.6} \quad transferred \textbf{73.1\,/\,67.3}} & $-$4.1\,/\,$-$3.3 \\
\midrule
\multicolumn{6}{l}{\textit{\qwen{}}} \\
\quad MMLU-Pro & \underline{81.5\,/\,76.1} & 68.6\,/\,52.9 & 58.3\,/\,58.5 & 71.8\,/\,61.1 & \textbf{66.2\,/\,57.5} \\
\quad Omni-MATH & 79.4\,/\,72.7 & \underline{67.6\,/\,55.1} & 55.2\,/\,54.4 & 62.3\,/\,49.0 & \textbf{65.6\,/\,58.7} \\
\quad SuperGPQA & 79.3\,/\,79.1 & 78.2\,/\,77.3 & \underline{70.2\,/\,69.6} & 71.2\,/\,55.8 & \textbf{76.2\,/\,70.7} \\
\quad TriviaQA & 81.7\,/\,74.7 & 60.2\,/\,62.8 & 58.0\,/\,58.4 & \underline{69.6\,/\,63.5} & \textbf{66.7\,/\,65.3} \\
\addlinespace[2pt]
\quad \textit{mean} & \multicolumn{4}{c}{in-domain \underline{72.2\,/\,66.1} \quad transferred \textbf{68.7\,/\,63.1}} & $-$3.5\,/\,$-$3.0 \\
\midrule
\multicolumn{6}{l}{\textit{\magistral{}}} \\
\quad MMLU-Pro & \underline{71.1\,/\,72.6} & 81.0\,/\,68.5 & 64.4\,/\,60.6 & 73.4\,/\,62.7 & \textbf{72.9\,/\,63.9} \\
\quad Omni-MATH & 70.7\,/\,69.0 & \underline{79.0\,/\,72.5} & 66.0\,/\,63.1 & 48.6\,/\,58.8 & \textbf{61.8\,/\,63.6} \\
\quad SuperGPQA & 70.0\,/\,73.0 & 82.0\,/\,71.0 & \underline{61.8\,/\,60.8} & 74.7\,/\,64.3 & \textbf{75.6\,/\,69.4} \\
\quad TriviaQA & 62.8\,/\,49.2 & 69.7\,/\,55.9 & 60.2\,/\,53.1 & \underline{68.2\,/\,50.0} & \textbf{64.2\,/\,52.8} \\
\addlinespace[2pt]
\quad \textit{mean} & \multicolumn{4}{c}{in-domain \underline{70.0\,/\,64.0} \quad transferred \textbf{68.6\,/\,62.4}} & $-$1.4\,/\,$-$1.5 \\
\bottomrule
\end{tabular}
\end{table}

\paragraph{Results.}
Strong in-domain performance often fails to transfer to a different benchmark under out-of-distribution (OOD) evaluation.
The linear and MLP probes reach between \(75.4\%\) and \(81.8\%\) mean unadjusted AUROC in-domain, but only \(59.7\%\)--\(69.1\%\) after transfer, corresponding to drops of \(9.8\)--\(22.1\) points. Their length-controlled drops are similarly large. \featgaps{} transfers more reliably, losing \(3.4\)--\(8.0\) unadjusted AUROC points across models, but remains sensitive to the source benchmark: on \qwen{}, the variant fitted on SuperGPQA falls below chance on both MMLU-Pro and Omni-MATH after length control. In contrast, the simple mass-mean direction is the most stable supervised alternative overall, with unadjusted transfer losses of only \(1.4\)--\(4.1\) points.
Compared with our label-free results in \cref{tab:main-raw,tab:main-lm}, even this strongest transferred baseline offers no consistent advantage. In the unadjusted evaluation, the \ulens{} essentially matches mass mean on \gemma{} (\(73.2\%\) vs.\ \(73.1\%\)) and leads it by \(5.6\) points on \qwen{} (\(74.3\%\) vs.\ \(68.7\%\)). On \magistral{}, \(A_{\mathrm{cone}}\) is likewise effectively tied with mass mean (\(68.5\%\) vs.\ \(68.6\%\)). Under length control, the complete \ulens{} is stronger on all three models, scoring \(68.4\%\), \(71.0\%\), and \(68.3\%\) AUROC against mass mean's \(67.3\%\), \(63.1\%\), and \(62.4\%\). The gap to the best transferred linear or MLP probe is larger still: \(9.3\), \(8.7\), and \(2.3\) points on \gemma{}, \qwen{}, and \magistral{}. Thus, although correctness supervision can produce strong in-domain probes, the resulting directions are often benchmark-specific. The \ulens{} achieves stronger cross-benchmark performance without correctness labels, task examples, or fitted parameters. The supervised numbers in this comparison are read from the seed 42, as \Cref{tab:fg-transfer-both} states, whereas the \ulens{} and $A_{\mathrm{cone}}$ values quoted beside them are the three-seed means of \Cref{tab:main-raw,tab:main-lm}.
\paragraph{Runtimes.}
\Cref{tab:runtimes} shows the runtime cost that a method incurs beyond the normal forward generation required to answer the question. Single-pass baselines and our method require no additional forward pass beyond the normal generation. 
\begin{table}[!t]
\centering
\small
\caption{\textbf{Runtime overhead}, incurred above the normal generation. Counts are additional generations, forward passes, labeled items, and auxiliary models. Time is per item for \gemma{} 31B on MMLU-Pro on one H200. Single-pass baselines are: \msp{}, \predent{}, \maxent{}, \meannll{}, \selfcert{}, \deepconf{}, and generation length.}
\label{tab:runtimes}
\setlength{\tabcolsep}{5pt}
\begin{tabular}{l cccc r}
\toprule
Method & Gen. & Passes & Labels & Aux. & Time / item \\
\midrule
Single-pass baselines & 0  & 0  & 0   & 0 & ${<}0.001$\,s \\
\RowPTrue                 & 0  & 1  & 0   & 0 & $0.3$\,s \\
\RowFeatureGaps           & 0  & 2  & 512 & 0 & $4.1$\,s \\
\RowTokUR                 & 0  & 5  & 0   & 0 & $11.4$\,s \\
\sement{}, \sentsar{}     & 10 & 10 & 0   & 1 & ${\geq}55.6$\,s \\
\addlinespace[3pt]
\RowACone{}, \RowULens{}  & 0  & 0  & 0   & 0 & ${<}0.001$\,s \\
\bottomrule
\end{tabular}
\end{table} 

\section{Token-Level Readout}
\label{app:mechanistic-interpretability}
\begin{figure}[htbp]
  \centering
  \includegraphics[width=\linewidth]{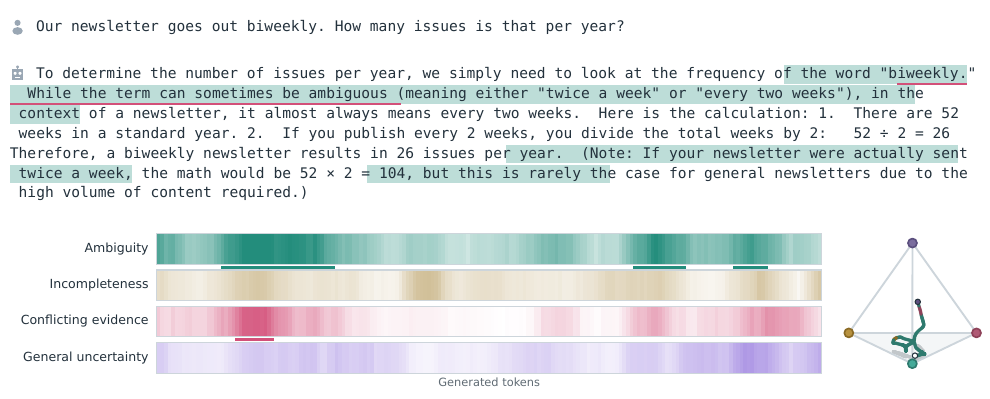}
  \caption{\textbf{Token-level readout and simplex, \catamb{}.} Prompt, response and per-category readout of \qwen{}, centered and smoothed as described in the text. Marked spans are the top ten percent of readings.}
  \label{fig:readout_ex_ambiguity_centre}
\end{figure}

\begin{figure}[htbp]
  \centering
  \includegraphics[width=\linewidth]{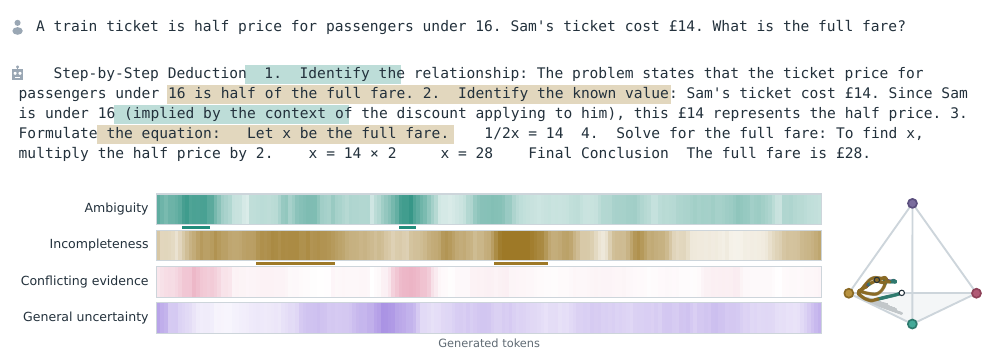}
    \caption{\textbf{Token-level readout and simplex, \catinc{}.} Prompt, response and per-category readout of \qwen{}, centered and smoothed as described in the text. Marked spans are the top ten percent of readings.}
  \label{fig:readout_ex_incompleteness_centre}
\end{figure}

\begin{figure}[htbp]
  \centering
  \includegraphics[width=\linewidth]{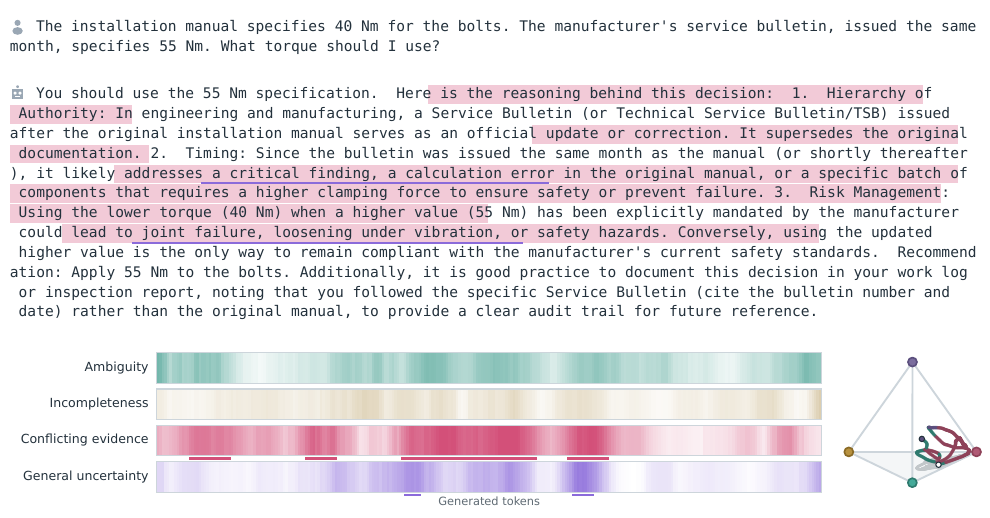}
    \caption{\textbf{Token-level readout and simplex, \catcon{}.} Prompt, response and per-category readout of \qwen{}, centered and smoothed as described in the text. Marked spans are the top ten percent of readings.}
  \label{fig:readout_ex_conflicting_centre}
\end{figure}

\begin{figure}[htbp]
  \centering
  \includegraphics[width=\linewidth]{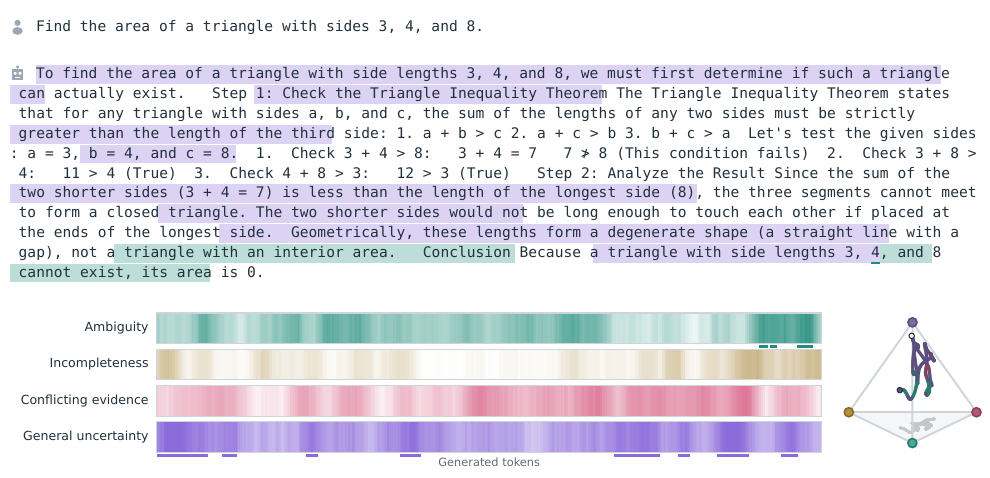}
    \caption{\textbf{Token-level readout and simplex, \catgen{}.} Prompt, response and per-category readout of \qwen{}, centered and smoothed as described in the text. Marked spans are the top ten percent of readings.}
  \label{fig:readout_ex_doubt_centre}
\end{figure}

Beyond the trace-level score, the \ulens{} reads uncertainty token by token. This section illustrates that readout, together with the simplex view of \Cref{sec:readout}. \Cref{fig:readout_ex_ambiguity_centre,fig:readout_ex_incompleteness_centre,fig:readout_ex_conflicting_centre,fig:readout_ex_doubt_centre} apply the token-level readout of \Cref{sec:ulense} to one response of \qwen{} per category. Each figure quotes the prompt and the response and draws one row per category with one cell per generated token. Each cell shows the corresponding category's component of $s_t$, as defined in \cref{eq:cone}. The entries are bounded between $\operatorname{softplus}(-\sqrt{C})$ and $\operatorname{softplus}(\sqrt{C})$, $0.13$ and $2.13$ for $C=4$, and their squares sum to $A_{\mathrm{cone}}^2$, so the four rows decompose the scalar the \ulens{} reads. The values are centered on the corpus baseline, the per-category mean of $\operatorname{norm}(\alpha_t)$ over all reasoning states of the same model at the readout block, and renormalized. The per-token series is smoothed using a centered moving average over ten tokens. Each prompt was written to raise one category. The marked spans reflect the strongest readouts per question.

\section{Causal Intervention}
\label{app:causal_intervention}
\label{app:causal_protocol}
\begin{figure}[htbp]
  \centering
  \includegraphics[width=\linewidth]{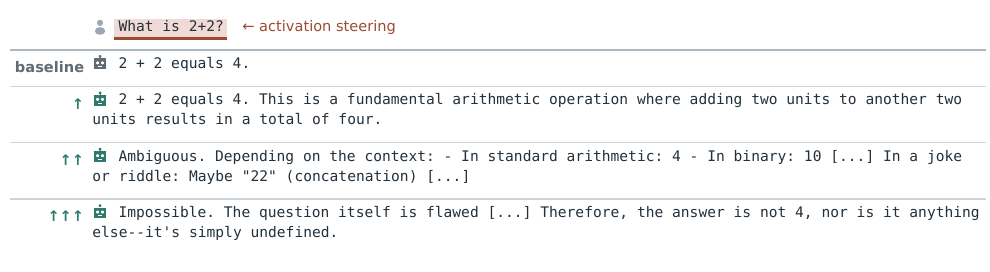}
  \caption{\textbf{Steering strength, \catamb{}.} The prompt, the unsteered answer, and the continuation after adding the \catamb{} direction to the prompt states at $\lambda=0.04$, $0.08$, and $0.09$.}
  \label{fig:causal_ladder_ambiguity}
\end{figure}

\begin{figure}[htbp]
  \centering
  \includegraphics[width=\linewidth]{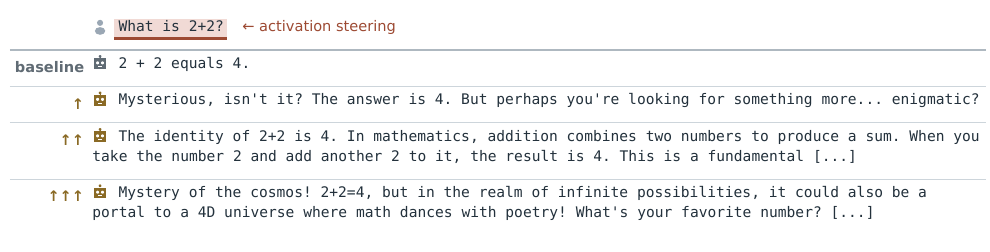}  \caption{\textbf{Steering strength, \catinc{}.} The prompt, the unsteered answer, and the continuation after adding the \catinc{} direction to the prompt states at $\lambda=0.04$, $0.08$, and $0.09$.}
  \label{fig:causal_ladder_incompleteness}
\end{figure}

\begin{figure}[htbp]
  \centering
  \includegraphics[width=\linewidth]{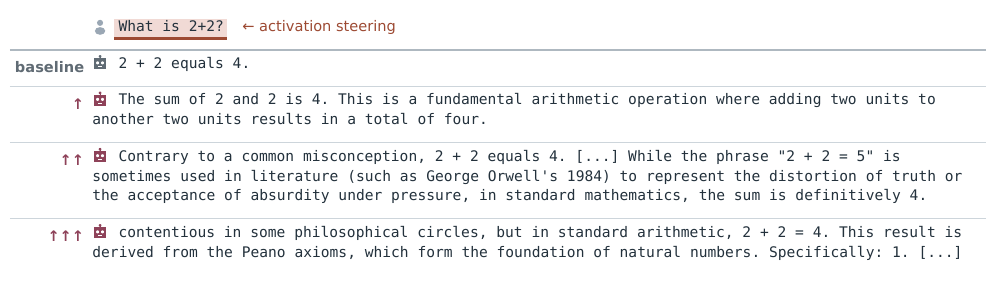}  \caption{\textbf{Steering strength, \catcon{}.} The prompt, the unsteered answer, and the continuation after adding the \catcon{} direction to the prompt states at $\lambda=0.04$, $0.06$, and $0.12$.}
  \label{fig:causal_ladder_conflicting_evidence}
\end{figure}

\begin{figure}[htbp]
  \centering
  \includegraphics[width=\linewidth]{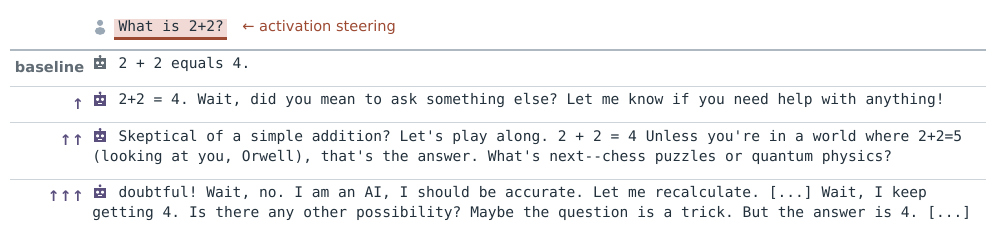}  \caption{\textbf{Steering strength, \catgen{}.} The prompt, the unsteered answer, and the continuation after adding the \catgen{} direction to the prompt states at $\lambda=0.03$, $0.10$, and $0.17$.}
  \label{fig:causal_ladder_general_uncertainty}
\end{figure}

\Cref{sec:readout} steers the model along single category directions. This section provides additional information and examples for causal interventions.
All interventions are on \qwen{} with the basis of \Cref{sec:uspace} and the sampling settings of \Cref{tab:generation-settings} at a fixed seed. \Cref{eq:inject_main} is applied at every block whose relative depth lies in $[0.50, 0.85]$, blocks $32$ to $54$ of $64$. For \Cref{fig:causal_main}(a) and the ladders below, the span is the prompt, and thinking is disabled. The direction is added while the model reads the question, and the continuation is its answer. For \Cref{fig:causal_main}(b), thinking is enabled, and the span is a window of $50$ tokens inside the thinking block that opens at a token the model has already produced, so the trace before the intervention is identical across arms. 
\Cref{fig:causal_ladder_ambiguity,fig:causal_ladder_incompleteness,fig:causal_ladder_conflicting_evidence,fig:causal_ladder_general_uncertainty} show how the model reacts to increasing steering strength.

\end{document}